\documentclass[11pt]{article}
\usepackage[utf8]{inputenc}
\usepackage{amsfonts, amsmath, amsthm, amssymb}
\usepackage[authoryear]{natbib}
\usepackage[margin=1.25in]{geometry}
\usepackage[colorlinks,citecolor=blue,urlcolor=blue]{hyperref}
\usepackage{booktabs}
\usepackage{graphicx}

\newtheorem{theorem}{Theorem}[section]
\newtheorem{lemma}{Lemma}[section]
\newtheorem{assumption}{Assumption}[section]
\newtheorem{corollary}{Corollary}[section]
\newtheorem{example}{Example}[section]

\theoremstyle{remark}
\newtheorem{definition}{Definition}[section]

\newtheorem*{remark}{Remark}

\usepackage{xargs}
\usepackage{amsmath, amsthm, amssymb, amsfonts}
\usepackage{bm}
\usepackage{mathtools}
\usepackage{algorithm, algorithmicx}
\usepackage{subfig}

\newcommand{\bfe}{\mathbf{e}}

\newcommand{\bfh}{\mathbf{h}}

\newcommand{\bfu}{\mathbf{u}}

\newcommand{\bfx}{\mathbf{x}}
\newcommand{\bfy}{\mathbf{y}}

\newcommand{\bfA}{\mathbf{A}}
\newcommand{\bfB}{\mathbf{B}}

\newcommand{\bfD}{\mathbf{D}}

\newcommand{\bfG}{\mathbf{G}}
\newcommand{\bfH}{\mathbf{H}}
\newcommand{\bfI}{\mathbf{I}}
\newcommand{\bfJ}{\mathbf{J}}

\newcommand{\bfL}{\mathbf{L}}

\newcommand{\bfP}{\mathbf{P}}
\newcommand{\bfQ}{\mathbf{Q}}
\newcommand{\bfR}{\mathbf{R}}

\newcommand{\bfT}{\mathbf{T}}
\newcommand{\bfU}{\mathbf{U}}
\newcommand{\bfV}{\mathbf{V}}

\newcommand{\bfX}{\mathbf{X}}
\newcommand{\bfY}{\mathbf{Y}}
\newcommand{\bfZ}{\mathbf{Z}}

\newcommand{\rmE}{\mathrm{E}}

\newcommand{\rmP}{\mathrm{P}}

\newcommand{\calA}{\mathcal{A}}

\newcommand{\calC}{\mathcal{C}}
\newcommand{\calD}{\mathcal{D}}
\newcommand{\calE}{\mathcal{E}}

\newcommand{\calG}{\mathcal{G}}

\newcommand{\calJ}{\mathcal{J}}

\newcommand{\calL}{\mathcal{L}}

\newcommand{\calN}{\mathcal{N}}

\newcommand{\calP}{\mathcal{P}}
\newcommand{\calQ}{\mathcal{Q}}
\newcommand{\calR}{\mathcal{R}}

\newcommand{\calV}{\mathcal{V}}

\newcommand{\calX}{\mathcal{X}}
\newcommand{\calY}{\mathcal{Y}}
\newcommand{\calZ}{\mathcal{Z}}

\newcommand{\bbR}{\mathbb{R}}

\newcommand{\bmalpha}{\bm{\alpha}}

\newcommand{\bmmu}{\bm{\mu}}

\newcommand{\bmtheta}{\bm{\theta}}
\newcommand{\bmTheta}{\bm{\Theta}}
\newcommand{\bmone}{\bm{1}}

\newcommandx{\norm}[3][2={}, 3={}]{\vert\vert #1 \vert\vert_{#2}^{#3}}

\newcommand{\trace}[1]{\mathrm{trace}\{#1\}}
\newcommand{\nullspace}[1]{\mathrm{null}\{#1\}}
\newcommand{\rank}[1]{\mathrm{rank}\{#1\}}
\newcommand{\spanof}[1]{\mathrm{span}\{#1\}}
\newcommand{\lambdamax}[1]{\lambda_{\max}(#1)}
\newcommand{\lambdamin}[1]{\lambda_{\min}(#1)}

\newcommandx{\sumlim}[3][1={i}, 2={1}]{\sum_{#1 = #2}^{#3}}
\newcommandx{\prodlim}[3][1={i}, 2={1}]{\prod_{#1 = #2}^{#3}}

\newcommand{\abs}[1]{\vert #1 \vert}

\newcommand{\const}{\mathrm{const}}

\title{\textbf{On consistency of constrained spectral clustering under representation-aware stochastic block model}}
\author{\textbf{Shubham Gupta} and \textbf{Ambedkar Dukkipati} \\
Department of Computer Science and Automation, \\
Indian Institute of Science, Bangalore-560012, India. \\
\texttt{[shubhamg, ambedkar]@iisc.ac.in}}
\date{}

\begin{document}

\maketitle

\begin{abstract}
    \noindent
    Spectral clustering is widely used in practice due to its flexibility, computational efficiency, and well-understood theoretical performance guarantees. Recently, spectral clustering has been studied to find balanced clusters under population-level constraints. These constraints are specified by additional information available in the form of auxiliary categorical node attributes. In this paper, we consider a scenario where these attributes may not be observable, but manifest as latent features of an auxiliary graph. Motivated by this, we study constrained spectral clustering with the aim of finding balanced clusters in a given \textit{similarity graph} $\calG$, such that each individual is adequately represented with respect to an auxiliary graph $\calR$ (we refer to this as representation graph). We propose an individual-level balancing constraint that formalizes this idea. Our work leads to an interesting stochastic block model that not only plants the given partitions in $\calG$ but also plants the auxiliary information encoded in the representation graph $\calR$. We develop unnormalized and normalized variants of spectral clustering in this setting. These algorithms use $\calR$ to find clusters in $\calG$ that approximately satisfy the proposed constraint. We also establish the first statistical consistency result for constrained spectral clustering under individual-level constraints for graphs sampled from the above-mentioned variant of the stochastic block model. Our experimental results corroborate our theoretical findings.
\end{abstract}

% ============================ %

\section{Introduction}
\label{section:introduction}

Theoretical performance guarantees and adaptability to real-world constraints determine the deployability, and hence the practical utility, of statistical methods and machine learning algorithms. For example, spectral clustering \citep{NgEtAl:2001:OnSpectralClustering,Luxburg:2007:ATutorialOnSpectralClustering}, one of the most sought after algorithms for clustering and community detection, has been studied under various constraints such as \textit{must-link} and \textit{cannot-link} constraints \citep{KamvarEtAl:2003:SpectralLearning, WangDavidson:2010:FlexibleConstrainedSpectralClustering}, size-balanced clusters \citep{BanerjeeGhosh:2006:ScalableClusteringAlgorithmsWithBalancingConstraints}, and statistical fairness \citep{KleindessnerEtAl:2019:GuaranteesForSpectralClusteringWithFairnessConstraints}. Commonly used constraints in practice can be categorized as \textit{population-level} (also known as statistical-level) or \textit{individual-level} constraints. However, the only known consistency guarantees for constrained spectral clustering were established in \citet{KleindessnerEtAl:2019:GuaranteesForSpectralClusteringWithFairnessConstraints} for the first case, where the goal is to find balanced clusters with respect to an auxiliary categorical attribute. In this paper, we study a problem setting where the auxiliary information is encoded in a graph $\calR$, which we refer to as a \textit{representation graph}. We study spectral clustering in a given \textit{similarity graph} $\calG$ under an individual-level balancing constraint that is specified using the representation graph $\calR$. There are two-fold advantages of this setting: \textbf{(i)} our constraint selects clusters that are balanced from each individual's perspective, and \textbf{(ii)} it enables us to define a new variant of the stochastic block model (SBM) \citep{HollandEtAl:1983:StochasticBlockmodelsFirstSteps} that plants the properties of $\calR$ into the sampled graphs, making this variant of SBM \textit{representation-aware} (aware of the representation graph $\calR$). 

%%%%%%%%%%%%%%%%%%%%%%%%%%%%%%%%%%%%%%%%%%%%%%

\subsection{Problem setting and applications}
\label{section:problem_setting_and_applications}

Let $\calG$ denote a similarity graph based on which clusters have to be discovered. We consider a setting where each node in $\calG$ specifies a list of its representatives (other nodes in $\calG$) using an auxiliary representation graph $\calR$. The graph $\calR$ is defined on the same set of nodes as $\calG$ and its edges specify the ``is representative of'' relationship. For example, $\calR$ may be a result of the interactions between individuals that result from values of certain latent node attributes like age and gender. This motivates a \textit{representation constraint} that requires the clusters in $\calG$ to be such that each node has a sufficient number of representatives (as per $\calR$) in all the clusters. Our goal is to develop and analyze variants of the spectral clustering algorithm with respect to this constraint. 

We begin by briefly describing two applications that motivate the above discussed problem. The first application is concerned with the ``fairness'' of clusters. Informally, a node or individual finds the clusters fair if it has a sufficient representation in all the clusters. %Recently, there has been a significant interest in finding fair clusters and various fairness notions have been proposed. 
The work of 
\citet{ChierichettiEtAl:2017:FairClusteringThroughFairlets} requires the clusters to be balanced with respect to various \textit{protected groups} (like gender or race). For example, if $50\%$ of the population is female then the same proportion should be respected in all clusters. This idea of proportional representation has been extended in various forms \citep{RosnerSchmidt:2018:PrivacyPreservingClusteringWithConstraints,BerceaEtAl:2019:OnTheCostOfEssentiallyFairClusterings,BeraEtAl:2019:FairAlgorithmsForClustering} and several efficient algorithms for discovering fair clusters under this notion have been proposed \citep{SchmidtEtAl:2018:FairCoresetsAndStreamingAlgorithmsForFairKMeansClustering,AhmadianEtAl:2019:ClusteringWithoutOverRepresentation,KleindessnerEtAl:2019:GuaranteesForSpectralClusteringWithFairnessConstraints}. While the fairness notion mentioned above is a \textit{statistical} fairness notion (i.e., constraints are applied on protected groups as a whole), \citet{ChenEtAl:2019:ProportionallyFairClustering} and \citet{MahabadiEtAl:2020:IndividualFairnessForKClustering} develop \textit{individual} fairness notions that require examples to be “sufficiently close" to their cluster centroids. \citet{AndersonEtAl:2020:DistributionalIndividualFairnessInClustering} pursue a different direction and adapt the fairness notion proposed by \citet{DworkEtAl:2012:FairnessThroughAwareness} to the problem of clustering. Only \citet{KleindessnerEtAl:2019:GuaranteesForSpectralClusteringWithFairnessConstraints} study spectral clustering in the context of (statistical) fairness. In contrast, we show in Section \ref{section:constraint} that our proposed constraint interpolates between statistical and individual fairness based on the structure of $\calR$.

\begin{figure}[t]
    \centering
    \subfloat[][Protected groups]{\includegraphics[width=0.3\textwidth]{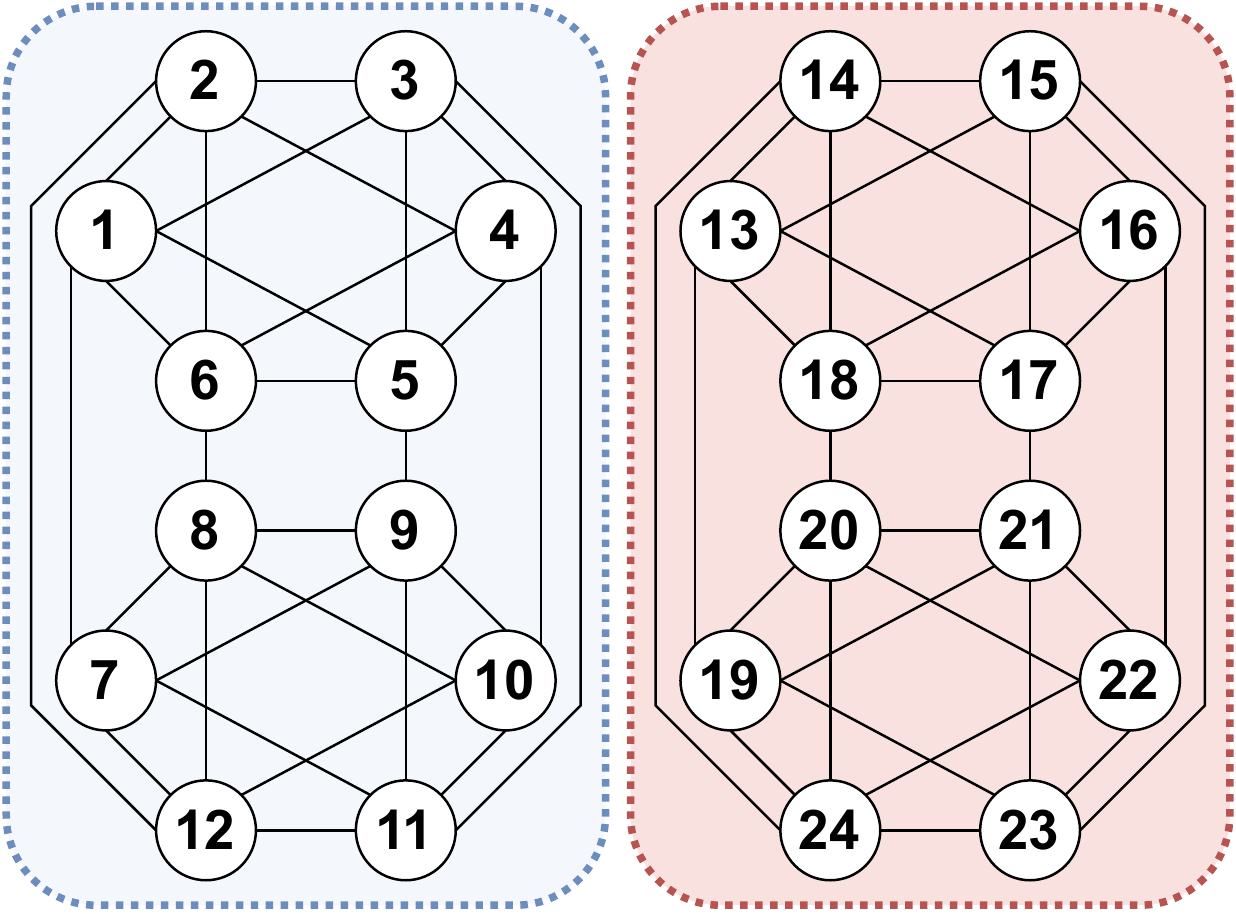}\label{fig:toy_example:protected_groups}}%
    \hspace{5mm}\subfloat[][Statistically fair clusters]{\includegraphics[width=0.3\textwidth]{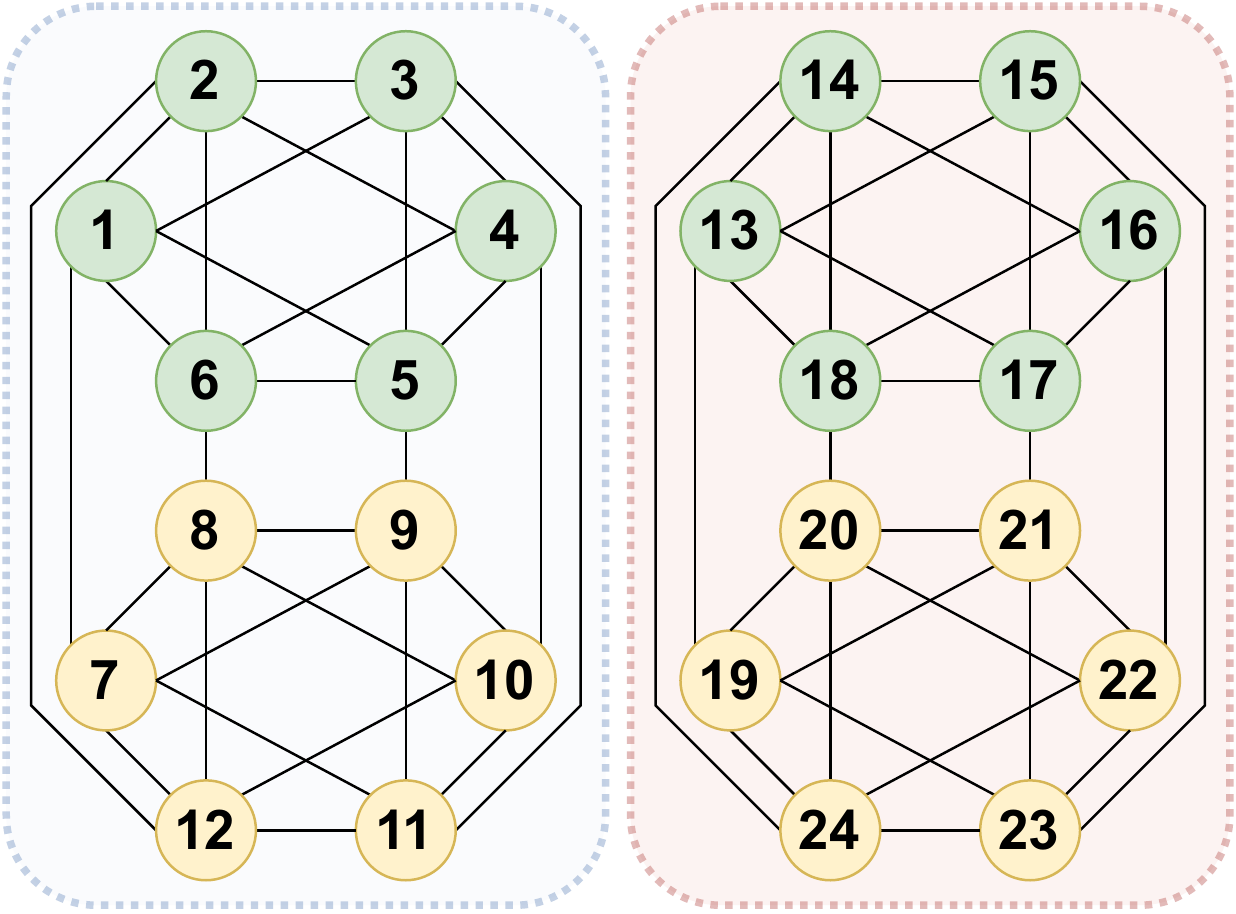}\label{fig:toy_example:fairsc}}%
    \hspace{5mm}\subfloat[][Individually fair clusters]{\includegraphics[width=0.3\textwidth]{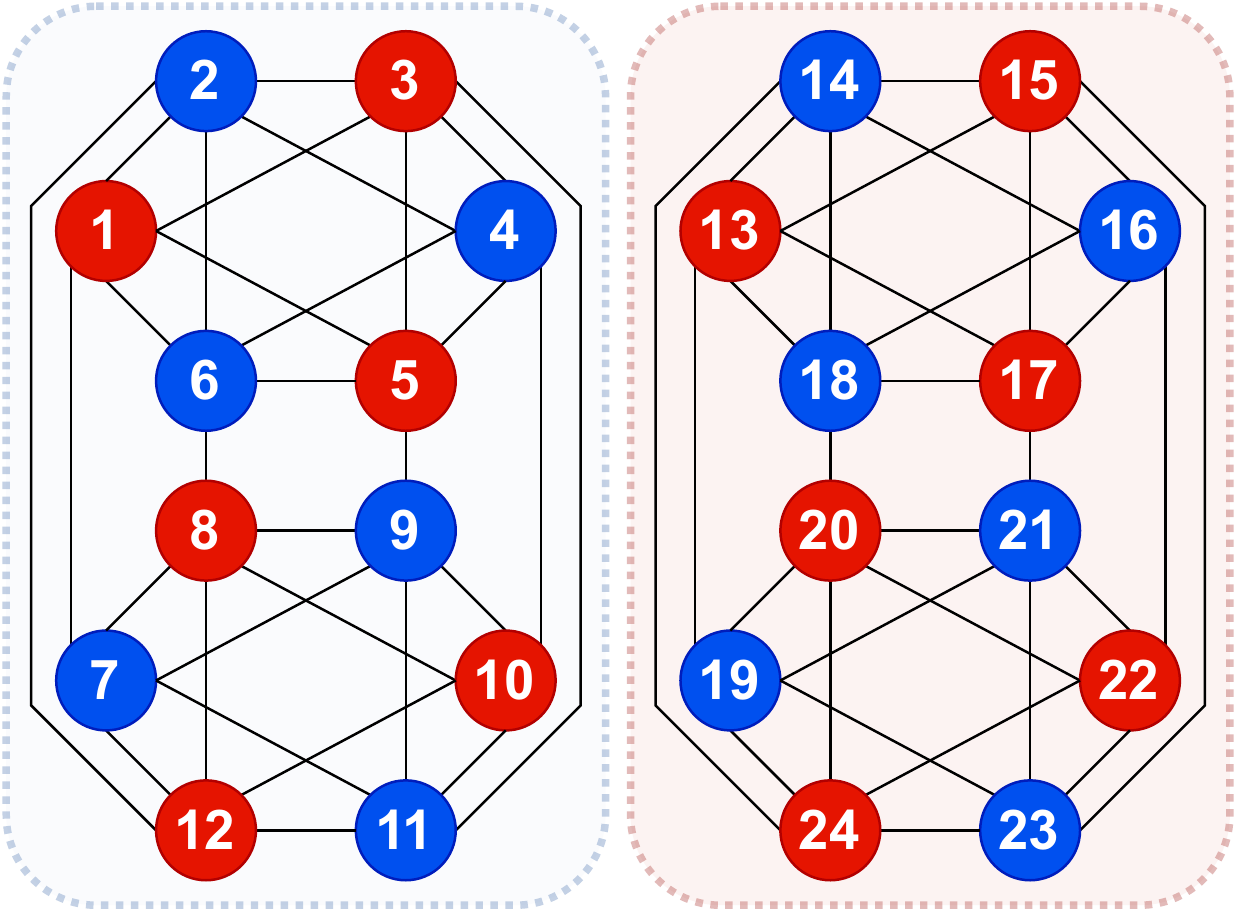}\label{fig:toy_example:repfairsc}}
    \caption{An example representation graph $\calR$. Panel (a) shows the protected groups recovered from $\calR$. Panel (b) shows the clusters recovered by a statistically fair clustering algorithm. Panel (c) shows the ideal individually fair clusters. (Best viewed in color)}
    \label{fig:fairness:toy_example}
\end{figure}

\begin{example}
    \label{example:statistical_vs_individual_fairness}
    To understand the need for individual fairness notions, consider the representation graph $\calR$ specified in Figure \ref{fig:toy_example:protected_groups}. All the nodes have a self-loop associated with them that has not been shown for clarity. In this example, $N = 24$, $K = 2$, and every node is connected to $d=6$ nodes (including the self-loop). To use a statistical fairness notion \citep{ChierichettiEtAl:2017:FairClusteringThroughFairlets}, one would begin by clustering the nodes in $\calR$ to approximate the protected groups, as the members of these protected groups will be each other's representatives to the first order of approximation. A natural choice is to have two protected groups, as shown in Figure~\ref{fig:toy_example:protected_groups} using different colors. However, clustering nodes based on these protected groups can produce the green and yellow clusters shown in Figure~\ref{fig:toy_example:fairsc}. It is easy to verify that these clusters satisfy the statistical fairness criterion as they have an equal number of members from both protected groups. However, these clusters are very "unfair" from the perspective of each individual. For example, node $v_1$ does not have enough representation in the yellow cluster as only one of its six representatives are in this cluster, despite the equal size of both the clusters. A similar argument can be made for every other node in this graph. This example highlights an extreme case where a statistically fair clustering is highly unfair from the perspective of each individual. Figure~\ref{fig:toy_example:repfairsc} shows another clustering assignment, and it is easy to verify that each node in this assignment has the same representation in both red and blue clusters, making it individually fair with respect to $\calR$. Our goal is to develop algorithms that prefer the clusters in Figure~\ref{fig:toy_example:repfairsc} over the clusters in Figure~\ref{fig:toy_example:fairsc}.
\end{example}

Another possible application could be in balancing the load on computing resources in a cloud platform. Here, nodes in $\calG$ correspond to processes and edges encode similarity among these processes in terms of shareable resources such as a read-only file. The edges in $\calR$ on the other hand connect processes that share resources that can only be accessed by one process at a time (say a network channel). The goal is to cluster similar processes in $\calG$ while ensuring that neighbors in $\calR$ are spread out across clusters to avoid a collision. 
%Appendix \ref{appendix:constraint} describes these applications in detail.

%%%%%%%%%%%%%%%%%%%%%%%%%%%%%%%%%%%%%%%%%%%%%%

\subsection{Contributions and results}
\label{section:contributions_and_results}

Our primary goal is to establish the statistical consistency of constrained spectral clustering, where the constraints are specified from the perspective of each individual node in the graph. Towards this end, we make four contributions.

First, in Section \ref{section:constraint}, we introduce the notion of balance from the perspective of each individual node in the graph and  formally specify our representation constraint as a simple linear expression. While we focus on spectral clustering in this paper, the proposed constraint may be of independent interest for other clustering techniques as well. 

Second, in Sections~\ref{section:unnormalized_repsc} and \ref{section:normalized_repsc}, we develop \textit{representation-aware} variants of the unnormalized and normalized spectral clustering. The proposed algorithms incorporate our representation constraint as a linear constraint in spectral clustering's optimization objective. The resulting problem can be easily solved using eigen-decomposition and the returned clusters approximately satisfy the constraint. 

Third, in Section~\ref{section:rsbm}, we propose a variant of SBM called \textit{representation-aware} stochastic block model ($\calR$-SBM). $\calR$-SBM encodes a probability distribution over similarity graphs $\calG$ conditioned on a given representation graph $\calR$. It can be viewed as a model that plants the properties of $\calR$ into $\calG$. We show that $\calR$-SBM generates similarity graphs that present a ``hard'' problem instance to the spectral algorithms in a constrained setting. In Section~\ref{section:consistency_results}, we consider the class of $d$-regular representation graphs and establish the weak-consistency of our algorithms (Theorems \ref{theorem:consistency_result_unnormalized} and \ref{theorem:consistency_result_normalized}) for graphs sampled from $\calR$-SBM. To the best of our knowledge, these are the first consistency results for constrained spectral clustering under individual-level constraints.

Fourth, in Section~\ref{section:numerical_results}, we present empirical studies on both simulated and real-world data to verify our theoretical guarantees. A comparison between the performance of the proposed algorithms and their closest counterparts in the literature demonstrates their practical utility. In particular, our experiments show that the $d$-regularity assumption on the representation graph is not necessary in practice. 

We conclude the paper in Section \ref{section:conclusion} with a few remarks on promising directions for future work. The proofs of all technical lemmas are presented in the supplementary material \citep{ThisPaperSupp}.

%%%%%%%%%%%%%%%%%%%%%%%%%%%%%%%%%%%%%%%%%%%%%%

\subsection{Related results}
\label{section:related_results}
Several algorithms for unconstrained clustering such as $k$-means \citep{HofmannBuhmann:1997:PairwiseDataClusteringByDeterministicAnnealing, WagstaffEtAl:2001:ConstrainedKMeansClusteringWithBackgroundKnowledge}, expectation-maximization based clustering \citep{ShentalEtAl:2003:ComputingGaussianMixtureModelsWithEMUsingEquivalenceConstraints}, and spectral clustering \citep{KamvarEtAl:2003:SpectralLearning} have been modified to satisfy the given \textit{must-link} (ML) and \textit{cannot-link} (CL) constraints \citep{BasuEtAl:2008:ConstrainedClustering} that specify pairs of nodes that should or should not belong to the same cluster. In this paper, we restrict our focus to spectral clustering as it provides a deterministic solution to the clustering problem in polynomial time and can detect arbitrarily shaped clusters \citep{Luxburg:2007:ATutorialOnSpectralClustering}. Existing approaches modify spectral clustering by preprocessing the input similarity graph \citep{KamvarEtAl:2003:SpectralLearning, LuCarreiraPerpinan:2008:ConstrainedSpectralClusteringThroughAffinityPropagation}, post-processing the eigenvectors of the Laplacian matrix \citep{LiEtAl:2009:ConstrainedClusteringViaSpectralRegularization}, or modifying the optimization problem solved by spectral clustering \citep{YuShi:2001:GroupingWithBias,YuShi:2004:SegmentationGivenPartialGroupingConstraints, BieEtAl:2004:LearningFromGeneralLabelConstraints, WangDavidson:2010:FlexibleConstrainedSpectralClustering,ErikssonEtAl:2011:NormalizedCutsRevisited, KawaleBoley:2013:ConstrainedSpectralClusteringUsingL1Regularization, WangEtAl:2014:OnConstrainedSpectralClusteringAndItsApplications}. Researchers have also studied spectral approaches that, for example, handle inconsistent \citep{ColemanEtAl:2008:SpectralClusteringWithInconsistentAdvice} or sparse \citep{ZhuEtAl:2013:ConstrainedClustering} constraints, actively solicit constraints \citep{WangDavidson:2010:ActiveSpectralClustering}, or modify variants of spectral clustering \citep{RangapuramHein:2012:Constrained1SpectralClustering, WangEtAl:2009:IntegratedKLClustering}, to name but a few. Other types of constraints such as those on cluster sizes \citep{BanerjeeGhosh:2006:ScalableClusteringAlgorithmsWithBalancingConstraints, DemirizEtAl:2008:UsingAssignmentConstraintsToAvoidEmptyClustersInKMeansClustering} and those that can be expressed as linear expressions \citep{XuEtAl:2009:FastNormalizedCutWithLinearConstraints} have also been explored. While this has been an active area of research, theoretical consistency guarantees on the performance of these constrained spectral clustering algorithms are largely missing from the literature.

(Unconstrained) spectral clustering is backed by strong statistical guarantees that usually take the form ``the algorithm makes $o(N)$ mistakes with probability $1 - o(1)$.'' Here, $N$ is the number of nodes in the similarity graph. These results consider a random model that generates problem instances for the algorithm (similarity graph in this case) with known ground-truth clusters. The high probability bound is with respect to this random model and the mistakes are computed against the ground-truth clusters. An algorithm that satisfies the condition above is called \textit{weakly consistent} \citep{Abbe:2018:CommunityDetectionAndStochasticBlockModels}. A common choice for the random model is the Stochastic Block Model (SBM) \citep{HollandEtAl:1983:StochasticBlockmodelsFirstSteps}. In this model, nodes have predefined community memberships that are used to sample edges with different probabilities. \citet{RoheEtAl:2011:SpectralClusteringAndTheHighDimensionalSBM} established the weak consistency of spectral clustering under the SBM. \citet{LeiEtAl:2015:ConsistencyOfSpectralClusteringInSBM} instead used a variant of SBM to sample networks with a more realistic degree distribution \citep{KarrerNewman:2011:StochasticBlockmodelsAndCommunityStructureInNetworks}. Several other variants of SBM have also been used to provide appropriate problem instances such as graphs with overlapping clusters \citep{ZhangEtAl:2014:DetectingOverlappingCommunitiesInNetworksUsingSpectralMethods}, observable node covariates \citep{BinkiewiczEtAl:2017:CovariateAssistedSpectralClustering}, or two alternative set of clusters, one \textit{unfair} and one \textit{fair} \citep{KleindessnerEtAl:2019:GuaranteesForSpectralClusteringWithFairnessConstraints}. \citet{LuxburgEtAl:2008:ConsistencyOfSpectralClustering} use a different random model where the similarity graph encodes pairwise cosine similarity between input feature vectors that follow a particular probability distribution. \citet{TremblayEtAl:2016:CompressiveSpectralClustering} study a variant of spectral clustering that is faster than the traditional algorithm. \textit{Strong consistency} results are also known in some cases \citep{GaoEtAl:2017:AchievingOptimalMisclassificationProportionInStochasticBlockModels,LeiZhu:2017:AGenericSampleSplittingApproachForRefinedCommunityRecoveryInSBMs, VuEtAl:2018:ASimpleSVDAlgorithmForFindingHiddenPartitions}. Finally, \citet{GhoshdastidarDukkipati:2017:UniformHypergraphPartitioning, GhoshdastidarDukkipati:2017:ConsistencyOfSpectralHypergraphPartitioningUnderPlantedPartitionModel} propose a variant of SBM for hypergraphs and establish the weak consistency of spectral algorithms in this setting. 

Theoretical results for constrained clustering are primarily concerned with the computational complexity of the problem. It is known that the problem is NP-hard if one hopes for an exact solution that satisfies all the CL constraints \citep{DavidsonRavi:2005:ClusteringWithConstraints} or the statistical fairness constraint \citep{ChierichettiEtAl:2017:FairClusteringThroughFairlets}. Hence, most existing methods only satisfy the constraints approximately \citep{CucuringuEtAl:2016:SimpleAndScalableConstrainedClustering}. Results related to the cluster quality in the constrained setting only consider the algorithm's convergence to the global optima of a relaxed optimization problem \citep{XuEtAl:2009:FastNormalizedCutWithLinearConstraints}. \citet{KleindessnerEtAl:2019:GuaranteesForSpectralClusteringWithFairnessConstraints} is a notable exception, however, even they consider constraints that apply at the level of protected groups as explained above, and not at the level of individuals. They follow a similar strategy as ours and modify spectral clustering to add a statistical fairness constraint. In Section~\ref{section:constraint}, we argue that a particular configuration of the representation graph $\calR$ reduces our constraint to the statistical fairness criterion. Thus, the algorithms proposed in \citet{KleindessnerEtAl:2019:GuaranteesForSpectralClusteringWithFairnessConstraints} are strictly special cases of the algorithms presented in this paper. To the best of our knowledge, we are the first to establish statistical consistency results for constrained spectral clustering for individual-level constraints. A subset of results presented in this paper are available online as a preliminary draft \citep{ThisPaperArXiv}.

%%%%%%%%%%%%%%%%%%%%%%%%%%%%%%%%%%%%%%%%%%%%%%

\section{Notation and preliminaries}
\label{section:notation_and_preliminaries}

Let $\calG = (\calV, \calE)$ denote a similarity graph, where $\calV = \{v_1, v_2, \dots, v_N\}$ is the set of $N$ nodes, and $\calE \subseteq \calV \times \calV$ is the set of edges. The aim of clustering is to partition the nodes into $K \geq 2$ clusters $\calC_1, \dots, \calC_K \subseteq \calV$ such that each node in $\calV$ belongs to exactly one cluster, i.e., $\calC_i \cap \calC_j = \phi$ if $i \neq j$ and $\cup_{k=1}^K \calC_k = \calV$. We further assume the availability of a \textit{representation graph}, denoted by $\calR = (\calV, \hat{\calE})$, that encodes auxiliary information. Notice that $\calR$ is defined on the same set of vertices as the similarity graph $\calG$, but has a different set of edges $\hat{\calE} \subseteq \calV \times \calV$. The discovered clusters $\calC_1, \dots, \calC_K$ are required to satisfy the constraint encoded by $\calR$, as described in Section \ref{section:constraint}. $\bfA \in \{0, 1\}^{N \times N}$ and $\bfR \in \{0, 1\}^{N \times N}$ denote the adjacency matrices of graphs $\calG$ and $\calR$, respectively. We assume that $\calG$ and $\calR$ are undirected. Further, $\calG$ has no self-loops. Thus, $\bfA$ and $\bfR$ are symmetric and $A_{ii} = 0$ for all $i \in [N]$, where $[n] \coloneqq \{1, 2, \dots, n\}$ for any integer $n$.

We propose modified variants of spectral clustering in Section \ref{section:algorithms}. Before describing the proposed algorithms, we begin with a brief review of the standard spectral clustering algorithm.

%%%%%%%%%%%%%%%%%%%%%%%%%%%%%%%%%%%%%%%%%%%%%%

\subsection{Unnormalized spectral clustering}
\label{section:unnormalized_spectral_clustering}

Given a similarity graph $\calG$, unnormalized spectral clustering finds clusters by approximately optimizing a quality metric known as the ratio-cut defined as \citep{Luxburg:2007:ATutorialOnSpectralClustering}
\begin{equation*}
    \mathrm{RCut}(\calC_1, \dots, \calC_K) = \sum_{i = 1}^K \frac{\mathrm{Cut}(\calC_i, \calV \backslash \calC_i)}{\abs{\calC_i}}.
\end{equation*}
Here, $\calV \backslash \calC_i$ denotes the set difference between sets $\calV$ and $\calC_i$. For any two subsets $\calX, \calY \subseteq \calV$, $\mathrm{Cut}(\calX, \calY)$ is defined as
$\mathrm{Cut}(\calX, \calY) = \frac{1}{2} \sum_{v_i \in \calX, v_j \in \calY} A_{ij}$. That is, $\mathrm{Cut}(\calX, \calY)$ counts the number of edges that have one endpoint in $\calX$ and another endpoint in $\calY$. 
%Assuming that similar nodes tend to connect more often than dissimilar ones, it is reasonable to expect that a \textit{good} community $\calC_i$ to have a low $\mathrm{Cut}(\calC_i, \calV \backslash \calC_i)$ value. However, on its own, cut is not a good measure of the quality of a community. As communities get larger, $\mathrm{Cut}(\calC_i, \calV \backslash \calC_i)$ invariably increases because nodes in real-world networks do connect with nodes outside their community, albeit with a smaller probability. The ratio-cut objective addresses this issue by dividing $\mathrm{Cut}(\calC_i, \calV \backslash \calC_i)$ by the size of the community $\calC_i$. Communities $\calC_1$, \dots, $\calC_K$ that minimize the ratio-cut objective are thus sparsely connected to each other.
%We need additional notation to specify the optimization problem solved by unnormalized spectral clustering. 
The Laplacian matrix $\bfL$ of the similarity graph $\calG$ is defined as 
\begin{equation}
    \label{eq:L_def}
    \bfL = \bfD - \bfA.
\end{equation}
Here, $\bfD \in \bbR^{N \times N}$ is the degree matrix, which is a diagonal matrix such that $D_{ii} = \sum_{j = 1}^N A_{ij}$, for all $i \in [N]$. 
%The matrix $\bfD$ is often referred to as the \textit{degree} matrix of $\calG$.
Further, define $\bfH \in \bbR^{N \times K}$ as
\begin{equation}
    \label{eq:H_def}
    H_{ij} = \begin{cases}
        \frac{1}{\sqrt{\abs{\calC_j}}} & \text{ if }v_i \in \calC_j \\
        0 & \text{ otherwise.}
    \end{cases}
\end{equation}
%Note that the matrix $\bfH$ uniquely identifies clusters $\calC_1, \dots, \calC_K$ and vice-versa. While $\bfH$ is a function of the clusters under consideration, we suppress this in the notation to avoid unnecessary clutter. 
One can easily verify that $\mathrm{RCut}(\calC_1, \dots, \calC_K) = \trace{\bfH^\intercal \bfL \bfH}$, where $\bfH$ corresponds to clusters $\calC_1, \dots, \calC_K$. Thus, to find good clusters, one can solve:
\begin{equation*}
    \min_{\bfH \in \bbR^{N \times K}}  \,\,\,\, \trace{\bfH^\intercal \bfL \bfH} \,\,\,\, \text{s.t.} \,\,\,\, \bfH \text{ is of the form \eqref{eq:H_def}.}
\end{equation*}
It is computationally hard to solve this optimization problem due to the combinatorial nature of the constraint \citep{WagnerWagner:1993:BetweenMinCutAndGraphBisection}. Unnormalized spectral clustering instead solves the following relaxed optimization problem:
\begin{equation}
    \label{eq:opt_problem_normal}
    \min_{\bfH \in \bbR^{N \times K}}  \,\,\,\, \trace{\bfH^\intercal \bfL \bfH} \,\,\,\, \text{s.t.} \,\,\,\, \bfH^\intercal \bfH = \bfI.
\end{equation}
The above relaxation is often referred to as the spectral relaxation. 
%Note that if $\bfH$ is of the form \eqref{eq:H_def}, then indeed $\bfH^\intercal \bfH = \bfI$, where $\bfI$ is an identity matrix of appropriate dimensions. 
By Rayleigh-Ritz theorem \citep[Section 5.2.2]{Lutkepohl:1996:HandbookOfMatrices}, 
%the solution to this trace minimization problem is given by the $K$ leading eigenvectors of $\bfL$. That is, 
the optimal matrix $\bfH^*$ is such that it has $\bfu_1, \bfu_2, \dots, \bfu_K \in \bbR^N$ as its columns, where $\bfu_i$ is the eigenvector corresponding to the $i^{th}$ smallest eigenvalue of $\bfL$ for all $i \in [K]$.
%Because \eqref{eq:opt_problem_normal} solves a relaxed variant of the ratio-cut minimization problem, the optimal matrix $\bfH^*$ is unlikely to have the form given in \eqref{eq:H_def}. Thus, 
The algorithm clusters the rows of $\bfH^*$ into $K$ clusters using $k$-means clustering \citep{Lloyd:1982:LeastSquaresQuantisationInPCM} to return $\hat{\calC}_1, \dots, \hat{\calC}_K$. Algorithm \ref{alg:unnormalized_spectral_clustering} summarizes this procedure.

The Laplacian given in \eqref{eq:L_def} is more specifically known as unnormalized Laplacian. The next subsection describes a variant of spectral clustering, known as normalized spectral clustering \citep{ShiMalik:2000:NormalizedCutsAndImageSegmentation, NgEtAl:2001:OnSpectralClustering}, that uses the normalized Laplacian. Unless stated otherwise, we will use spectral clustering (without any qualification) to refer to unnormalized spectral clustering.

\begin{algorithm}[t]
    \begin{algorithmic}[1]
        \State \textbf{Input:} Adjacency matrix $\bfA$, number of clusters $K \geq 2$
        \State Compute the Laplacian matrix $\bfL = \bfD - \bfA$.
        \State Compute the first $K$ eigenvectors $\bfu_1, \dots, \bfu_K$ of $\bfL$. Let $\bfH^* \in \bbR^{N \times K}$ be a matrix that has $\bfu_1, \dots, \bfu_K$ as its columns.
        \State Let $\bfh^*_i$ denote the $i^{th}$ row of $\bfH^*$. Cluster $\bfh^*_1, \dots, \bfh^*_N$ into $K$ clusters using $k$-means clustering.
        \State \textbf{Output:} Clusters $\hat{\calC}_1, \dots, \hat{\calC}_K$, \textrm{s.t.} $\hat{\calC}_i = \{v_j \in \calV : \bfh^*_j \text{ was assigned to the }i^{th} \text{ cluster}\}$.
    \end{algorithmic}
    \caption{Unnormalized spectral clustering}
    \label{alg:unnormalized_spectral_clustering}
\end{algorithm}

%%%%%%%%%%%%%%%%%%%%%%%%%%%%%%%%%%%%%%%%%%%%%%

\subsection{Normalized spectral clustering}
\label{section:normalized_spectral_clustering}

The ratio-cut objective divides $\mathrm{Cut}(\calC_i, \calV \backslash \calC_i)$ by the number of nodes in $\calC_i$ to balance the size of the clusters. The volume of a cluster $\calC \subseteq \calV$, defined as $\mathrm{Vol}(\calC) = \sum_{v_i \in \calC} D_{ii}$, is another popular notion of its size. The normalized cut or $\mathrm{NCut}$ objective divides $\mathrm{Cut}(\calC_i, \calV \backslash \calC_i)$ by $\mathrm{Vol}(\calC_i)$, and is defined as,
\begin{equation*}
    \mathrm{NCut}(\calC_1, \dots, \calC_K) = \sum_{i = 1}^K \frac{\mathrm{Cut}(\calC_i, \calV \backslash \calC_i)}{\mathrm{Vol}(\calC_i)}.
\end{equation*}
As before, one can show that $\mathrm{NCut}(\calC_1, \dots, \calC_K) = \trace{\bfT^\intercal \bfL \bfT}$ \citep{Luxburg:2007:ATutorialOnSpectralClustering}, where $\bfT \in \bbR^{N \times K}$ is specified below.
\begin{equation}
    \label{eq:T_def}
    T_{ij} = \begin{cases}
        \frac{1}{\sqrt{\mathrm{Vol}(\calC_j)}} & \text{ if }v_i \in \calC_j \\
        0 & \text{ otherwise.}
    \end{cases}
\end{equation}
Note that $\bfT^\intercal \bfD \bfT = \bfI$. Thus, the optimization problem for minimizing the NCut objective is
\begin{equation}
    \label{eq:normalized_ideal_opt_problem}
    \min_{\bfT \in \bbR^{N \times K}}  \,\,\,\, \trace{\bfT^\intercal \bfL \bfT} \,\,\,\, \text{s.t.} \,\,\,\, \bfT^\intercal \bfD \bfT = \bfI \text{ and } \bfT \text{ is of the form \eqref{eq:T_def}.}
\end{equation}
As before, this optimization problem is hard to solve, and normalized spectral clustering solves a relaxed variant of this problem. Let $\bfH = \bfD^{1/2} \bfT$ and define the normalized graph Laplacian as $\bfL_{\mathrm{norm}} = \bfI - \bfD^{-1/2} \bfA \bfD^{-1/2}$. Normalized spectral clustering solves the following relaxed problem:
\begin{equation}
    \label{eq:opt_problem_normal_normalized}
    \min_{\bfH \in \bbR^{N \times K}}  \,\,\,\, \trace{\bfH^\intercal \bfL_{\mathrm{norm}} \bfH} \,\,\,\, \text{s.t.} \,\,\,\, \bfH^\intercal \bfH = \bfI.
\end{equation}
Note that $\bfH^\intercal \bfH = \bfI \Leftrightarrow \bfT^\intercal \bfD \bfT = \bfI$. This is again the standard form of the trace minimization problem that can be solved using the Rayleigh-Ritz theorem. Algorithm \ref{alg:normalized_spectral_clustering} summarizes the normalized spectral clustering algorithm. 
%Besides computing the eigenvectors of the normalized Laplacian $\bfL_{\mathrm{norm}}$ instead of the unnormalized Laplacian $\bfL$, Algorithm \ref{alg:normalized_spectral_clustering} also normalizes the rows of the optimal matrix $\bfH^*$ before the $k$-means step. This improves the performance of the algorithm on nodes with a relatively small degree \citep{Luxburg:2007:ATutorialOnSpectralClustering}.

\begin{algorithm}[t]
    \begin{algorithmic}[1]
        \State \textbf{Input:} Adjacency matrix $\bfA$, number of clusters $K \geq 2$
        \State Compute the normalized Laplacian matrix $\bfL_{\mathrm{norm}} = \bfI - \bfD^{-1/2}\bfA \bfD^{-1/2}$.
        \State Compute the first $K$ eigenvectors $\bfu_1, \dots, \bfu_K$ of $\bfL_{\mathrm{norm}}$. Let $\bfH^* \in \bbR^{N \times K}$ be a matrix that has $\bfu_1, \dots, \bfu_K$ as its columns.
        \State Let $\bfh^*_i$ denote the $i^{th}$ row of $\bfH^*$. Compute $\tilde{\bfh}^*_i = \frac{\bfh^*_i}{\norm{\bfh^*_i}[2]}$ for all $i = 1, 2, \dots, N$.
        \State Cluster $\tilde{\bfh}^*_1, \dots, \tilde{\bfh}^*_N$ into $K$ clusters using $k$-means clustering.
        \State \textbf{Output:} Clusters $\hat{\calC}_1, \dots, \hat{\calC}_K$, \textrm{s.t.} $\hat{\calC}_i = \{v_j \in \calV : \tilde{\bfh}^*_j \text{ was assigned to the }i^{th} \text{ cluster}\}$.
    \end{algorithmic}
    \caption{Normalized spectral clustering}
    \label{alg:normalized_spectral_clustering}
\end{algorithm}

%%%%%%%%%%%%%%%%%%%%%%%%%%%%%%%%%%%%%%%%%%%%%%

\section{Representation constraint and representation-aware spectral clustering}
\label{section:algorithms}

%In this section, we first describe our representation constraint and then develop variants of unnormalized and normalized spectral clustering which return clusters that approximately satisfy this constraint. We call these algorithms \textit{Representation-Aware Spectral Clustering} or \textsc{RepSC} and use the prefix \textsc{U} or \textsc{N} to distinguish between the unnormalized and normalized variant, as in \textsc{URepSC} and \textsc{NRepSC}.

%%%%%%%%%%%%%%%%%%%%%%%%%%%%%%%%%%%%%%%%%%%%%%

\subsection{Representation constraint}
\label{section:constraint}

Here, an individual-level constraint for clustering the graph $\calG = (\calV, \calE)$ is specified using a representation graph $\calR = (\calV, \hat{\calE})$.
%This graph is defined on the same set of nodes $\calV$ as the similarity graph $\calG$, but has a different set of edges $\hat{\calE}$. 
%that enforces a representation requirement on the clusters $\calC_1, \dots, \calC_K$ discovered in $\calG$. 
One intuitive explanation for $\calR$ is that nodes connected in this graph represent each other in some form (say based on opinions if nodes correspond to people). 
%To motivate this, consider fairness as an example, and assume that two nodes are connected in $\calR$ if they can represent each other's viewpoint in different clusters. 
Let $\calN_{\calR}(i) = \{v_j \; : \; R_{ij} = 1\}$ denote the set of neighbors of node $v_i$ in $\calR$. The size of $\calN_{\calR}(i) \cap \calC_k$ denotes node $v_i$'s representation in cluster $\calC_k$. 
%is determined by the number of its representatives in $\calN_{\calR}(i)$ that belongs to $\calC_k$.
The goal of this constraint is to ensure that each node has an adequate amount of representation in all clusters. %Towards this end, we define a notion of balance for the clusters from the perspective of each individual.

%Central to the specification of the proposed constraint is the notion of a \textit{representation graph} $\calR = (\calV, \hat{\calE})$. This graph is defined on the same set of nodes $\calV$ as the similarity graph $\calG$, but has a different set of edges $\hat{\calE}$. We are interested in enforcing a \textit{representation} requirement on the clusters $\calC_1, \dots, \calC_K$ discovered in $\calG$. One intuitive explanation for $\calR$ is that nodes connected in this graph represent each other in some form (say based on opinions if nodes correspond to people). 
%To motivate this, consider fairness as an example, and assume that two nodes are connected in $\calR$ if they can represent each other's viewpoint in different clusters. Let $\calN_{\calR}(i) = \{v_j \; : \; R_{ij} = 1\}$ denote the set of neighbors of node $v_i$ in $\calR$. The size of $\calN_{\calR}(i) \cap \calC_k$ denotes node $v_i$'s representation in cluster $\calC_k$. 
%is determined by the number of its representatives in $\calN_{\calR}(i)$ that belongs to $\calC_k$. Informally, the goal of the proposed representation constraint is to ensure that each node has an adequate amount of representation in all clusters. Towards this end, we define a notion of balance for the clusters from the perspective of each individual.

\begin{definition}
    \label{def:balance}
    The balance of given clusters $\calC_1, \dots, \calC_K$ with respect to a node $v_i \in \calV$ is defined as
    \begin{equation}
        \label{eq:balance}
        \rho_i = \min_{k, \ell \in [K]} \;\; \frac{\abs{\calC_k \cap \calN_{\calR}(i)}}{\abs{\calC_\ell \cap \calN_{\calR}(i)}}.
\end{equation}
\end{definition}

It is easy to see that $0 \leq \rho_i \leq 1$ and higher values of $\rho_i$ indicate that node $v_i$ has an adequate representation in all clusters. Thus, one objective could be to find clusters $\calC_1, \dots, \calC_K$ that solve the following constrained optimization problem.
\begin{equation}
    \label{eq:general_optimization_problem}
    \min_{\calC_1, \dots, \calC_K} \;\; f(\calC_1, \dots, \calC_K) \;\;\;\; \text{s.t.} \;\;\;\; \rho_i \geq \alpha, \; \forall \; i \in [N].
\end{equation}
Here, $f(\cdot)$ is a function that is inversely proportional to the quality of clusters (such as $\mathrm{RCut}$ or $\mathrm{NCut}$), and $\alpha \in [0, 1]$ is a user specified threshold. However, it not clear as to how this approach can be combined with spectral clustering to develop a consistent algorithm. Therefore, we take a slightly different approach, as described below.

First, note that $\rho_i \leq \min_{k, \ell \in [K]} \; \frac{\abs{\calC_k}}{\abs{\calC_\ell}}$ for all $i \in [N]$. Therefore, the balance $\rho_i$ is maximized when the representatives $\calN_{\calR}(i)$ of node $v_i$ are split across clusters $\calC_1, \dots, \calC_K$ in proportion of the sizes of these clusters. Our constraint, formally defined below, requires this proportionality condition to be satisfied for each node $v_i \in \calV$.

\begin{definition}[Representation constraint]
    \label{def:representation_constraint}
    Given a representation graph $\calR$, clusters $\calC_1, \dots, \calC_K$ in $\calG$ satisfy the representation constraint if $\abs{\calC_k \cap \calN_{\calR}(i)} \propto \abs{\calC_k}$, for all $i \in [N]$ and $k \in [K]$, or equivalently,
    \begin{equation}
        \label{eq:representation_constraint}
        \frac{\abs{\calC_k \cap \calN_{\calR}(i)}}{\abs{\calC_k}} = \frac{\abs{\calN_{\calR}(i)}}{N}, \;\; \forall k \in [K], \; \forall i \in [N].
    \end{equation}
\end{definition}

In other words, the representation constraint requires the representatives of any given node $v_i$ to have a proportional membership in all clusters. For example, if a node $v_i$ is connected to $30\%$ of all the nodes in $\calR$, then the clusters discovered in $\calG$ must be such that this node has $30\%$ representation in all clusters. We wish to re-emphasize that the representation constraint applies at the level of individual nodes unlike the constraint in \citep{ChierichettiEtAl:2017:FairClusteringThroughFairlets} that applies at the level of protected groups. In Sections \ref{section:unnormalized_repsc} and \ref{section:normalized_repsc}, we show that \eqref{eq:representation_constraint} can be integrated with the optimization problem solved by spectral clustering. 

%Appendix \ref{appendix:constraint} expands on the two example applications of this constraint that were mentioned in Section \ref{section:introduction}. It also 
Appendix \ref{appendix:constraint} presents additional remarks about the properties of this constraint, in particular, its relation to the statistical-level constraint for categorical attributes \citep{ChierichettiEtAl:2017:FairClusteringThroughFairlets, KleindessnerEtAl:2019:GuaranteesForSpectralClusteringWithFairnessConstraints}. It shows that \eqref{eq:representation_constraint} recovers the statistical-level constraint for a particular configuration of the representation graph, and generalizes it to individual-level constraint for other configurations. Next, we turn to the feasibility of the constraint.

\paragraph*{Feasibility} The optimization problem in \eqref{eq:general_optimization_problem} can always be solved for a small enough value of $\alpha$ (with the convention that $0/0 = 1$). On the other hand, the constraint in Definition \ref{def:representation_constraint} may not always be feasible. For example, if a node has only two representatives, i.e., $\abs{\calN_{\calR}(i)} = 2$, and there are $K > 2$ clusters, then \ref{eq:representation_constraint} can never be satisfied as there will always be at least one cluster $\calC_k$ for which $\abs{\calC_k \cap \calN_{\calR}(i)} = 0$. We argue in the subsequent subsections that spectral relaxation ensures the approximate satisfiability of the constraint when it is added to spectral clustering's optimization problem. We also describe a necessary assumption that is needed to ensure feasibility in practice and two additional assumptions that are required for our theoretical analysis. Thus, the way we use the proposed constraint makes it widely applicable, even though it is very strong. The general form of the constraint in \eqref{eq:general_optimization_problem} may be of independent interest in the context of other clustering algorithms, but we do not pursue this direction here.

%%%%%%%%%%%%%%%%%%%%%%%%%%%%%%%%%%%%%%%%%%%%%%

\subsection{Unnormalized representation-aware spectral clustering (\textsc{URepSC})}
\label{section:unnormalized_repsc}

%This and the following subsection describe our proposed algorithms. 
%We begin with the unnormalized variant \textsc{URepSC}. 
Recall from Section \ref{section:unnormalized_spectral_clustering} that unnormalized spectral clustering approximately minimizes the ratio-cut objective by relaxing an NP-hard optimization problem to solve  \eqref{eq:opt_problem_normal}. The lemma below specifies a sufficient condition that implies the constraint in \eqref{eq:representation_constraint}. 
% See Appendix \ref{appendix:proof_of_technical_lemmas_from_algorithms} for its proof.

\begin{lemma}
    \label{lemma:constraint_matrix_unnorm}
    Let $\bfH \in \bbR^{N \times K}$ have the form specified in \eqref{eq:H_def}. The condition 
    \begin{equation}
      \label{eq:matrix_fairness_criteria}
      \bfR \left( \bfI - \frac{1}{N}\bmone\bmone^\intercal \right) \bfH = \mathbf{0}
    \end{equation}
    implies that the corresponding clusters $\calC_1, \dots, \calC_K$ satisfy the constraint in \eqref{eq:representation_constraint}. Here, $\bfI$ is the $N \times N$ identity matrix and $\bmone$ is a $N$-dimensional all-ones vector.
\end{lemma}

Ideally, we would like to solve the following optimization problem to get the clusters that satisfy the representation constraint,
\begin{equation}
    \label{eq:optimization_problem_ideal}
    \min_{\bfH} \;\;\;\; \trace{\bfH^\intercal \bfL \bfH} \;\;\;\; \text{s.t.} \;\;\;\;\bfH \text{ is of the form \eqref{eq:H_def}} ; \;\;\;\;
    \bfR \left( \bfI - \frac{1}{N}\bmone\bmone^\intercal \right) \bfH = \mathbf{0},
\end{equation}
where $\bfL$ is the unnormalized graph Laplacian defined in \eqref{eq:L_def}. However, as noted in Section \ref{section:unnormalized_spectral_clustering}, the first constraint on $\bfH$ makes this problem NP-hard. Thus, we solve the following relaxed problem,
\begin{equation}
    \label{eq:opt_problem_with_eq_constraint}
    \min_{\bfH} \;\;\;\; \trace{\bfH^\intercal \bfL \bfH} \;\;\;\; \text{s.t.} \;\;\;\; \bfH^\intercal \bfH = \bfI; \;\;\;\; \bfR \left( \bfI - \frac{1}{N}\bmone\bmone^\intercal \right)\bfH = \mathbf{0}.
\end{equation}
Clearly, the columns of any feasible $\bfH$ must belong to the null space of $\bfR(\bfI - \bmone\bmone^\intercal / N)$. Thus, any feasible $\bfH$ can be expressed as $\bfH = \bfY \bfZ$ for some matrix $\bfZ \in \bbR^{N - r \times K}$, where $\bfY \in \bbR^{N \times N - r}$ is an orthonormal matrix containing the basis vectors of the null space of $\bfR(\bfI - \bmone\bmone^\intercal / N)$ as its columns. Here, $r$ is the rank of $\bfR(\bfI - \bmone\bmone^\intercal / N)$. Because $\bfY^\intercal \bfY = \bfI$, $\bfH^\intercal \bfH = \bfZ^\intercal \bfY^\intercal \bfY \bfZ = \bfZ^\intercal \bfZ$. Thus, $\bfH^\intercal \bfH = \bfI \Leftrightarrow \bfZ^\intercal \bfZ = \bfI$. The following optimization problem is equivalent to \eqref{eq:opt_problem_with_eq_constraint} by setting $\bfH = \bfY \bfZ$.
\begin{equation}
    \label{eq:optimization_problem}
    \min_{\bfZ} \;\;\;\; \trace{\bfZ^\intercal \bfY^\intercal \bfL \bfY \bfZ} \;\;\;\; \text{s.t.} \;\;\;\; \bfZ^\intercal \bfZ = \bfI.
\end{equation}
As in standard spectral clustering, the solution to \eqref{eq:optimization_problem} is given by the $K$ leading eigenvectors of $\bfY^\intercal \bfL \bfY$. Of course, for $K$ eigenvectors to exist, $N - r$ must be at least $K$, as $\bfY^\intercal \bfL \bfY$ has dimensions $N - r \times N - r$. The clusters can then be recovered by using $k$-means clustering to cluster the rows of $\bfH = \bfY \bfZ$, as in Algorithm \ref{alg:unnormalized_spectral_clustering}. Algorithm \ref{alg:urepsc} summarizes this procedure. We refer to this algorithm as unnormalized representation-aware spectral clustering (\textsc{URepSC}).

\begin{algorithm}[t]
    \begin{algorithmic}[1]
        \State \textbf{Input: }Adjacency matrix $\bfA$, representation graph $\bfR$, number of clusters $K \geq 2$
        \State Compute $\bfY$ containing orthonormal basis vectors of $\nullspace{\bfR(\bfI - \frac{1}{N}\bmone\bmone^\intercal)}$
        \State Compute Laplacian $\bfL = \bfD - \bfA$
        \State Compute leading $K$ eigenvectors of $\bfY^\intercal \bfL \bfY$. Let $\bfZ$ contain these vectors as its columns.
        \State Apply $k$-means clustering to rows of $\bfH = \bfY \bfZ$ to get clusters $\hat{\calC}_1, \hat{\calC}_2, \dots, \hat{\calC}_K$
        \State \textbf{Return:} Clusters $\hat{\calC}_1, \hat{\calC}_2, \dots, \hat{\calC}_K$
    \end{algorithmic}
    \caption{\textsc{URepSC}}
    \label{alg:urepsc}
\end{algorithm}

%%%%%%%%%%%%%%%%%%%%%%%%%%%%%%%%%%%%%%%%%%%%%%

\subsection{Normalized representation-aware spectral clustering (\textsc{NRepSC})}
\label{section:normalized_repsc}

We use a similar strategy as the previous section to develop the normalized variant of \textsc{RepSC}. Recall from Section \ref{section:normalized_spectral_clustering} that normalized spectral clustering approximately minimizes the $\mathrm{NCut}$ objective. The lemma below is a counterpart of Lemma \ref{lemma:constraint_matrix_unnorm}. It formulates a sufficient condition that implies our constraint in \eqref{eq:representation_constraint}, but this time in terms of the matrix $\bfT$ defined in \eqref{eq:T_def}. 
% See Appendix \ref{appendix:proof_of_technical_lemmas_from_algorithms} for its proof.

\begin{lemma}
    \label{lemma:constraint_matrix_norm}
    Let $\bfT \in \bbR^{N \times K}$ have the form specified in \eqref{eq:T_def}. The condition 
    \begin{equation}
      \label{eq:normalized_matrix_fairness_criteria}
      \bfR \left( \bfI - \frac{1}{N}\bmone\bmone^\intercal \right) \bfT = \mathbf{0}
    \end{equation}
    implies that the corresponding clusters $\calC_1, \dots, \calC_K$ satisfy \eqref{eq:representation_constraint}. Here, $\bfI$ is the $N \times N$ identity matrix and $\bmone$ is a $N$ dimensional all-ones vector.
\end{lemma}

For \textsc{NRepSC}, we assume that the similarity graph $\calG$ is connected so that the diagonal entries of $\bfD$ are strictly positive. We proceed as before to incorporate constraint \eqref{eq:normalized_matrix_fairness_criteria} in optimization problem \eqref{eq:normalized_ideal_opt_problem}. After applying the spectral relaxation, we get
\begin{equation}
    \label{eq:optimization_problem_normalized}
    \min_{\bfT} \;\;\;\; \trace{\bfT^\intercal  \bfL \bfT} \;\;\;\; \text{s.t.} \;\;\;\; \bfT^\intercal \bfD \bfT = \bfI; \;\;\;\; \bfR(\bfI - \bmone \bmone^\intercal / N) \bfT = \mathbf{0}.
\end{equation}
As before, $\bfT = \bfY \bfZ$ for some $\bfZ \in \bbR^{N - r \times K}$, where recall that columns of $\bfY$ contain orthonormal basis for $\nullspace{\bfR(\bfI - \bmone \bmone^\intercal / N)}$. This reparameterization yields
\begin{equation*}
    \min_{\bfZ} \;\;\;\; \trace{\bfZ^\intercal \bfY^\intercal  \bfL \bfY \bfZ} \;\;\;\; \text{s.t.} \;\;\;\; \bfZ^\intercal \bfY^\intercal \bfD \bfY \bfZ = \bfI.
\end{equation*}
Define $\bfQ \in \bbR^{N - r \times N - r}$ such that $\bfQ^2 = \bfY^\intercal \bfD \bfY$. Note that $\bfQ$ exists as the entries of $\bfD$ are non-negative.%\footnote{Let $\bfP = \bfY^\intercal \bfD \bfY$. As $\bfP$ is symmetric, it can be written as $\bfP = \bfU \bmSigma \bfU^\intercal$. Then, $\bfQ$ is defined as $\bfQ = \sqrt{\bfP} = \bfU \sqrt{\bmSigma} \bfU^\intercal$, where the square root is applied to the entries of $\bmSigma$ element by element.}.
Let $\bfV = \bfQ \bfZ$. Then, $\bfZ = \bfQ^{-1} \bfV$ and $\bfZ^\intercal \bfQ^2 \bfZ = \bfV^\intercal \bfV$ as $\bfQ$ is symmetric. Reparameterizing again, we get:
\begin{equation*}
    \min_{\bfV} \;\;\;\; \trace{\bfV^\intercal \bfQ^{-1} \bfY^\intercal  \bfL \bfY \bfQ^{-1} \bfV} \;\;\;\; \text{s.t.} \;\;\;\; \bfV^\intercal \bfV = \bfI.
\end{equation*}
This again is the standard form of the trace minimization problem and the optimal solution is given by the leading $K$ eigenvectors of $\bfQ^{-1} \bfY^\intercal  \bfL \bfY \bfQ^{-1}$. Algorithm \ref{alg:nrepsc} summarizes the normalized representation-aware spectral clustering algorithm, which we denote by \textsc{NRepSC}. Note that the algorithm assumes that $\bfQ$ is invertible, which requires the absence of isolated nodes in the similarity graph $\calG$. 

\begin{algorithm}[t]
    \begin{algorithmic}[1]
        \State \textbf{Input: }Adjacency matrix $\bfA$, representation graph $\bfR$, number of clusters $K \geq 2$
        \State Compute $\bfY$ containing orthonormal basis vectors of $\nullspace{\bfR(\bfI - \frac{1}{N}\bmone\bmone^\intercal)}$
        \State Compute Laplacian $\bfL = \bfD - \bfA$
        \State Compute $\bfQ = \sqrt{\bfY^\intercal \bfD \bfY}$ using the matrix square root
        \State Compute leading $K$ eigenvectors of $\bfQ^{-1} \bfY^\intercal  \bfL \bfY \bfQ^{-1}$. Set them as columns of $\bfV \in \bbR^{N-r \times K}$
        \State Apply $k$-means clustering to the rows of $\bfT = \bfY \bfQ^{-1} \bfV$ to get clusters $\hat{\calC}_1, \hat{\calC}_2, \dots, \hat{\calC}_K$
        \State \textbf{Return:} Clusters $\hat{\calC}_1, \hat{\calC}_2, \dots, \hat{\calC}_K$
    \end{algorithmic}
    \caption{\textsc{NRepSC}}
    \label{alg:nrepsc}
\end{algorithm}

%%%%%%%%%%%%%%%%%%%%%%%%%%%%%%%%%%%%%%%%%%%%%%

\subsection{Comments on the proposed algorithms}
\label{section:comments_on_the_proposed_algorithms}

Before proceeding with the theoretical analysis in Section \ref{section:analysis}, we first make two remarks about the proposed algorithms.

\paragraph*{Spectral relaxation} Note that the constraints $\bfR(\bfI - \bmone\bmone^\intercal / N) \bfH = \mathbf{0}$ and $\bfR(\bfI - \bmone\bmone^\intercal / N) \bfT = \mathbf{0}$ imply the satisfaction of our representation constraint only when $\bfH$ and $\bfT$ have the form given in \eqref{eq:H_def} and \eqref{eq:T_def}, respectively. Thus, a feasible solution to the relaxed optimization problem in~\eqref{eq:opt_problem_with_eq_constraint} or \eqref{eq:optimization_problem_normalized} may not necessarily result in \textit{representation-aware} clusters. In fact, even in the unconstrained case, there are no general guarantees that bound the difference between the optimal solution of~\eqref{eq:opt_problem_normal} or \eqref{eq:opt_problem_normal_normalized} and the respective optimal solutions of the original NP-hard ratio-cut/normalized-cut problems \citep{KleindessnerEtAl:2019:GuaranteesForSpectralClusteringWithFairnessConstraints}. Thus, the representation-aware nature of the clusters discovered by solving \eqref{eq:optimization_problem} or \eqref{eq:optimization_problem_normalized} cannot be guaranteed in the general case. Nonetheless, we show in Section~\ref{section:analysis} that the discovered clusters indeed satisfy the representation constraint under certain additional assumptions.

\paragraph*{Computational complexity} Algorithms \ref{alg:urepsc} and \ref{alg:nrepsc} have a time complexity of $O(N^3)$ and space complexity of $O(N^2)$. Finding the null space of $\bfR(\bfI - \bmone \bmone^\intercal / N)$ to calculate $\bfY$ and computing the eigenvectors of appropriate matrices are the computationally dominant steps in both cases. This matches the worst-case complexity of the standard spectral clustering algorithm. For small $K$, several approximations can reduce this complexity, but most such techniques require $K = 2$ \citep{YuShi:2004:SegmentationGivenPartialGroupingConstraints,XuEtAl:2009:FastNormalizedCutWithLinearConstraints}.

%%%%%%%%%%%%%%%%%%%%%%%%%%%%%%%%%%%%%%%%%%%%%%

\section{Analysis}
\label{section:analysis}

In this section, we show that Algorithms~\ref{alg:urepsc} and \ref{alg:nrepsc} recover the ground truth clusters with a high probability under certain assumptions on the representation graph. As we will see in Section \ref{section:consistency_results}, the ground truth clusters satisfy \eqref{eq:representation_constraint} by construction for similarity graphs $\calG$ sampled from a modified variant of the Stochastic Block Model (SBM) \citep{HollandEtAl:1983:StochasticBlockmodelsFirstSteps}, described in Section \ref{section:rsbm}. Section \ref{section:consistency_results} presents our main results that establish a high probability upper bound on the number of mistakes made by the proposed algorithms. Corollaries \ref{corollary:weak_consistency_urepsc} and \ref{corollary:weak_consistency_nrepsc} then establish their weak-consistency.

%%%%%%%%%%%%%%%%%%%%%%%%%%%%%%%%%%%%%%%%%%%%%%

\subsection{$\calR$-SBM}
\label{section:rsbm}

The well known Stochastic Block Model (SBM) \citep{HollandEtAl:1983:StochasticBlockmodelsFirstSteps} allows one to sample random graphs with known clusters. It takes a function $\pi: \calV \rightarrow [K]$ as input. This function assigns each node $v_i \in \calV$ to one of the $K$ clusters. Then, independently for all node pairs $(v_i, v_j)$ such that $i > j$, $\rmP(A_{ij} = 1) = B_{\pi(v_i) \pi(v_j)},$ where $\bfB \in [0, 1]^{K \times K}$ is a symmetric matrix. The $(k, \ell)^{th}$ entry of $\bfB$ specifies the probability of a connection between two nodes that belong to clusters $\calC_k$ and $\calC_\ell$, respectively. A commonly used variant of SBM assumes $B_{kk} = \alpha$ and $B_{k\ell} = \beta$ for all $k, \ell \in [K]$ such that $k \neq \ell$. We define a variant of SBM with respect to a representation graph $\calR$ below and refer to it as Representation-Aware SBM or $\calR$-SBM.

\begin{definition}[$\calR$-SBM]
    \label{def:conditioned_sbm}
    A $\calR$-SBM is defined by the tuple $(\pi, \calR, p, q, r, s)$, where $\pi: \calV \rightarrow [K]$ maps nodes in $\calV$ to clusters, $\calR$ is a representation graph, and $1 \geq p \geq q \geq r \geq s \geq 0$ are probabilities used for sampling edges. Under this model, for all $i > j$,
    \begin{equation}
        \label{eq:sbm_specification}
        \rmP(A_{ij} = 1) = \begin{cases}
          p & \text{if } \pi(v_i) = \pi(v_j) \text{ and } R_{ij} = 1, \\
          q & \text{if } \pi(v_i) \neq \pi(v_j) \text{ and } R_{ij} = 1, \\
          r & \text{if } \pi(v_i) = \pi(v_j) \text{ and } R_{ij} = 0,  \\
          s & \text{if } \pi(v_i) \neq \pi(v_j) \text{ and } R_{ij} = 0.
        \end{cases}
      \end{equation}
\end{definition}

The similarity graphs sampled from a $\calR$-SBM have two interesting properties. First, everything else being equal, nodes have a higher tendency to connect with other nodes in the same cluster as $p \geq q$ and $r \geq s$. Thus, $\calR$-SBM plants the clusters specified by $\pi$ in the sampled graph $\calG$. Second, and more importantly, $\calR$-SBM also plants the properties of the given representation graph $\calR$ in the sampled graphs $\calG$. To see this, note that nodes that are connected in $\calR$ have a higher probability of being connected in $\calG$ as well ($p \geq r$ and $q \geq s$).

Recall that our algorithms must discover clusters in $\calG$ in which the connected nodes in $\calR$ are proportionally distributed. However, $\calR$-SBM makes two nodes connected in $\calR$ more likely to connect in $\calG$, even if they do not belong to the same cluster ($q \geq r$). In this sense, graphs sampled from $\calR$-SBM are ``hard'' instances from the perspective of our algorithms. When $\calR$ itself has a community structure, there are two natural ways to cluster the nodes: \textbf{(i)} based on the ground-truth clusters $\calC_1$, $\calC_2$, \dots, $\calC_K$ specified by $\pi$; and \textbf{(ii)} based on the communities in $\calR$. The clusters based on communities in $\calR$ are likely to not satisfy the representation constraint in Definition \ref{def:representation_constraint} as tightly connected nodes in $\calR$ will be assigned to the same cluster in this case rather than being distributed across clusters. 

We show in Section \ref{section:consistency_results} that, under certain assumptions on $\calR$, the ground-truth clusters can be constructed so that they satisfy the representation constraint \eqref{eq:representation_constraint}. Assuming that the ground-truth clusters indeed satisfy \eqref{eq:representation_constraint}, the goal is to show that Algorithms \ref{alg:urepsc} and \ref{alg:nrepsc} recover the ground-truth clusters with high probability rather than returning any other natural but ``representation-unaware'' clusters.

%%%%%%%%%%%%%%%%%%%%%%%%%%%%%%%%%%%%%%%%%%%%%%

\subsection{Consistency results}
\label{section:consistency_results}

As noted in Section \ref{section:constraint}, some representation graphs lead to constraints that cannot be satisfied. For our theoretical analysis, we restrict our focus to cases where the constraint in \eqref{eq:representation_constraint} is feasible. Towards this end, an additional assumption on $\calR$ is required.

\begin{assumption}
    \label{assumption:R_is_d_regular}
    $\calR$ is a $d$-regular graph for some $K \leq d \leq N$. Moreover, $R_{ii} = 1$ for all $i \in [N]$, and each node in $\calR$ is connected to $d / K$ nodes from cluster $\calC_i$, for all $i \in [K]$ (including the self-loop).
\end{assumption}

%Recall that the function $\pi: \calV \rightarrow [K]$ assigns each node to a cluster in $\calR$-SBM.
Assumption~\ref{assumption:R_is_d_regular} ensures the existence of a $\pi$ for which the corresponding ground-truth clusters satisfy the representation constraint in \eqref{eq:representation_constraint}. Namely, assuming equal-sized clusters, let $\pi(v_i) = k$, if $(k - 1) \frac{N}{K} \leq i \leq k \frac{N}{K}$ for all $i \in [N]$, and $k \in [K]$. It can be easily verified that the resulting clusters $\calC_k = \{v_i : \pi(v_i) = k \}$, $k \in [K]$ satisfy \eqref{eq:representation_constraint}.

Before presenting our main results, we need to set up additional notation. Let $\bmTheta \in \{0, 1\}^{N \times K}$ indicate the ground-truth cluster memberships, i.e., $\Theta_{ij} = 1 \Leftrightarrow v_i \in \calC_j$. Similarly, $\hat{\bmTheta} \in \{0, 1\}^{N \times K}$ indicates the clusters returned by the algorithm, i.e., $\hat{\Theta}_{ij} = 1 \Leftrightarrow v_i \in \hat{\calC}_j$. Further, let $\calJ$ be the set of all $K \times K$ permutation matrices. The fraction of misclustered nodes \citep{LeiEtAl:2015:ConsistencyOfSpectralClusteringInSBM} is defined as 
\begin{equation*}
    M(\bmTheta, \hat{\bmTheta}) = \min_{\bfJ \in \calJ} \frac{1}{N} \norm{\bmTheta - \hat{\bmTheta} \bfJ}[0].
\end{equation*}
As the ground truth clusters $\calC_1, \dots, \calC_K$ satisfy \eqref{eq:representation_constraint} by construction, a low $M(\bmTheta, \hat{\bmTheta})$ indicates that the clusters returned by the algorithm approximately satisfy \eqref{eq:representation_constraint}. Theorems \ref{theorem:consistency_result_unnormalized} and \ref{theorem:consistency_result_normalized} also use the eigenvalues of the Laplacian matrix in the expected case. We use $\calL$ to denote this matrix, and define it as $\calL = \calD - \calA$, where $\calA = \rmE[\bfA]$ is the expected adjacency matrix of a graph sampled from $\calR$-SBM and $\calD \in \bbR^{N \times N}$ is a diagonal matrix such that $\calD_{ii} = \sum_{j = 1}^N \calA_{ij}$, for all $i \in [N]$. The next two results establish high-probability upper bounds on the fraction of misclustered nodes for \textsc{URepSC} and \textsc{NRepSC} for similarity graphs $\calG$ sampled from $\calR$-SBM.

\begin{theorem}[Error bound for \textsc{URepSC}]
    \label{theorem:consistency_result_unnormalized}
    Let $\rank{\bfR} \leq N - K$ and assume that all clusters have equal sizes. Let $\mu_1 \leq \mu_2 \leq \dots \leq \mu_{N - r}$ denote the eigenvalues of $\bfY^\intercal \calL \bfY$, where $\bfY$ was defined in Section \ref{section:unnormalized_repsc}. Define $\gamma = \mu_{K + 1} - \mu_{K}$. Under Assumption \ref{assumption:R_is_d_regular}, there exists a universal constant $\const(C, \alpha)$, such that if $\gamma$ satisfies $$\gamma^2 \geq \const(C, \alpha) (2 + \epsilon) p N K \ln N,$$ and $p \geq C \ln N / N$ for some $C > 0$, then,
    $$M(\bmTheta, \hat{\bmTheta}) \leq \const(C, \alpha) \frac{(2 + \epsilon)}{\gamma^2} p N \ln N,$$
    for every $\epsilon > 0$ with probability at least $1 - 2 N^{-\alpha}$ when a $(1 + \epsilon)$-approximate algorithm for $k$-means clustering is used in Step 5 of Algorithm \ref{alg:urepsc}.
\end{theorem}

\begin{theorem}[Error bound for \textsc{NRepSC}]
    \label{theorem:consistency_result_normalized}
    Let $\rank{\bfR} \leq N - K$ and assume that all clusters have equal sizes. Let $\mu_1 \leq \mu_2 \leq \dots \leq \mu_{N - r}$ denote the eigenvalues of $\calQ^{-1} \bfY^\intercal \calL \bfY \calQ^{-1}$, where $\calQ = \sqrt{\bfY^\intercal \calD \bfY}$ and $\bfY$ was defined in Section \ref{section:unnormalized_repsc}. Define $\gamma = \mu_{K + 1} - \mu_{K}$ and $\lambda_1 = qd + s(N - d) + (p - q) \frac{d}{K} + (r - s) \frac{N - d}{K}$. Under Assumption \ref{assumption:R_is_d_regular}, there are universal constants $\const_1(C, \alpha)$, $\const_4(C, \alpha)$, and $\const_5(C, \alpha)$ such that if:
    \begin{enumerate}
        \item $\left(\frac{\sqrt{p N \ln N}}{\lambda_1 - p}\right) \left(\frac{\sqrt{p N \ln N}}{\lambda_1 - p} + \frac{1}{6\sqrt{C}}\right) \leq \frac{1}{16(\alpha + 1)}$,
        \item $\frac{\sqrt{p N \ln N}}{\lambda_1 - p} \leq \const_4(C, \alpha)$, and
        \item $16(2 + \epsilon)\left[ \frac{8 \const_5(C, \alpha) \sqrt{K}}{\gamma} + \const_1(C, \alpha)\right]^2 \frac{p N^2 \ln N}{(\lambda_1 - p)^2} < \frac{N}{K}$,
    \end{enumerate}
    and $p \geq C \ln N / N$ for some $C > 0$, then,
    $$M(\bmTheta, \hat{\bmTheta}) \leq 32(2 + \epsilon)\left[ \frac{8 \const_5(C, \alpha) \sqrt{K}}{\gamma} + \const_1(C, \alpha)\right]^2 \frac{p N \ln N}{(\lambda_1 - p)^2},$$
    for every $\epsilon > 0$ with probability at least $1 - 2 N^{-\alpha}$ when a $(1 + \epsilon)$-approximate algorithm for $k$-means clustering is used in Step 6 of Algorithm \ref{alg:nrepsc}.
\end{theorem}

Next, we discuss our assumptions and use the error bounds above to establish the weak consistency of our algorithms.

\subsection{Discussion}
\label{section:discussion}

Note that $\bfI - \bmone \bmone^\intercal / N$ is a projection matrix and $\bmone$ is its eigenvector with eigenvalue $0$. Any vector orthogonal to $\bmone$ is also an eigenvector with eigenvalue $1$. Thus, $\rank{\bfI - \bmone \bmone^\intercal / N} = N - 1$. Because $\rank{\bfR (\bfI - \bmone \bmone^\intercal / N)} \leq \min(\rank{\bfR}, \rank{\bfI - \bmone \bmone^\intercal / N})$, requiring $\rank{\bfR} \leq N - K$ ensures that $\rank{\bfR(\bfI - \bmone \bmone^\intercal / N)} \leq N - K$, which is necessary for \eqref{eq:optimization_problem} and \eqref{eq:optimization_problem_normalized} to have a solution.

The assumption on the size of the clusters, together with the $d$-regularity assumption on $\calR$, allows us to compute the smallest $K$ eigenvalues of the Laplacian matrix in the expected case. This is a crucial step in the proof of our main consistency results. The additional assumptions in Theorem \ref{theorem:consistency_result_normalized} are easy to satisfy as $\lambda_1$ scales linearly with $N$ for appropriate values of $p, q, r,$ and $s$. Similar assumptions were also used in \citet{KleindessnerEtAl:2019:GuaranteesForSpectralClusteringWithFairnessConstraints}.

\begin{remark}
    In practice, Algorithms \ref{alg:urepsc} and \ref{alg:nrepsc} only require the rank assumption on $\bfR$ to ensure the existence of solutions to the corresponding optimization problems. The assumptions on the size of clusters and $d$-regularity of $\calR$ are only needed for our theoretical analysis.
\end{remark}

The next two corollaries are direct consequences of Theorems \ref{theorem:consistency_result_unnormalized} and \ref{theorem:consistency_result_normalized}, respectively.

\begin{corollary}[Weak consistency of \textsc{URepSC}]
    \label{corollary:weak_consistency_urepsc}
    Under the same setup as Theorem \ref{theorem:consistency_result_unnormalized}, for \textsc{URepSC}, $M(\bmTheta, \hat{\bmTheta}) = o(1)$ with probability $1 - o(1)$ if $\gamma = \omega(\sqrt{pN K \ln N})$.
\end{corollary}

\begin{corollary}[Weak consistency of \textsc{NRepSC}]
    \label{corollary:weak_consistency_nrepsc}
    Under the same setup as Theorem \ref{theorem:consistency_result_normalized}, for \textsc{NRepSC}, $M(\bmTheta, \hat{\bmTheta}) = o(1)$ with probability $1 - o(1)$ if $\gamma = \omega(\sqrt{pN K \ln N} / (\lambda_1 - p))$.
\end{corollary}

Thus, under the assumptions in the corollaries above, Algorithms \ref{alg:urepsc} and \ref{alg:nrepsc} are weakly consistent \citep{Abbe:2018:CommunityDetectionAndStochasticBlockModels}. The conditions on $\gamma$ are satisfied in many interesting cases. For example, when there are $P$ protected groups, as is the case for statistical-level constraints, the equivalent representation graph has $P$ cliques that are not connected to each other (see Appendix \ref{appendix:constraint}). \citet{KleindessnerEtAl:2019:GuaranteesForSpectralClusteringWithFairnessConstraints} show that $\gamma = \theta(N/K)$ in this case (for the unnormalized variant), which satisfies the criterion given above if $K$ is not too large.

Finally, Theorems \ref{theorem:consistency_result_unnormalized} and \ref{theorem:consistency_result_normalized} require a $(1 + \epsilon)$-approximate solution to $k$-means clustering. Several efficient algorithms have been proposed in the literature for this task \citep{KumarEtAl:2004:ASimpleLinearTimeApproximateAlgorithmForKMeansClusteringInAnyDimension,ArthurVassilvitskii:2007:KMeansTheAdvantagesOfCarefulSeeding,AhmadianEtAl:2017:BetterGuaranteesForKMeansAndEuclideanKMedianByPrimalDualAlgorithms}. Such algorithms are also available in commonly used software packages like MATLAB and scikit-learn. The assumption that $p \geq C \ln N / N$ controls the sparsity of the graph, and is required in the the consistency proofs for standard spectral clustering as well \citep{LeiEtAl:2015:ConsistencyOfSpectralClusteringInSBM}.

%%%%%%%%%%%%%%%%%%%%%%%%%%%%%%%%%%%%%%%%%%%%%%

\subsection{Proof of Theorems \ref{theorem:consistency_result_unnormalized} and \ref{theorem:consistency_result_normalized}}
\label{section:proof_of_theorems}

The proof of Theorems \ref{theorem:consistency_result_unnormalized} and \ref{theorem:consistency_result_normalized} follow the commonly used template for such results \citep{RoheEtAl:2011:SpectralClusteringAndTheHighDimensionalSBM, LeiEtAl:2015:ConsistencyOfSpectralClusteringInSBM}. In the context of \textsc{URepSC} (similar arguments work for \textsc{NRepSC} as well), we
\begin{enumerate}
    \item Compute the expected Laplacian matrix $\calL$ under $\calR$-SBM and show that its top $K$ eigenvectors can be used to recover the ground-truth clusters (Lemmas \ref{lemma:introducing_uks}--\ref{lemma:orthonormal_eigenvectors_y2_yK}).
    \item Show that these top $K$ eigenvectors lie in the null space of $\bfR(\bfI - \bmone \bmone^\intercal / N)$, and hence are also the top $K$ eigenvectors of $\bfY^\intercal \calL \bfY$ (Lemma \ref{lemma:first_K_eigenvectors_of_L}). This implies that Algorithm \ref{alg:urepsc} returns the ground truth clusters in the expected case.
    \item Use matrix perturbation arguments to establish a high probability mistake bound in the general case when the graph $\calG$ is sampled from a $\calR$-SBM (Lemmas \ref{lemma:bound_on_D-calD}--\ref{lemma:k_means_error}).
\end{enumerate}

We begin with a series of lemmas that highlight certain useful properties of eigenvalues and eigenvectors of the expected Laplacian $\calL$. These lemmas will be used in Sections \ref{section:proof_consistency_unnormalized} and \ref{section:proof_consistency_normalized} to prove Theorem \ref{theorem:consistency_result_unnormalized} and \ref{theorem:consistency_result_normalized}, respectively. See the supplementary material \citep{ThisPaperSupp} for the proofs of all technical lemmas. For the remainder of this section, we assume that all appropriate assumptions made in Theorems \ref{theorem:consistency_result_unnormalized} and \ref{theorem:consistency_result_normalized} are satisfied.

The first lemma shows that certain vectors that can be used to recover the ground-truth clusters indeed satisfy the representation constraint in \eqref{eq:matrix_fairness_criteria} and \eqref{eq:normalized_matrix_fairness_criteria}.

\begin{lemma}
\label{lemma:introducing_uks}
The $N$-dimensional vector of all ones, denoted by $\bmone$, is an eigenvector of $\bfR$ with eigenvalue $d$. Define $\bfu_k \in \bbR^N$ for $k \in [K - 1]$ as,
\begin{equation*}
    u_{ki} = \begin{cases}
    1 & \text{ if }v_i \in \calC_k \\
    -\frac{1}{K - 1} & \text{ otherwise,}
    \end{cases}
\end{equation*}
where $u_{ki}$ is the $i^{th}$ element of $\bfu_k$. Then, $\bmone, \bfu_1, \dots, \bfu_{K - 1} \in \nullspace{\bfR(\bfI - \frac{1}{N}\bmone\bmone^\intercal)}$. Moreover, $\bmone, \bfu_1, \dots, \bfu_{K - 1}$ are linearly independent.
\end{lemma}

Recall that we use $\calA \in \bbR^{N \times N}$ to denote the expected adjacency matrix of the similarity graph $\calG$. Clearly, $\calA = \tilde{\calA} - p \bfI$, where $\tilde{\calA}$ is such that $\tilde{\calA}_{ij} = P(A_{ij} = 1)$ if $i \neq j$ (see \eqref{eq:sbm_specification}) and $\tilde{\calA}_{ii} = p$ otherwise. Note that 
\begin{equation}
    \label{eq:eigenvector_A_tildeA}
    \tilde{\calA} \bfx = \lambda \bfx \,\,\,\, \Leftrightarrow \,\,\,\, \calA \bfx = (\lambda - p) \bfx.
\end{equation}
Simple algebra shows that $\tilde{\calA}$ can be written as
\begin{equation}
    \label{eq:tilde_cal_A_def}
    \tilde{\calA} = q \bfR + s (\bmone \bmone^\intercal - \bfR) + (p - q)\sum_{k = 1}^K \bfG_k \bfR \bfG_k + (r - s) \sum_{k = 1}^K \bfG_k (\bmone \bmone^\intercal - \bfR) \bfG_k,
\end{equation}
where, for all $k \in [K]$, $\bfG_k \in \bbR^{N \times N}$ is a diagonal matrix such that $(\bfG_k)_{ii} = 1$ if $v_i \in \calC_k$ and $0$ otherwise. The next lemma shows that $\bmone, \bfu_1, \dots, \bfu_{K - 1}$ defined in Lemma \ref{lemma:introducing_uks} are eigenvectors of $\tilde{\calA}$.

\begin{lemma}
    \label{lemma:uk_eigenvector_of_tildeA}
    Let $\bmone, \bfu_1, \dots, \bfu_{K - 1}$ be as defined in Lemma \ref{lemma:introducing_uks}. Then,
    \begin{eqnarray*}
        \tilde{\calA} \bmone &=& \lambda_1 \bmone \text{ where } \lambda_1 = qd + s(N - d) + (p - q) \frac{d}{K} + (r - s) \frac{N - d}{K}, \text{ and } \\
        \tilde{\calA} \bfu_k &=& \lambda_{1 + k} \bfu_k \text{ where } \lambda_{1 + k} = (p - q) \frac{d}{K} + (r - s) \frac{N - d}{K}.
    \end{eqnarray*}
\end{lemma}

Let $\calL = \calD - \calA$ be the expected Laplacian matrix, where $\calD$ is a diagonal matrix with $\calD_{ii} = \sum_{j=1}^N \calA_{ij}$ for all $i \in [N]$. It is easy to see that $\calD_{ii} = \lambda_1 - p$ for all $i \in [N]$ as $\calA \bmone = (\lambda_1 - p) \bmone$ by \eqref{eq:eigenvector_A_tildeA} and Lemma \ref{lemma:uk_eigenvector_of_tildeA}. Thus, $\calD = (\lambda_1 - p) \bfI$ and hence any eigenvector of $\tilde{\calA}$ with eigenvalue $\lambda$ is also an eigenvector of $\calL$ with eigenvalue $\lambda_1 - \lambda$. That is, if $\tilde{\calA} \bfx = \lambda \bfx$,
\begin{equation}
    \label{eq:eigenvectors_of_L}
    \calL \bfx = (\calD - \calA)\bfx = ((\lambda_1 - p) \bfI - (\tilde{\calA} - p \bfI)) \bfx = (\lambda_1 - \lambda) \bfx.
\end{equation}
Hence, the eigenvectors of $\calL$ corresponding to the $K$ smallest eigenvalues are the same as the eigenvectors of $\tilde{\calA}$ corresponding to the $K$ largest eigenvalues.

Recall that the columns of the matrix $\bfY$ used in Algorithms \ref{alg:urepsc} and \ref{alg:nrepsc} contain the orthonormal basis for the null space of $\bfR(\bfI - \bmone \bmone^\intercal/N)$. To solve \eqref{eq:optimization_problem} and \eqref{eq:optimization_problem_normalized}, we only need to optimize over vectors that belong to this null space. By Lemma \ref{lemma:introducing_uks}, $\bmone, \bfu_1, \dots, \bfu_{K - 1} \in \nullspace{\bfR(\bfI - \bmone \bmone^\intercal/N)}$ and these vectors are linearly independent. However, we need an orthonormal basis to compute $\bfY$. Let $\bfy_1 = \bmone / \sqrt{N}$ and $\bfy_2, \dots, \bfy_K$ be orthonormal vectors that span the same space as $\bfu_1, \dots, \bfu_{K - 1}$. The next lemma computes such $\bfy_2, \dots, \bfy_K$. The matrix $\bfY \in \bbR^{N \times N - r}$ contains these vectors $\bfy_1, \dots, \bfy_K$ as its first $K$ columns.

\begin{lemma}
\label{lemma:orthonormal_eigenvectors_y2_yK}
Define $\bfy_{1 + k} \in \bbR^N$ for $k \in [K - 1]$ as
\begin{equation*}
    y_{1 + k, i} = \begin{cases}
    0 & \text{ if } v_i \in \calC_{k{'}} \text{ s.t. } k{'} < k \\
    \frac{K - k}{\sqrt{\frac{N}{K}(K - k)(K - k + 1)}} & \text{ if } v_i \in \calC_k \\
    -\frac{1}{\sqrt{\frac{N}{K}(K - k)(K - k + 1)}} & \text{ otherwise.}
    \end{cases}
\end{equation*}
Then, for all $k \in [K - 1]$, $\bfy_{1 + k}$ are orthonormal vectors that span the same space as $\bfu_1, \bfu_2, \dots, \bfu_{K - 1}$ and $\bfy_1^\intercal \bfy_{1 + k} = 0$. As before, $y_{1 + k, i}$ refers to the $i^{th}$ element of $\bfy_{1 + k}$.
\end{lemma}

Let $\bfX \in \bbR^{N \times K}$ be such that it has $\bfy_1, \dots, \bfy_{K}$ as its columns. If two nodes belong to the same cluster, the rows corresponding to these nodes in $\bfX \bfU$ will be identical for any $\bfU \in \bbR^{K \times K}$ such that $\bfU^\intercal \bfU = \bfU \bfU^\intercal = \bfI$. Thus, any $K$ orthonormal vectors belonging to the span of $\bfy_1, \dots, \bfy_K$ can be used to recover the ground truth clusters.  With the general properties of the eigenvectors and eigenvalues established in the lemmas above, we next move on to the proof of Theorem \ref{theorem:consistency_result_unnormalized} in the next section and Theorem \ref{theorem:consistency_result_normalized} in Section \ref{section:proof_consistency_normalized}.

%%%%%%%%%%%%%%%%%%%%%%%%%%%%%%%%%%%%%%%%%%%%%%

\subsubsection{Proof of Theorem \ref{theorem:consistency_result_unnormalized}}
\label{section:proof_consistency_unnormalized}

Let $\calZ \in \bbR^{N - r \times K}$ be a solution to the optimization problem \eqref{eq:optimization_problem} in the expected case with $\calA$ as input. The next lemma shows that columns of $\bfY \calZ$ indeed lie in the span of $\bfy_1, \dots, \bfy_K$. Thus, the $k$-means clustering step in Algorithm \ref{alg:urepsc} will return the correct ground truth clusters when $\calA$ is passed as input.

\begin{lemma}
    \label{lemma:first_K_eigenvectors_of_L}
    Let $\bfy_1 = \bmone / \sqrt{N}$ and $\bfy_{1 + k}$ be as defined in Lemma \ref{lemma:orthonormal_eigenvectors_y2_yK} for all $k \in [K - 1]$. Further, let $\calZ$ be the optimal solution of the optimization problem in \eqref{eq:optimization_problem} with $\bfL$ set to $\calL$. Then, the columns of $\bfY \calZ$ lie in the span of $\bfy_1, \bfy_2, \dots, \bfy_K$.
\end{lemma}

Next, we use arguments from matrix perturbation theory to show a high-probability bound on the number of mistakes made by the algorithm. In particular, we need an upper bound on $\norm{\bfY^\intercal \bfL \bfY - \bfY^\intercal \calL \bfY}$, where $\bfL$ is the Laplacian matrix for a graph randomly sampled from $\calR$-SBM and $\norm{\bfP} = \sqrt{\lambdamax{\bfP^\intercal \bfP}}$ for any matrix $\bfP$. Note that  $\norm{\bfY} = \norm{\bfY^\intercal} = 1$ as $\bfY^\intercal \bfY = \bfI$. Thus, 
\begin{equation}
    \label{eq:reducing_YLY_to_L}
    \norm{\bfY^\intercal \bfL \bfY - \bfY^\intercal \calL \bfY} \leq \norm{\bfY^\intercal} \,\, \norm{\bfL - \calL} \,\, \norm{\bfY} = \norm{\bfL - \calL}.
\end{equation}
Moreover, $$\norm{\bfL - \calL} = \norm{\bfD - \bfA - (\calD - \calA)} \leq \norm{\bfD - \calD} + \norm{\bfA - \calA}.$$ The next two lemmas bound the two terms on the right hand side of the inequality above, thus providing an upper bound on $\norm{\bfL - \calL}$, and hence on $\norm{\bfY^\intercal \bfL \bfY - \bfY^\intercal \calL \bfY}$ by \eqref{eq:reducing_YLY_to_L}.

\begin{lemma}
    \label{lemma:bound_on_D-calD}
    Assume that $p \geq C \frac{\ln N}{N}$ for some constant $C > 0$. Then, for every $\alpha > 0$, there exists a constant $\const_1(C, \alpha)$ that only depends on $C$ and $\alpha$ such that $$\norm{\bfD - \calD} \leq \const_1(C, \alpha) \sqrt{p N \ln N}$$ with probability at-least $1 - N^{-\alpha}$.
\end{lemma}

\begin{lemma}
    \label{lemma:bound_on_A-calA}
    Assume that $p \geq C \frac{\ln N}{N}$ for some constant $C > 0$. Then, for every $\alpha > 0$, there exists a constant $\const_2(C, \alpha)$ that only depends on $C$ and $\alpha$ such that $$\norm{\bfA - \calA} \leq \const_2(C, \alpha) \sqrt{p N}$$ with probability at-least $1 - N^{-\alpha}$.
\end{lemma}

From Lemmas \ref{lemma:bound_on_D-calD} and \ref{lemma:bound_on_A-calA}, we conclude that there is always a constant $\const_3(C, \alpha) = \max\{\const_1(C, \alpha), \const_2(C, \alpha)\}$ such that, for any $\alpha > 0$, with probability at least $1 - 2N^{-\alpha}$, 
\begin{equation}
    \label{eq:L-calL_bound}
    \norm{\bfY^\intercal \bfL \bfY - \bfY^\intercal \calL \bfY} \leq \norm{\bfL - \calL} \leq \const_3(C, \alpha) \sqrt{p N \ln N}. 
\end{equation}

Let $\calZ$ and $\bfZ$ denote the optimal solution of \eqref{eq:optimization_problem} in the expected ($\bfL$ replaced with $\calL$) and observed case.
We use \eqref{eq:L-calL_bound} to show a bound on $\norm{\bfY \calZ - \bfY Z}[F]$ in Lemma \ref{lemma:bound_on_eigenvector_diff} and then use this bound to argue that Algorithm \ref{alg:urepsc} makes a small number of mistakes when the graph is sampled from $\calR$-SBM.

\begin{lemma}
\label{lemma:bound_on_eigenvector_diff}
    Let $\mu_1 \leq \mu_2 \leq \dots \leq \mu_{N - r}$ be eigenvalues of $\bfY^\intercal \calL \bfY$. Further, let the columns of $\calZ \in \bbR^{N - r \times K}$ and $\bfZ \in \bbR^{N - r \times K}$ correspond to the leading $K$ eigenvectors of $\bfY^\intercal \calL \bfY$ and $\bfY^\intercal \bfL \bfY$, respectively. Define $\gamma = \mu_{K + 1} - \mu_{K}$. Then, with probability at least $1 - 2N^{-\alpha}$, $$\inf_{\bfU \in \bbR^{K \times K} : \bfU\bfU^\intercal = \bfU^\intercal \bfU = \bfI} \norm{\bfY \calZ - \bfY \bfZ \bfU}[F] \leq \const_3(C, \alpha) \frac{4\sqrt{2K}}{\gamma} \sqrt{p N \ln N},$$ where $\const_3(C, \alpha)$ is from \eqref{eq:L-calL_bound}.
\end{lemma}

Recall that $\bfX \in \bbR^{N \times K}$ is a matrix that contains $\bfy_1, \dots, \bfy_K$ as its columns. Let $\bfx_i$ denote the $i^{th}$ row of $\bfX$. Simple calculation using Lemma \ref{lemma:orthonormal_eigenvectors_y2_yK} shows that,
\begin{equation*}
    \norm{\bfx_i - \bfx_j}[2] = \begin{cases}
        0 & \text{ if }v_i \text{ and }v_j \text{ belong to the same cluster} \\
        \sqrt{\frac{2K}{N}} & \text{ otherwise.}
\end{cases}
\end{equation*}
By Lemma \ref{lemma:first_K_eigenvectors_of_L}, $\calZ$ can be chosen such that $\bfY \calZ = \bfX$. Let $\bfU$ be the matrix that solves $\inf_{\bfU \in \bbR^{K \times K} : \bfU\bfU^\intercal = \bfU^\intercal \bfU = \bfI} \norm{\bfY \calZ - \bfY \bfZ \bfU}[F]$. As $\bfU$ is orthogonal, $\norm{\bfx_i^\intercal \bfU - \bfx_j^\intercal \bfU}[2] = \norm{\bfx_i - \bfx_j}[2]$. The following lemma is a direct consequence of Lemma 5.3 in \citep{LeiEtAl:2015:ConsistencyOfSpectralClusteringInSBM}.

\begin{lemma}
    \label{lemma:k_means_error}
    Let $\bfX$ and $\bfU$ be as defined above. For any $\epsilon > 0$, let $\hat{\bmTheta} \in \bbR^{N \times K}$ be the assignment matrix returned by a $(1 + \epsilon)$-approximate solution to the $k$-means clustering problem when rows of $\bfY \bfZ$ are provided as input features. Further, let $\hat{\bmmu}_1$, $\hat{\bmmu}_2$, \dots, $\hat{\bmmu}_K \in \bbR^{K}$ be the estimated cluster centroids. Define $\hat{\bfX} = \hat{\bmTheta} \hat{\bmmu}$ where $\hat{\bmmu} \in \bbR^{K \times K}$ contains $\hat{\bmmu}_1, \dots, \hat{\bmmu}_K$ as its rows. Further, define $\delta = \sqrt{\frac{2K}{N}}$, and $S_k = \{v_i \in \calC_k : \norm{\hat{\bfx}_i - \bfx_i} \geq \delta/2\}$. Then,
    \begin{equation}
        \label{eq:num_mistakes_bound}
        \delta^2 \sum_{k = 1}^K \abs{S_k} \leq 8(2 + \epsilon) \norm{\bfX \bfU^\intercal - \bfY \bfZ}[F][2].
    \end{equation}
    Moreover, if $\gamma$ from Lemma \ref{lemma:bound_on_eigenvector_diff} satisfies $\gamma^2 > \const(C, \alpha) (2 + \epsilon) p NK \ln N$ for a universal constant $\const(C, \alpha)$, there exists a permutation matrix  $\bfJ \in \bbR^{K \times K}$ such that 
    \begin{equation}
        \label{eq:correct_solution_on_non-mistakes}
        \hat{\bmtheta}_i^\intercal \bfJ = \bmtheta_i^\intercal, \,\,\,\, \forall \,\, i \in [N] \backslash (\cup_{k=1}^K S_k).    
    \end{equation}
    Here, $\hat{\bmtheta}_i \bfJ$ and $\bmtheta_i$ represent the $i^{th}$ row of matrix $\hat{\bmTheta}\bfJ$ and $\bmTheta$ respectively.
\end{lemma}
    
By the definition of $M(\bmTheta, \hat{\bmTheta})$, for the matrix $\bfJ$ used in Lemma \ref{lemma:k_means_error}, $M(\bmTheta, \hat{\bmTheta}) \leq \frac{1}{N} \norm{\bmTheta - \hat{\bmTheta} \bfJ}[0]$. But, according to Lemma \ref{lemma:k_means_error},
$\norm{\bmTheta - \hat{\bmTheta} \bfJ}[0] \leq 2 \sum_{k = 1}^K \abs{S_k}$. Using Lemma \ref{lemma:bound_on_eigenvector_diff} and \ref{lemma:k_means_error}, we get:
\begin{eqnarray*}
    M(\bmTheta, \hat{\bmTheta}) \leq \frac{1}{N} \norm{\bmTheta - \hat{\bmTheta} \bfJ}[0] \leq \frac{2}{N} \sum_{k = 1}^K \abs{S_k}
    &\leq& \frac{16(2 + \epsilon)}{N \delta^2} \norm{\bfX \bfU^\intercal - \bfY \bfZ}[F][2] \\
    &\leq& \const_3(C, \alpha)^2 \frac{512(2 + \epsilon)}{N \delta^2 \gamma^2} p N K \ln N.
\end{eqnarray*}
Noting that $\delta = \sqrt{\frac{2K}{N}}$ and setting $\const(C, \alpha) = 256 \times \const_3(C, \alpha)^2$ finishes the proof.

%%%%%%%%%%%%%%%%%%%%%%%%%%%%%%%%%%%%%%%%%%%%%%

\subsubsection{Proof of Theorem \ref{theorem:consistency_result_normalized}}
\label{section:proof_consistency_normalized}

Recall that $\bfQ = \sqrt{\bfY^\intercal \bfD \bfY}$ and analogously define $\calQ = \sqrt{\bfY^\intercal \calD \bfY}$, where $\calD$ is the expected degree matrix. It was shown after Lemma \ref{lemma:uk_eigenvector_of_tildeA} that $\calD = (\lambda_1 - p) \bfI$. Thus, $\calQ = \sqrt{\lambda_1 - p} \;\bfI$ as $\bfY^\intercal \bfY = \bfI$. Hence $\calQ^{-1} \bfY^\intercal \calL \bfY \calQ^{-1} = \frac{1}{\lambda_1 - p} \bfY^\intercal \calL \bfY$. Therefore, $\calQ^{-1} \bfY^\intercal \calL \bfY \calQ^{-1} \bfx = \frac{\lambda}{\lambda_1 - p} \bfx \; \Longleftrightarrow \;  \bfY^\intercal \calL \bfY \bfx = \lambda \bfx$. Let $\calZ \in \bbR^{N - r \times K}$ contain the leading $K$ eigenvectors of $\calQ^{-1} \bfY^\intercal \calL \bfY \calQ^{-1}$ as its columns. Algorithm \ref{alg:nrepsc} will cluster the rows of $\bfY \calQ^{-1} \calZ$ to recover the clusters in the expected case. As $\calQ^{-1} =\frac{1}{\sqrt{\lambda_1 - p}} \bfI$, we have $\bfY \calQ^{-1} \calZ = \frac{1}{\sqrt{\lambda_1 - p}} \bfY \calZ$. By Lemma \ref{lemma:first_K_eigenvectors_of_L}, $\calZ$ can always be chosen such that $\bfY \calZ = \bfX$, where recall that $\bfX \in \bbR^{N \times K}$ has $\bfy_1, \dots, \bfy_K$ as its columns. Because the rows of $\bfX$ are identical for nodes that belong to the same cluster, Algorithm \ref{alg:nrepsc} returns the correct ground truth clusters in the expected case. 

To bound the number of mistakes made by Algorithm \ref{alg:nrepsc}, we show that $\bfY \bfQ^{-1} \bfZ$ is close to $\bfY \calQ^{-1} \calZ$. Here, $\bfZ \in \bbR^{N - r \times K}$ contains the top $K$ eigenvectors of $\bfQ^{-1} \bfY^\intercal \bfL \bfY \bfQ^{-1}$. As in the proof of Lemma \ref{lemma:bound_on_eigenvector_diff}, we use Davis-Kahan theorem to bound this difference. This requires us to compute $\norm{\calQ^{-1} \bfY^\intercal \calL \bfY \calQ^{-1} - \bfQ^{-1} \bfY^\intercal \bfL \bfY \bfQ^{-1}}$. Note that:
\begin{align*}
    \norm{\calQ^{-1} \bfY^\intercal \calL \bfY \calQ^{-1} - \bfQ^{-1} \bfY^\intercal \bfL \bfY \bfQ^{-1}} = &\norm{\calQ^{-1} - \bfQ^{-1}} \cdot \norm{\bfY^\intercal \calL \bfY} \cdot \norm{\calQ^{-1}} + \\
    &\norm{\bfQ^{-1}} \cdot \norm{\bfY^\intercal \calL \bfY - \bfY^\intercal \bfL \bfY} \cdot \norm{\calQ^{-1}} + \\
    &\norm{\bfQ^{-1}} \cdot \norm{\bfY^\intercal \bfL \bfY} \cdot \norm{\calQ^{-1} - \bfQ^{-1}}.
\end{align*}
We already have a bound on $\norm{\bfY^\intercal \calL \bfY - \bfY^\intercal \bfL \bfY}$ in \eqref{eq:L-calL_bound}. Also, note that $\norm{\calQ^{-1}} = \frac{1}{\sqrt{\lambda_1 - p}}$ as $\calQ^{-1} = \frac{1}{\sqrt{\lambda_1 - p}} \bfI$. Similarly, as $\bfY^\intercal \bfY = \bfI$, $\norm{\bfY^\intercal \calL \bfY} \leq \norm{\calL} = \lambda_1 - \bar{\lambda}$, where $\bar{\lambda} = \lambdamin{\tilde{\calA}}$. Finally,
\begin{align*}
    \norm{\bfQ^{-1}} &\leq \norm{\calQ^{-1} - \bfQ^{-1}} + \norm{\calQ^{-1}} = \norm{\calQ^{-1} - \bfQ^{-1}} + \frac{1}{\sqrt{\lambda_1 - p}} \text{, and} \\
    \norm{\bfY^\intercal \bfL \bfY} &\leq \norm{\bfY^\intercal \calL \bfY - \bfY^\intercal \bfL \bfY} + \norm{\bfY^\intercal \calL \bfY} = \norm{\bfY^\intercal \calL \bfY - \bfY^\intercal \bfL \bfY} + \lambda_1 - \bar{\lambda}.
\end{align*}
Thus, to compute a bound on $\norm{\calQ^{-1} \bfY^\intercal \calL \bfY \calQ^{-1} - \bfQ^{-1} \bfY^\intercal \bfL \bfY \bfQ^{-1}}$, we only need a bound on $\norm{\calQ^{-1} - \bfQ^{-1}}$. The next lemma provides this bound.

\begin{lemma}
    \label{lemma:calQ-Q_bound}
    Let $\calQ = \sqrt{\bfY^\intercal \calD \bfY}$, $\bfQ = \sqrt{\bfY^\intercal \bfD \bfY}$, and assume that
    $$\left(\frac{\sqrt{pN \ln N}}{\lambda_1 - p}\right)\left(\frac{\sqrt{pN \ln N}}{\lambda_1 - p} + \frac{1}{6\sqrt{C}} \right) \leq \frac{1}{16(\alpha + 1)},$$
    where $C$ and $\alpha$ are used in $\const_1(C, \alpha)$ defined in Lemma \ref{lemma:bound_on_D-calD}. Then,
    $$\norm{\calQ^{-1} - \bfQ^{-1}} \leq \sqrt{\frac{2}{(\lambda_1 - p)^3}} \norm{\bfD - \calD}.$$
\end{lemma}

\noindent
Using the lemma above with \eqref{eq:L-calL_bound}, we get
{
\small
\begin{align}
    \label{eq:calQYcalLYcalQ-QYLYQ_bound}
    \begin{aligned}
    \norm{\calQ^{-1} \bfY^\intercal \calL \bfY \calQ^{-1} - \bfQ^{-1} \bfY^\intercal \bfL &\bfY \bfQ^{-1}} \leq \frac{2(\lambda_1 - \bar{\lambda})}{(\lambda_1 - p)^2} \left[ \sqrt{2} + \frac{\norm{\bfD - \calD}}{\lambda_1 - p}\right] \norm{\bfD - \calD} + \\
    & \frac{\const_3(C, \alpha)}{\lambda_1 - p} \left[\frac{2\sqrt{2} \norm{\bfD - \calD}}{\lambda_1 - p} + \frac{2 \norm{\bfD - \calD}[][2]}{(\lambda_1 - p)^2} + 1  \right] \sqrt{p N \ln N}.
    \end{aligned}
\end{align}
}

\noindent
The next lemma uses the bound above to show that $\bfY \bfQ^{-1} \bfZ$ is close to $\bfY \calQ^{-1} \calZ$.

\begin{lemma}
    \label{lemma:eigenvector_diff_bound_normalized}
    Let $\mu_1 \leq \mu_2 \leq \dots \leq \mu_{N - r}$ be eigenvalues of $\calQ^{-1} \bfY^\intercal \calL \bfY \calQ^{-1}$. Further, let the columns of $\calZ \in \bbR^{N - r \times K}$ and $\bfZ \in \bbR^{N - r \times K}$ correspond to the leading $K$ eigenvectors of $\calQ^{-1} \bfY^\intercal \calL \bfY \calQ^{-1}$ and $\bfQ^{-1} \bfY^\intercal \bfL \bfY \bfQ^{-1}$, respectively. Define $\gamma = \mu_{K + 1} - \mu_{K}$ and let there be a constant $\const_4(C, \alpha)$ such that 
    $\frac{\sqrt{p N \ln N}}{\lambda_1 - p} \leq \const_4(C, \alpha)$. Then, with probability at least $1 - 2N^{-\alpha}$, there exists a constant $\const_5(C, \alpha)$ such that
    \begin{align*}
        \inf_{\bfU : \bfU^\intercal \bfU = \bfU\bfU^\intercal = \bfI} \norm{\bfY \calQ^{-1} \calZ - &\bfY \bfQ^{-1} \bfZ \bfU}[F] \leq \\
        &\left[\frac{16 K \const_5(C, \alpha)}{\gamma (\lambda_1 - p)^{3/2}} + \frac{2 \const_1(C, \alpha) \sqrt{K}}{(\lambda_1 - p)^{3/2}} \right] \sqrt{p N \ln N},
    \end{align*}
    where $\const_1(C, \alpha)$ is defined in Lemma \ref{lemma:bound_on_D-calD}.
\end{lemma}

Recall that, by Lemma \ref{lemma:first_K_eigenvectors_of_L}, $\calZ$ can always be chosen such that $\bfY \calZ = \bfX$, where $\bfX$ contains $\bfy_1, \dots, \bfy_K$ as its columns. As $\calQ^{-1} = \frac{1}{\sqrt{\lambda_1 - p}} \bfI$, one can show that:
\begin{equation*}
    \norm{(\calQ^{-1} \bfX)_i - (\calQ^{-1} \bfX)_j}[2] = \begin{cases}
        0 & \text{ if } v_i \text{ and } v_j \text{ belong to the same cluster} \\
        \sqrt{\frac{2K}{N(\lambda_1 - p)}} & \text{ otherwise.}
    \end{cases}
\end{equation*}
Here, $(\calQ^{-1} \bfX)_i$ denotes the $i^{th}$ row of the matrix $\bfY \calQ^{-1} \calZ$. Let $\bfU$ be the matrix that solves $\inf_{\bfU \in \bbR^{K \times K} : \bfU\bfU^\intercal = \bfU^\intercal \bfU = \bfI} \norm{\bfY \calQ^{-1} \calZ - \bfY \bfQ^{-1} \bfZ \bfU}[F]$. As $\bfU$ is orthogonal, $\norm{(\calQ^{-1} \bfX)_i^\intercal \bfU - (\calQ^{-1} \bfX)_j^\intercal \bfU}[2] = \norm{(\calQ^{-1} \bfX)_i - (\calQ^{-1} \bfX)_j}[2]$. As in the previous case, the following lemma is a direct consequence of Lemma 5.3 in \citep{LeiEtAl:2015:ConsistencyOfSpectralClusteringInSBM}.

\begin{lemma}
    \label{lemma:k_means_error_normalized}
    Let $\bfX$ and $\bfU$ be as defined above. For any $\epsilon > 0$, let $\hat{\bmTheta} \in \bbR^{N \times K}$ be the assignment matrix returned by a $(1 + \epsilon)$-approximate solution to the $k$-means clustering problem when rows of $\bfY \bfQ^{-1} \bfZ$ are provided as input features. Further, let $\hat{\bmmu}_1$, $\hat{\bmmu}_2$, \dots, $\hat{\bmmu}_K \in \bbR^{K}$ be the estimated cluster centroids. Define $\hat{\bfX} = \hat{\bmTheta} \hat{\bmmu}$ where $\hat{\bmmu} \in \bbR^{K \times K}$ contains $\hat{\bmmu}_1, \dots, \hat{\bmmu}_K$ as its rows. Further, define $\delta = \sqrt{\frac{2K}{N(\lambda_1 - p)}}$, and $S_k = \{v_i \in \calC_k : \norm{\hat{\bfx}_i - \bfx_i} \geq \delta/2\}$. Then,
    \begin{equation*}
        \delta^2 \sum_{k = 1}^K \abs{S_k} \leq 8(2 + \epsilon) \norm{\bfX \bfU^\intercal - \bfY \bfQ^{-1} \bfZ}[F][2].
    \end{equation*}
    Moreover, if $\gamma$ from Lemma \ref{lemma:eigenvector_diff_bound_normalized} satisfies $$16(2 + \epsilon)\left[ \frac{8 \const_5(C, \alpha) \sqrt{K}}{\gamma} + \const_1(C, \alpha)\right]^2 \frac{p N^2 \ln N}{(\lambda_1 - p)^2} < \frac{N}{K},$$ then, there exists a permutation matrix  $\bfJ \in \bbR^{K \times K}$ such that 
    \begin{equation*}
        \hat{\bmtheta}_i^\intercal \bfJ = \bmtheta_i^\intercal, \,\,\,\, \forall \,\, i \in [N] \backslash (\cup_{k=1}^K S_k).    
    \end{equation*}
    Here, $\hat{\bmtheta}_i \bfJ$ and $\bmtheta_i$ represent the $i^{th}$ row of matrix $\hat{\bmTheta}\bfJ$ and $\bmTheta$ respectively.
\end{lemma}

The proof of Lemma \ref{lemma:k_means_error_normalized} is similar to that of Lemma \ref{lemma:k_means_error}, and has been omitted. The result follows by using a similar calculation as was done after Lemma \ref{lemma:k_means_error} in Section \ref{section:proof_consistency_unnormalized}.

%%%%%%%%%%%%%%%%%%%%%%%%%%%%%%%%%%%%%%%%%%%%%%

\section{Numerical Results}
\label{section:numerical_results}

We perform three types of experiments. In the first two cases, we use synthetically generated data to validate our theoretical results using $d$-regular representation graphs (Section \ref{chapter:fairness:section:d_reg_experiments}) and non-$d$-regular representation graphs (Section \ref{chapter:fairness:section:sbm_experiments}). In the third case, we demonstrate the effectiveness of the proposed algorithms on
a real-world dataset (Section~\ref{chapter:fairness:section:trade_experiments}). Notably, our experiments in Sections \ref{chapter:fairness:section:sbm_experiments} and \ref{chapter:fairness:section:trade_experiments} demonstrate that the $d$-regularity assumption on $\calR$ is not needed in practice. Before proceeding further, we mention two important details below: \textbf{(i)} How do we compare with the algorithms presented in \cite{KleindessnerEtAl:2019:GuaranteesForSpectralClusteringWithFairnessConstraints}? and \textbf{(ii)} What do we do when the rank assumption on $\calR$ is not satisfied?

\begin{figure}[t]
    \centering
    \subfloat[][Accuracy vs no. of nodes]{\includegraphics[width=0.33\textwidth]{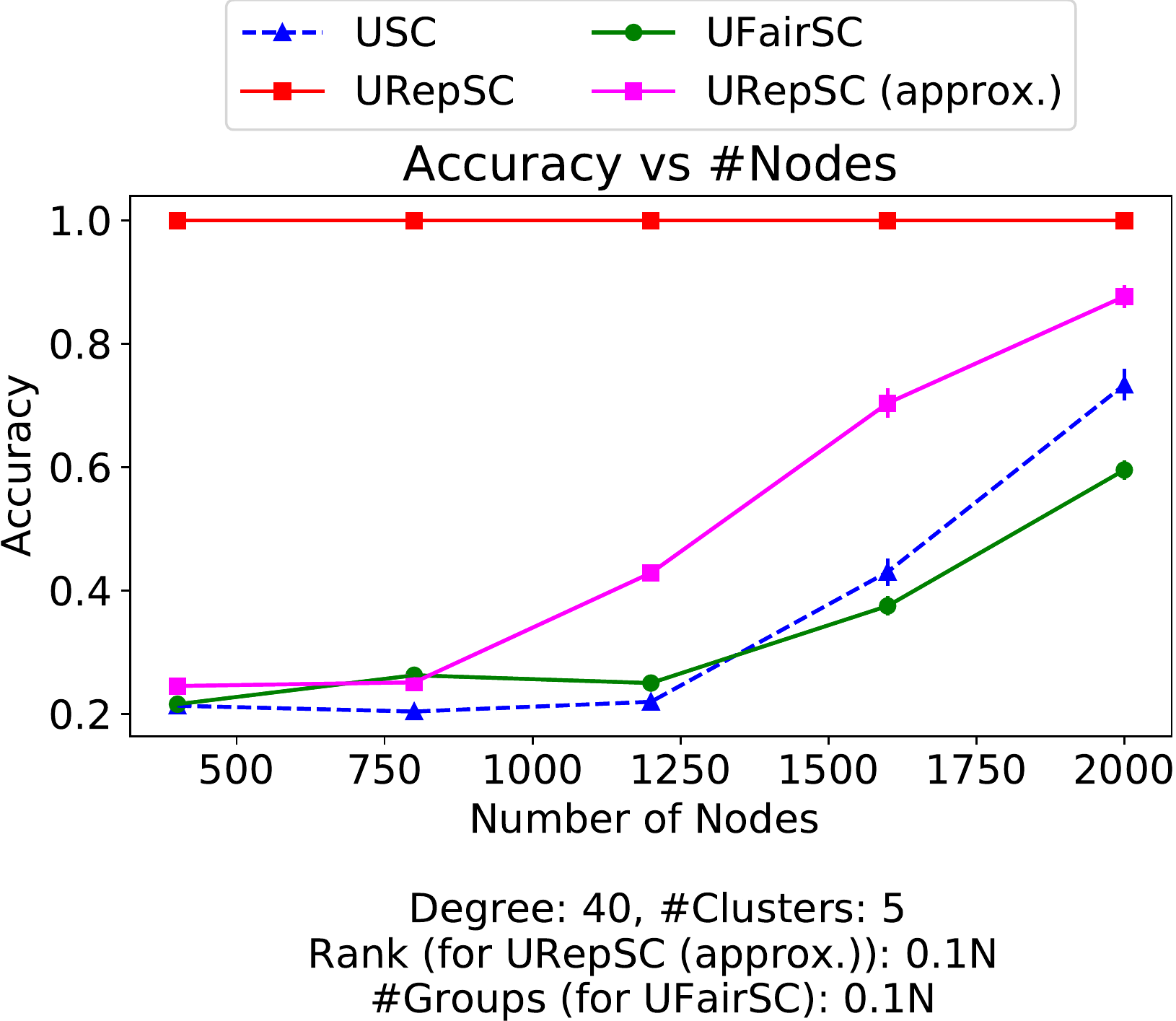}\label{fig:d_reg_unnorm:vs_N}}%
    \subfloat[][Accuracy vs no. of clusters]{\includegraphics[width=0.33\textwidth]{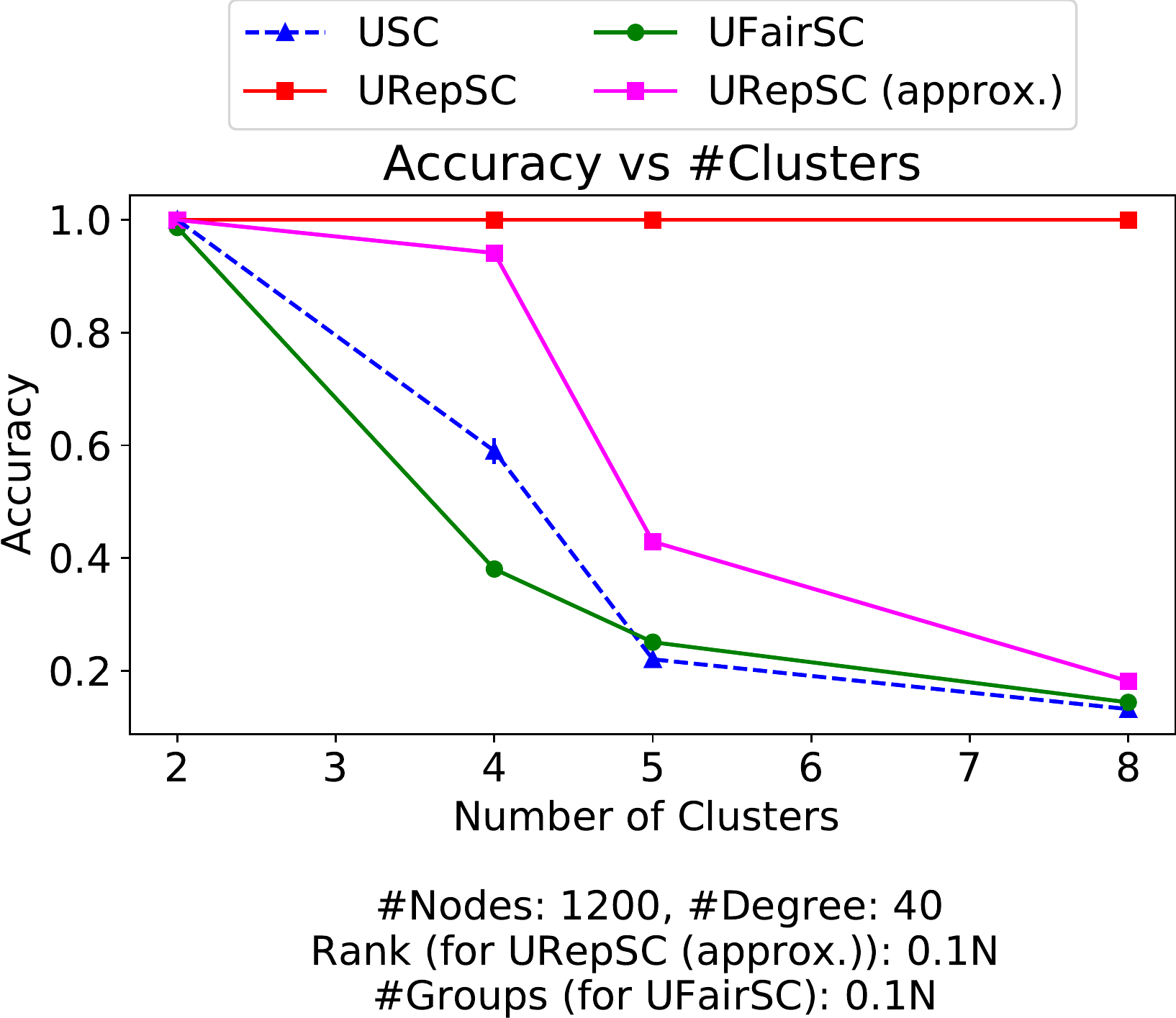}\label{fig:d_reg_unnorm:vs_K}}%
    \subfloat[][Accuracy vs degree of $\calR$]{\includegraphics[width=0.33\textwidth]{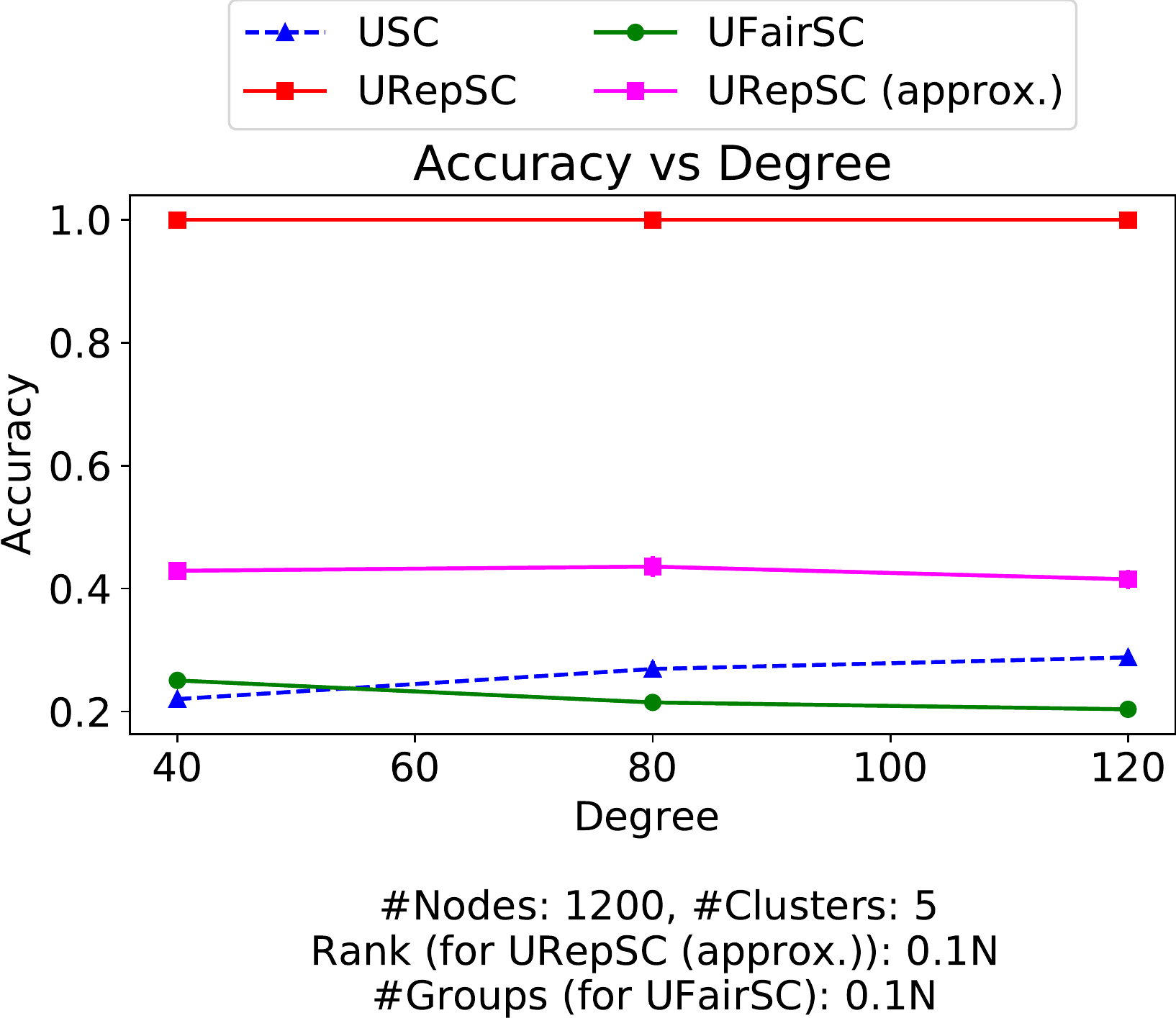}\label{fig:d_reg_unnorm:vs_d}}
    \caption{Comparing \textsc{URepSC} with other ``unnormalized'' algorithms using synthetically generated $d$-regular representation graphs.}
    \label{fig:d_reg_unnorm}
\end{figure}

\begin{figure}[t]
    \centering
    \subfloat[][Accuracy vs no. of nodes]{\includegraphics[width=0.33\textwidth]{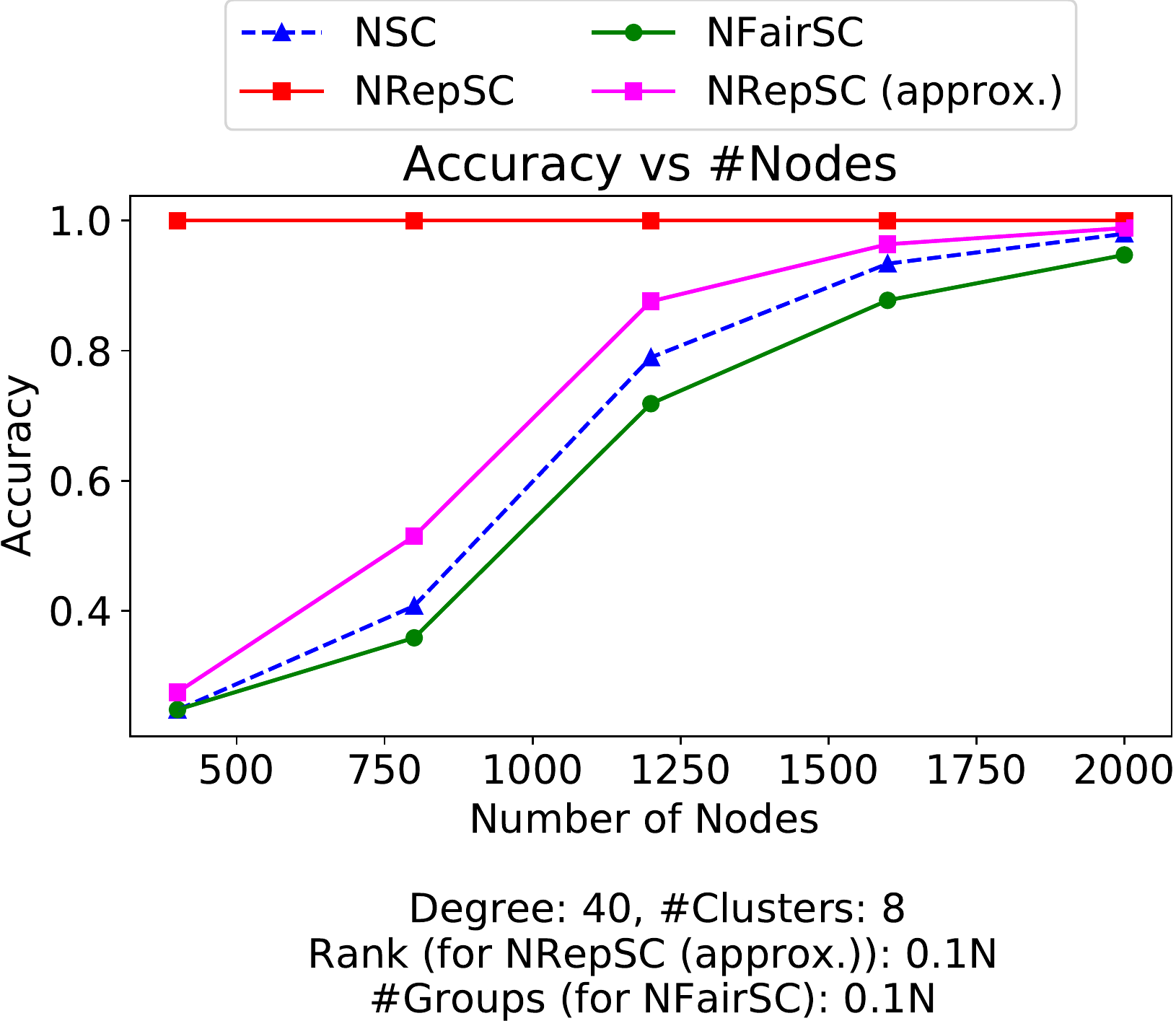}\label{fig:d_reg_norm:vs_N}}%
    \subfloat[][Accuracy vs no. of clusters]{\includegraphics[width=0.33\textwidth]{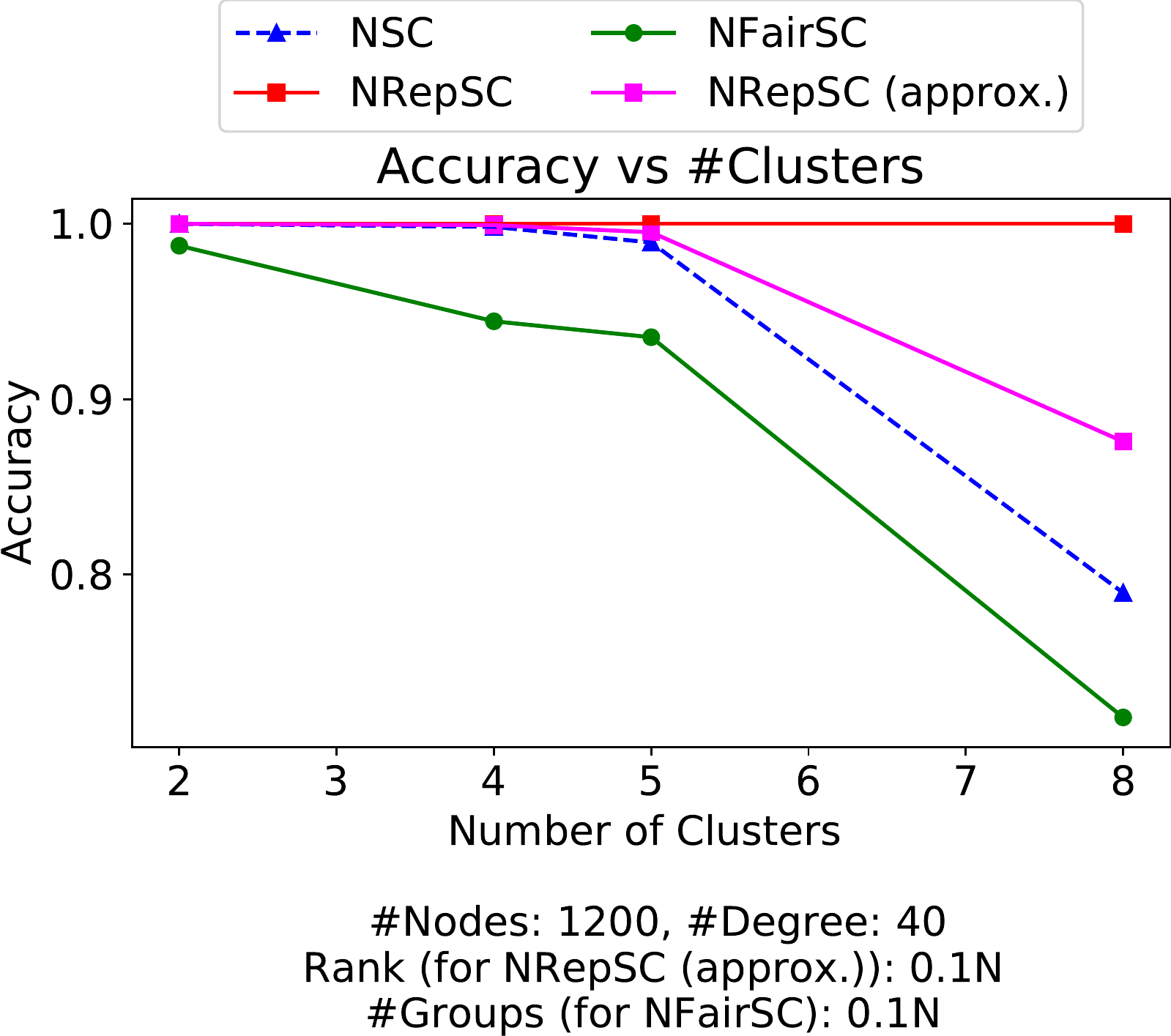}\label{fig:d_reg_norm:vs_K}}%
    \subfloat[][Accuracy vs degree of $\calR$]{\includegraphics[width=0.33\textwidth]{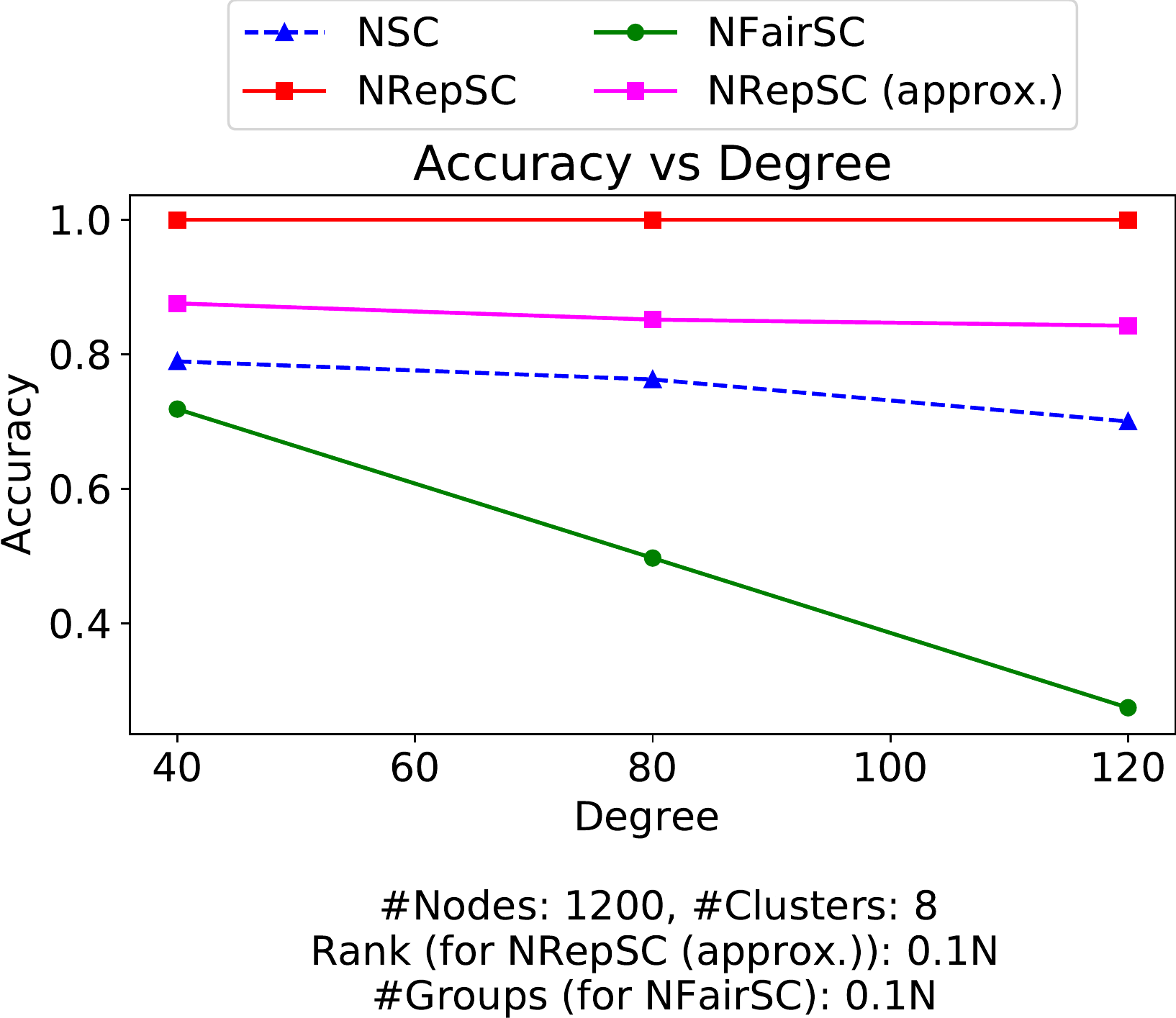}\label{fig:d_reg_norm:vs_d}}
    \caption{Comparing \textsc{NRepSC} with other ``normalized'' algorithms using synthetically generated $d$-regular representation graphs.}
    \label{fig:d_reg_norm}
\end{figure}

\paragraph*{Comparison with \citet{KleindessnerEtAl:2019:GuaranteesForSpectralClusteringWithFairnessConstraints}} We refer to the algorithms proposed in \citet{KleindessnerEtAl:2019:GuaranteesForSpectralClusteringWithFairnessConstraints} as \textsc{UFairSC} and \textsc{NFairSC}, corresponding to the unnormalized and normalized fair spectral clustering, respectively. These algorithms assume that each node belongs to one of the $P$ protected groups $\calP_1, \dots, \calP_P \subseteq \calV$ that are observed by the learner. Recall that these algorithms are special cases of our algorithms when $\calR$ is block diagonal (Section \ref{section:constraint}). To demonstrate the generality of our algorithms, we only experiment with representation graphs that are not of the form specified above. Naturally, \textsc{UFairSC} and \textsc{NFairSC} are not directly applicable in this setting. Nonetheless, to compare with these algorithms, we approximate the protected groups by clustering the nodes in $\calR$ using standard spectral clustering. Each discovered cluster is then treated as a protected group.

\paragraph*{Approximate \textsc{URepSC} and \textsc{NRepSC}} Recall that the rank assumption on $\calR$ requires $\rank{\bfR} \leq N - K$. It is not possible to find $K$ orthonormal eigenvectors in Algorithms \ref{alg:urepsc} and \ref{alg:nrepsc} if $\bfR$ violates this assumption. Unlike other assumptions in our theoretical analysis, this assumption is necessary in practice. If a graph $\calR$ violates the rank assumption, we instead use the best rank $R$ approximation of its adjacency matrix $\bfR$ ($R \leq N - K$). This approximation does not have binary elements, but it works well in practice. Whenever this approximation is used, we refer to \textsc{URepSC} and \textsc{NRepSC} as \textsc{URepSC (approx.)} and \textsc{NRepSC (approx.)}, respectively.

% ==================================================== %

\begin{figure}[t]
    \centering
    \subfloat[][Unnormalized case]{\includegraphics[width=0.4\textwidth]{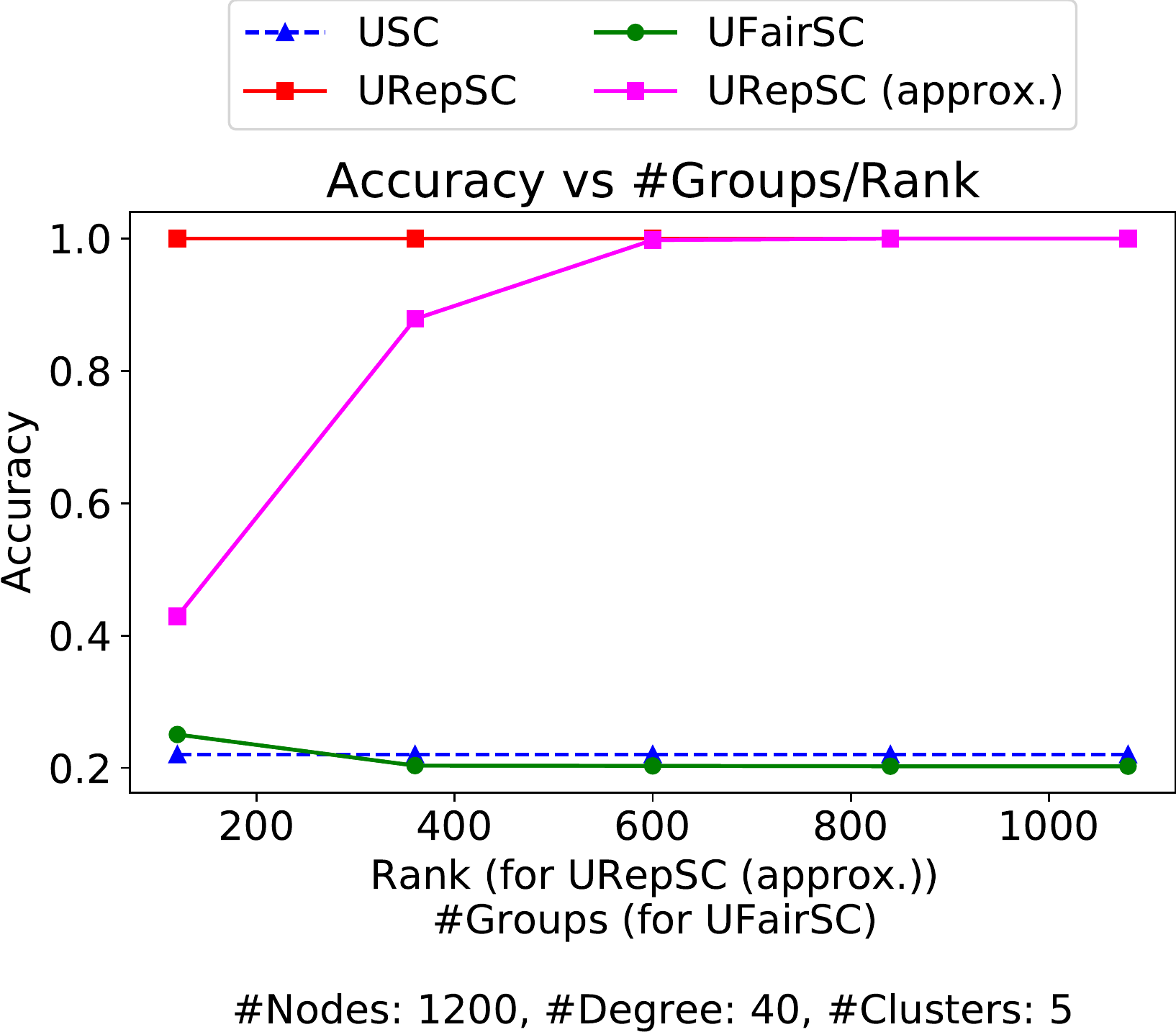}\label{fig:d_reg_unnorm:rank_groups}}%
    \hspace{1cm}\subfloat[][Normalized case]{\includegraphics[width=0.4\textwidth]{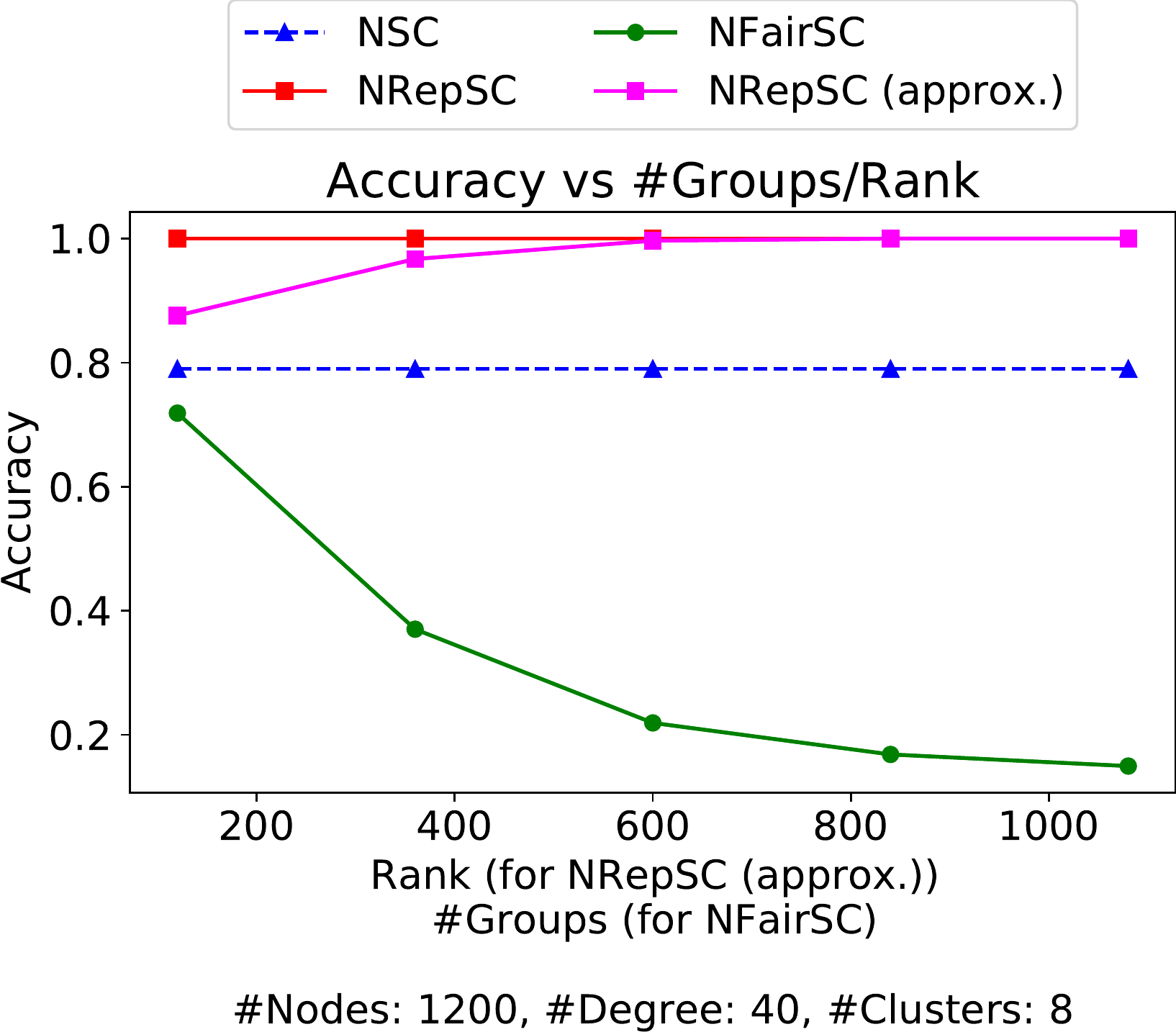}\label{fig:d_reg_norm:rank_groups}}%
    \caption{Accuracy vs the values of $P$ and $R$ used by \textsc{U/NFairSC} and \textsc{U/NRepSC}, respectively, for $d$-regular representation graphs.}
\end{figure}

\subsection{Experiments with $d$-regular representation graphs}
\label{chapter:fairness:section:d_reg_experiments}

For these experiments, we sampled $d$-regular representation graphs using $p=0.4$, $q=0.3$, $r=0.2$, and $s=0.1$, for various values of $d$, $N$, and $K$. We ensured that the sampled $\calR$ satisfies Assumption \ref{assumption:R_is_d_regular} and $\rank{\bfR} \leq N - k$. Further, the ground-truth clusters have equal size and are representation-aware by construction as described in Section \ref{section:consistency_results}. Figure \ref{fig:d_reg_unnorm} compares the performance of \textsc{URepSC} with unnormalized spectral clustering (\textsc{USC}) (Algorithm \ref{alg:unnormalized_spectral_clustering}) and \textsc{UFairSC}. Figure \ref{fig:d_reg_unnorm:vs_N} shows the effect of varying $N$ for a fixed $d = 40$ and $K=5$. Figure \ref{fig:d_reg_unnorm:vs_K} varies $K$ and keeps $N = 1200$ and $d = 40$ fixed. Similarly, Figure \ref{fig:d_reg_unnorm:vs_d} keeps $N = 1200$ and $K = 5$ fixed and varies $d$. In all cases, we use $R = P = N/10$, where recall that $R$ is the rank used for approximation in \textsc{URepSC (approx.)} and $P$ is the number of protected groups discovered in $\calR$ for running \textsc{UFairSC}. The figures plot the accuracy on $y$-axis and report the mean and standard deviation across $10$ independent executions of the algorithms in each case. 

As the ground truth clusters satisfy Definition \ref{def:representation_constraint} by construction, a high accuracy of cluster recovery implies that the algorithm returns representation-aware clusters. Figure \ref{fig:d_reg_norm} shows the corresponding results for \textsc{NRepSC}, where we compare it with the normalized variants of other algorithms. In Figures \ref{fig:d_reg_unnorm:vs_N} and \ref{fig:d_reg_norm:vs_N}, it appears that even the standard spectral clustering algorithm will return representation-aware clusters for a large enough graph. However, Figures \ref{fig:d_reg_unnorm:vs_K} and \ref{fig:d_reg_norm:vs_K} show that this is not true if the number of clusters also increases with $N$, as is more common in practice.

It may also be tempting to think that \textsc{UFairSC} and \textsc{NFairSC} may perform well with a more carefully chosen value of $P$, the number of protected groups. However, Figures \ref{fig:d_reg_unnorm:rank_groups} and \ref{fig:d_reg_norm:rank_groups} show that this is not true. These figures plot the performance of \textsc{UFairSC} and \textsc{NFairSC} as a function of the number of protected groups $P$. Also shown in these plots is the performance of the approximate variants of our algorithms for various values of rank $R$. As expected, the accuracy increases with $R$ as the approximation of $\bfR$ becomes better.

% ==================================================== %

\begin{figure}[t]
    \centering
    \subfloat[][$N = 1000$, $K = 4$]{\includegraphics[width=0.48\textwidth]{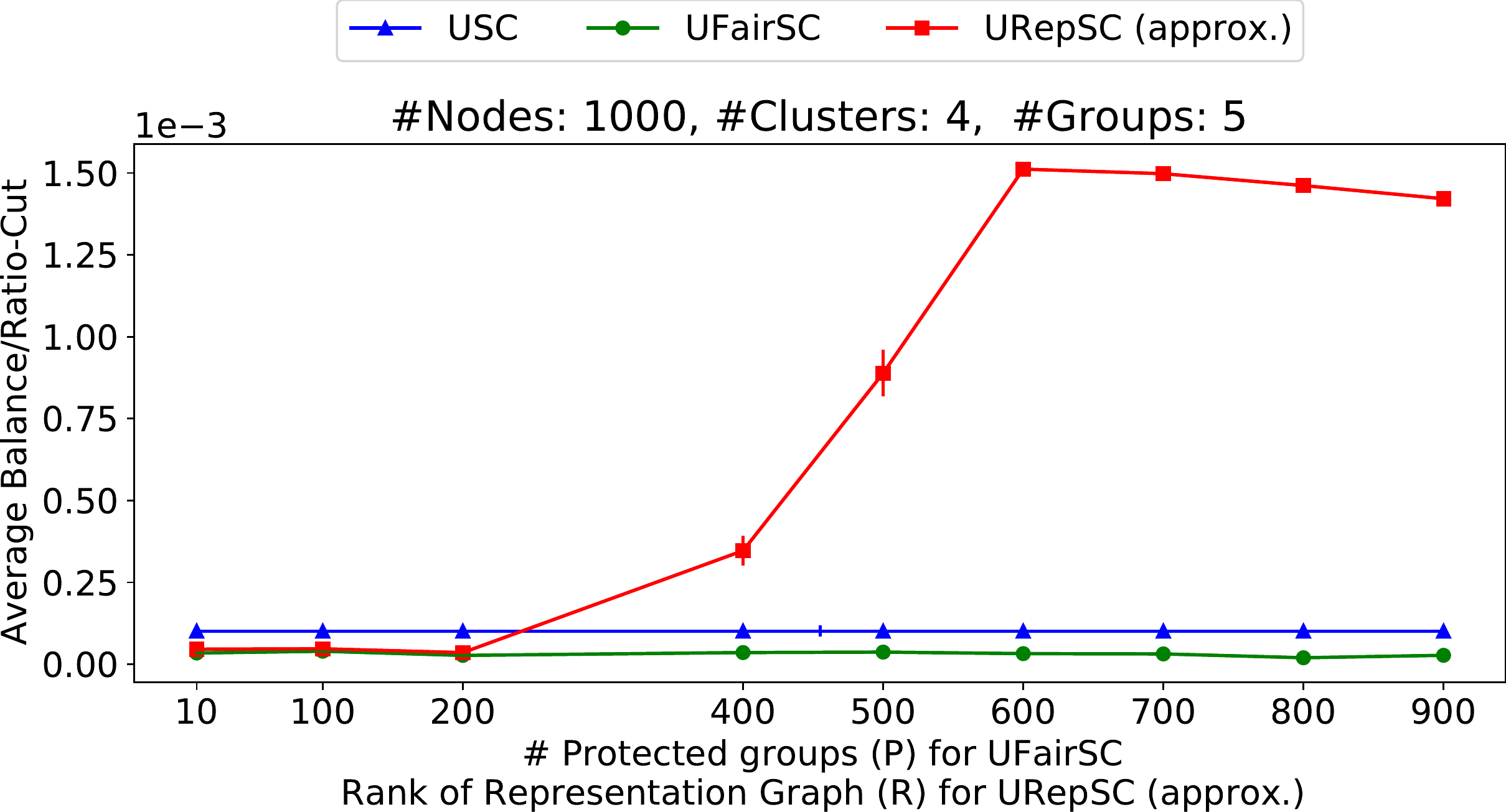}}%
    \hspace{0.5cm}\subfloat[][$N = 3000$, $K = 4$]{\includegraphics[width=0.48\textwidth]{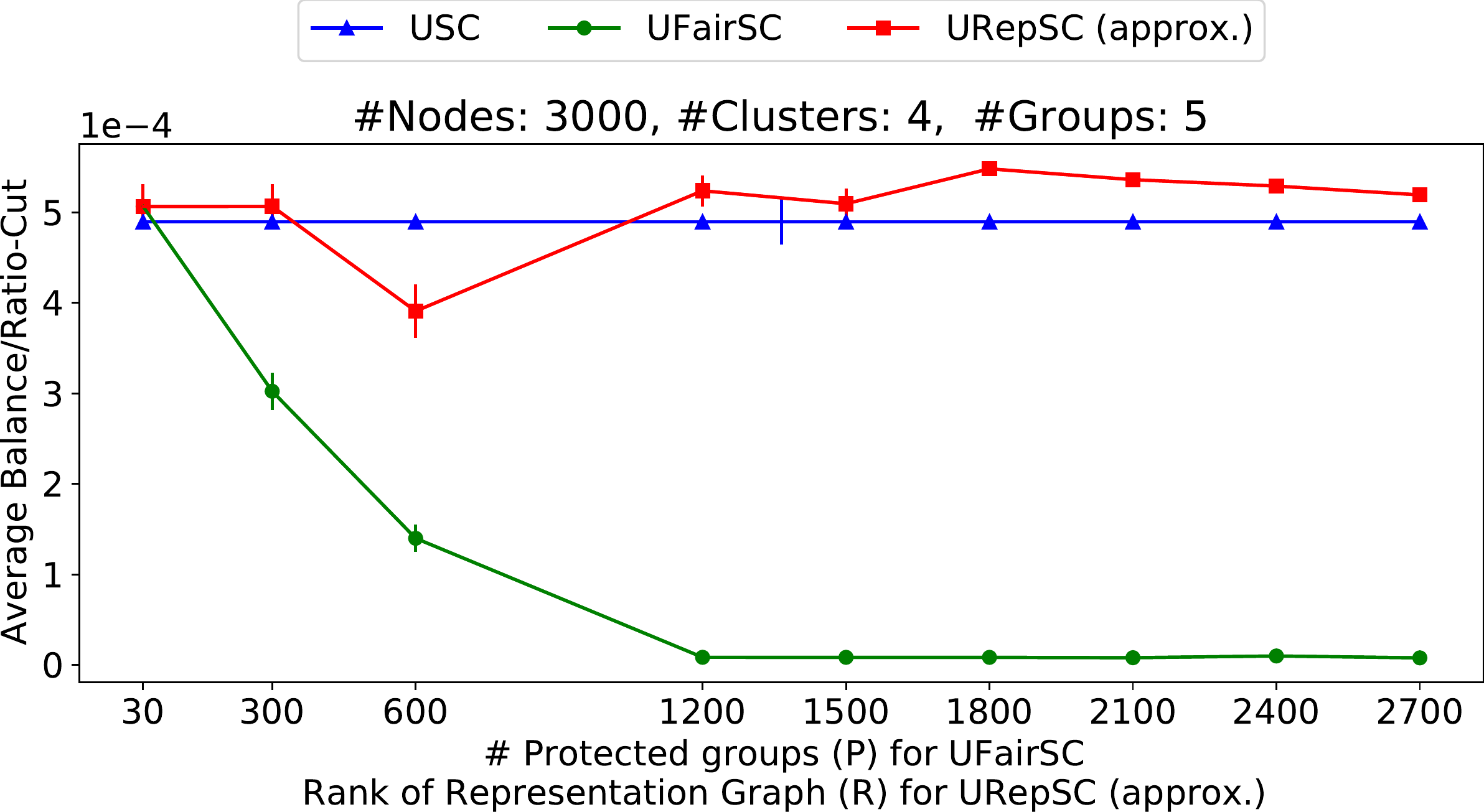}}

    \subfloat[][$N = 1000$, $K = 8$]{\includegraphics[width=0.48\textwidth]{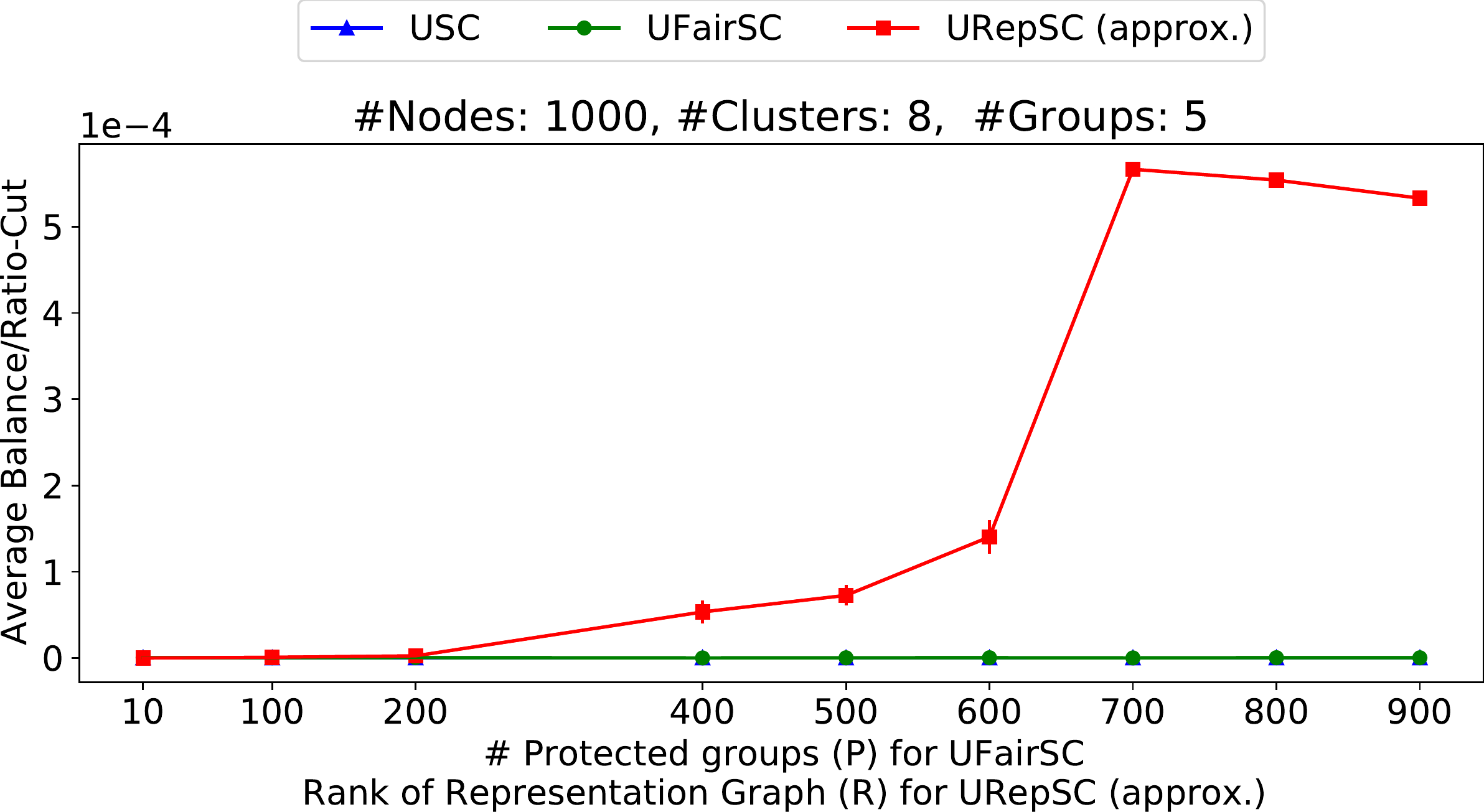}}%
    \hspace{0.5cm}\subfloat[][$N = 3000$, $K = 8$]{\includegraphics[width=0.48\textwidth]{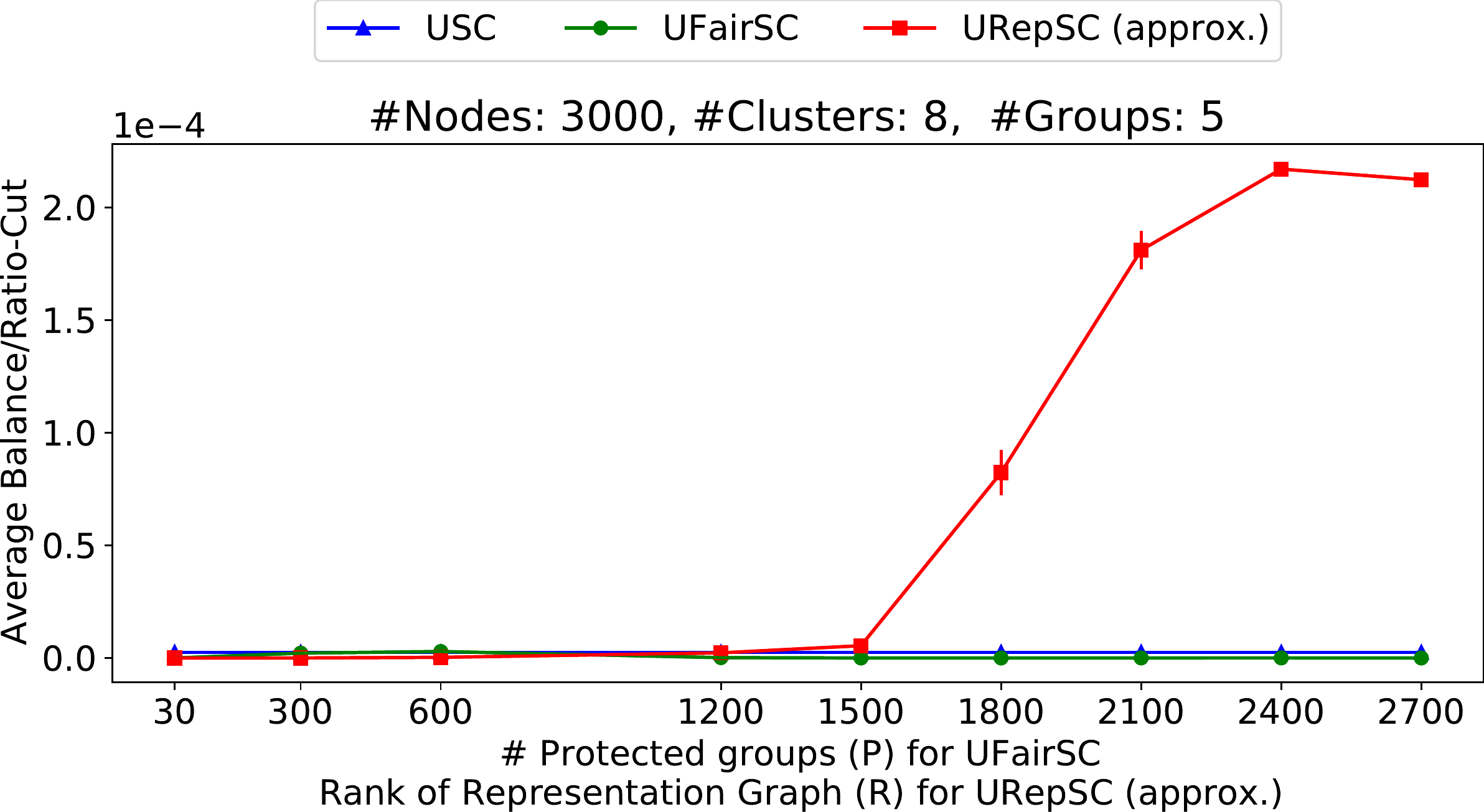}}
    \caption{Comparing \textsc{URepSC (approx.)} with \textsc{UFairSC} using synthetically generated representation graphs sampled from an SBM.}
    \label{fig:sbm_comparison_unnorm}
\end{figure}

\begin{figure}[t]
    \centering
    \subfloat[][$N = 1000$, $K = 4$]{\includegraphics[width=0.48\textwidth]{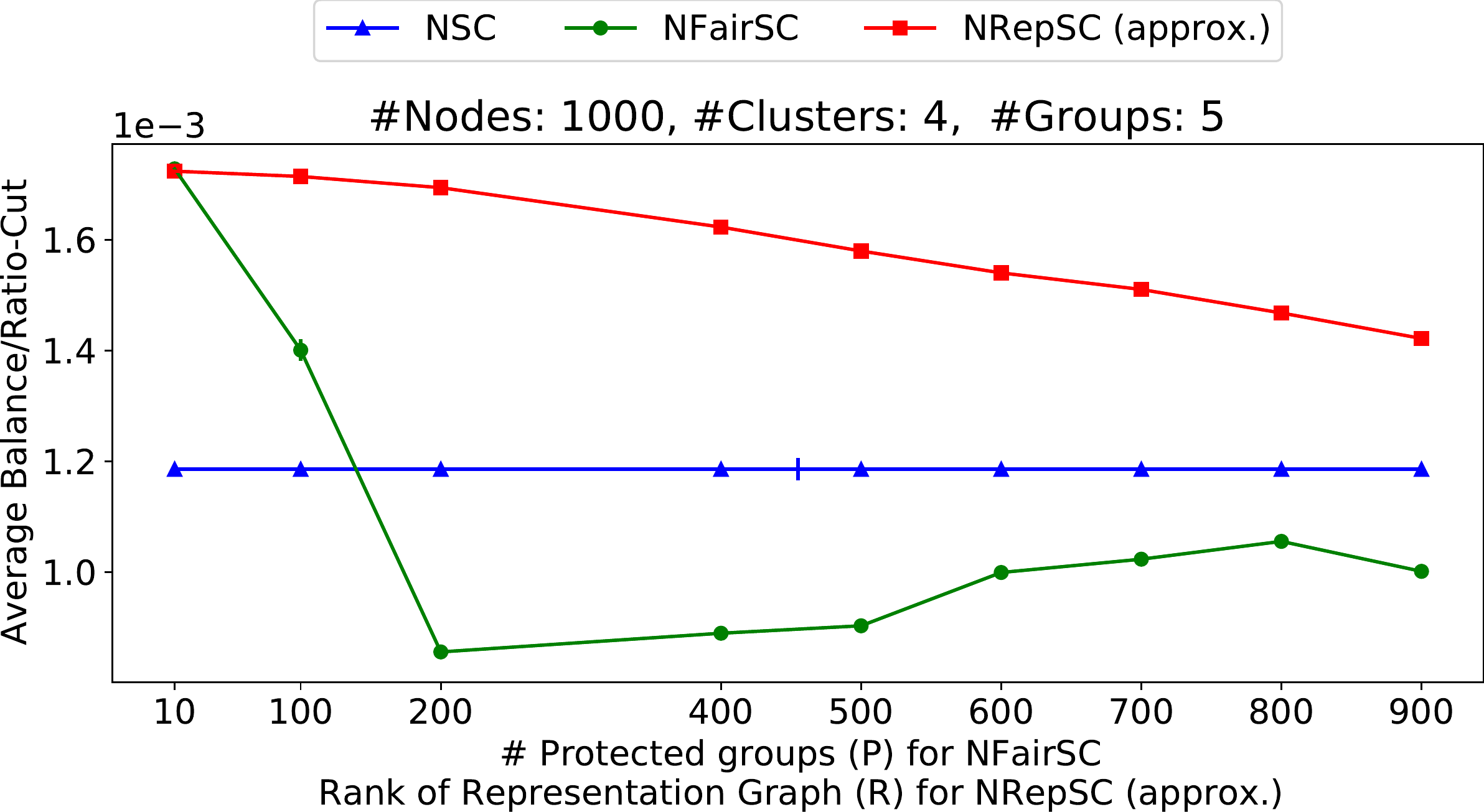}}%
    \hspace{0.5cm}\subfloat[][$N = 3000$, $K = 4$]{\includegraphics[width=0.48\textwidth]{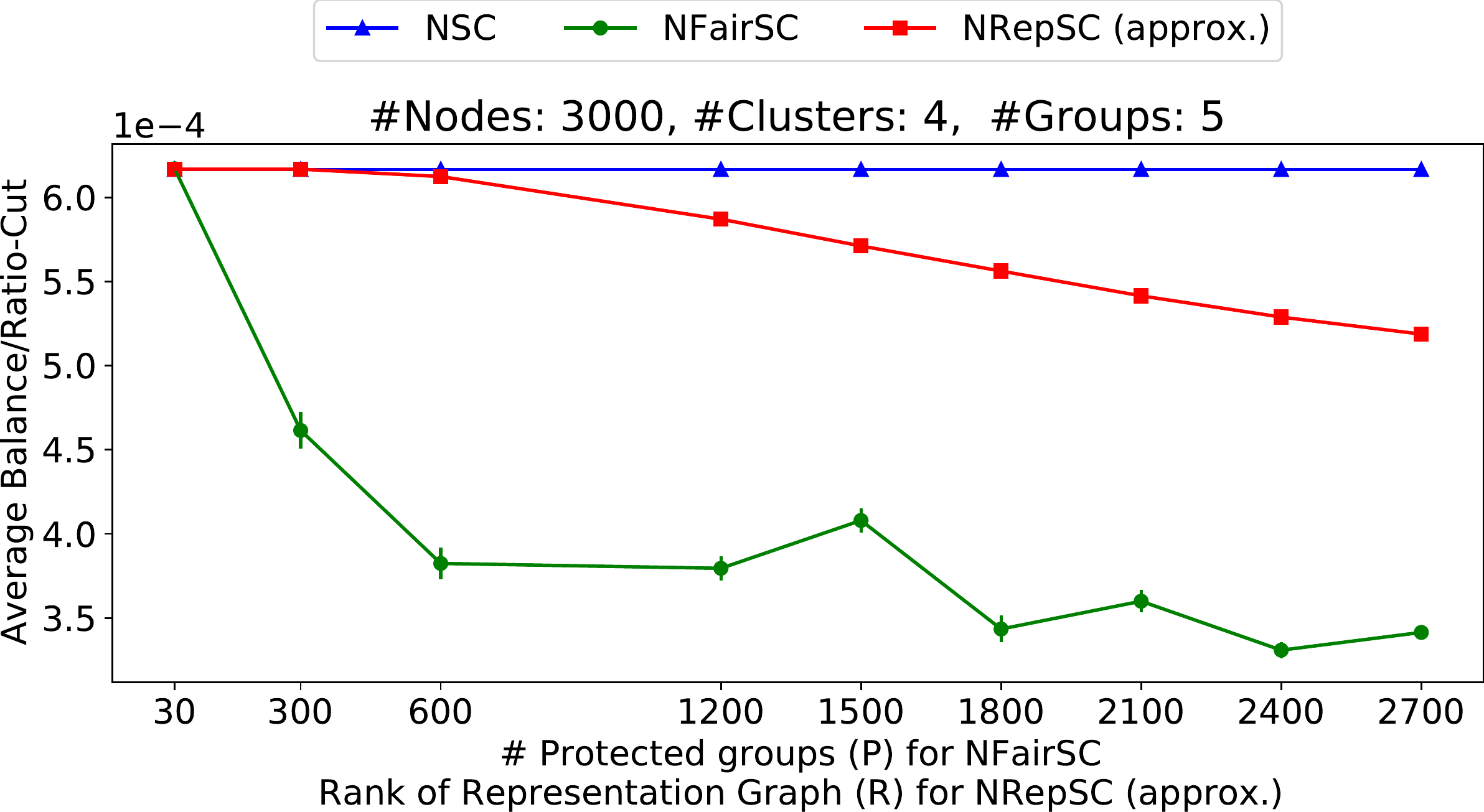}}

    \subfloat[][$N = 1000$, $K = 8$]{\includegraphics[width=0.48\textwidth]{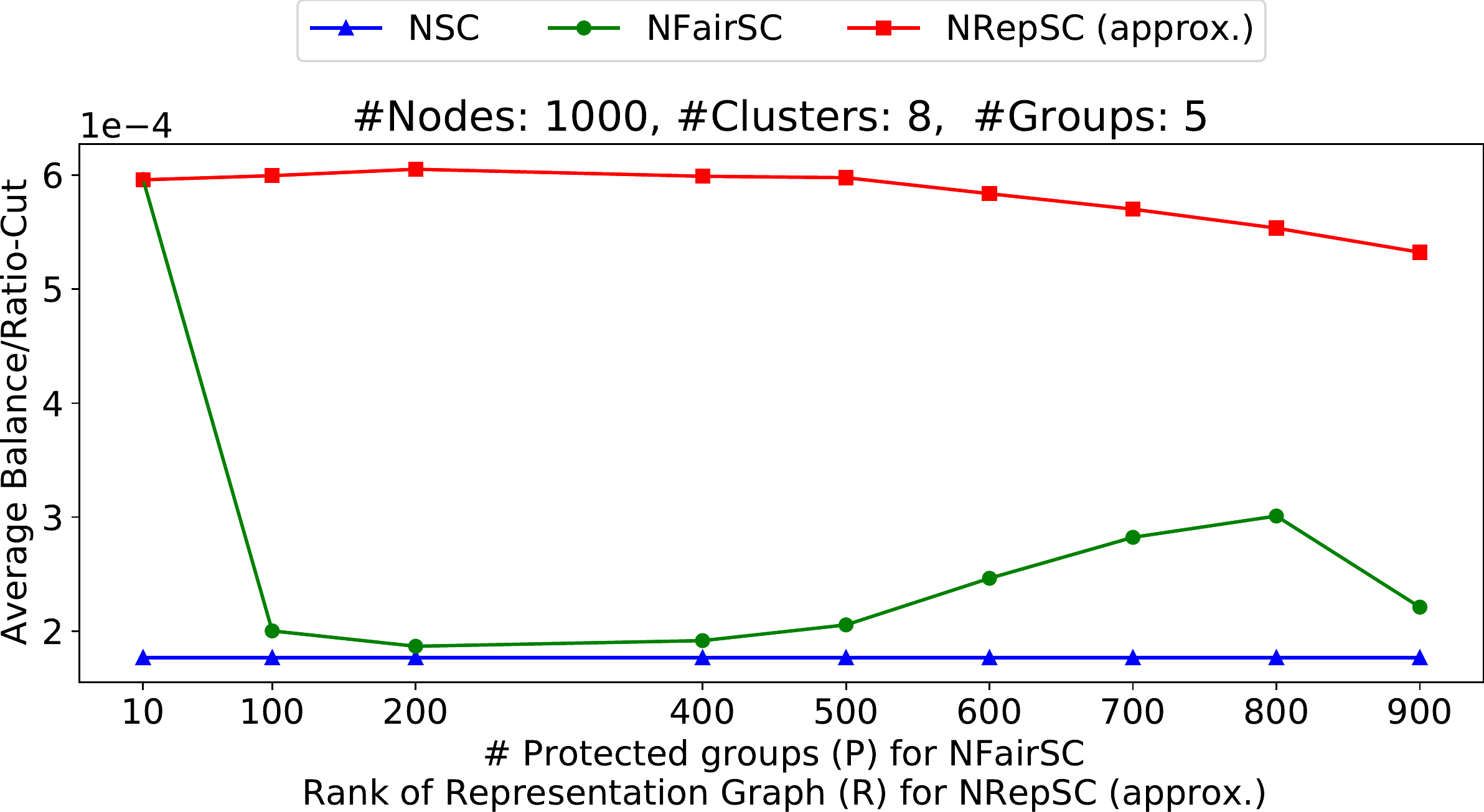}}%
    \hspace{0.5cm}\subfloat[][$N = 3000$, $K = 8$]{\includegraphics[width=0.48\textwidth]{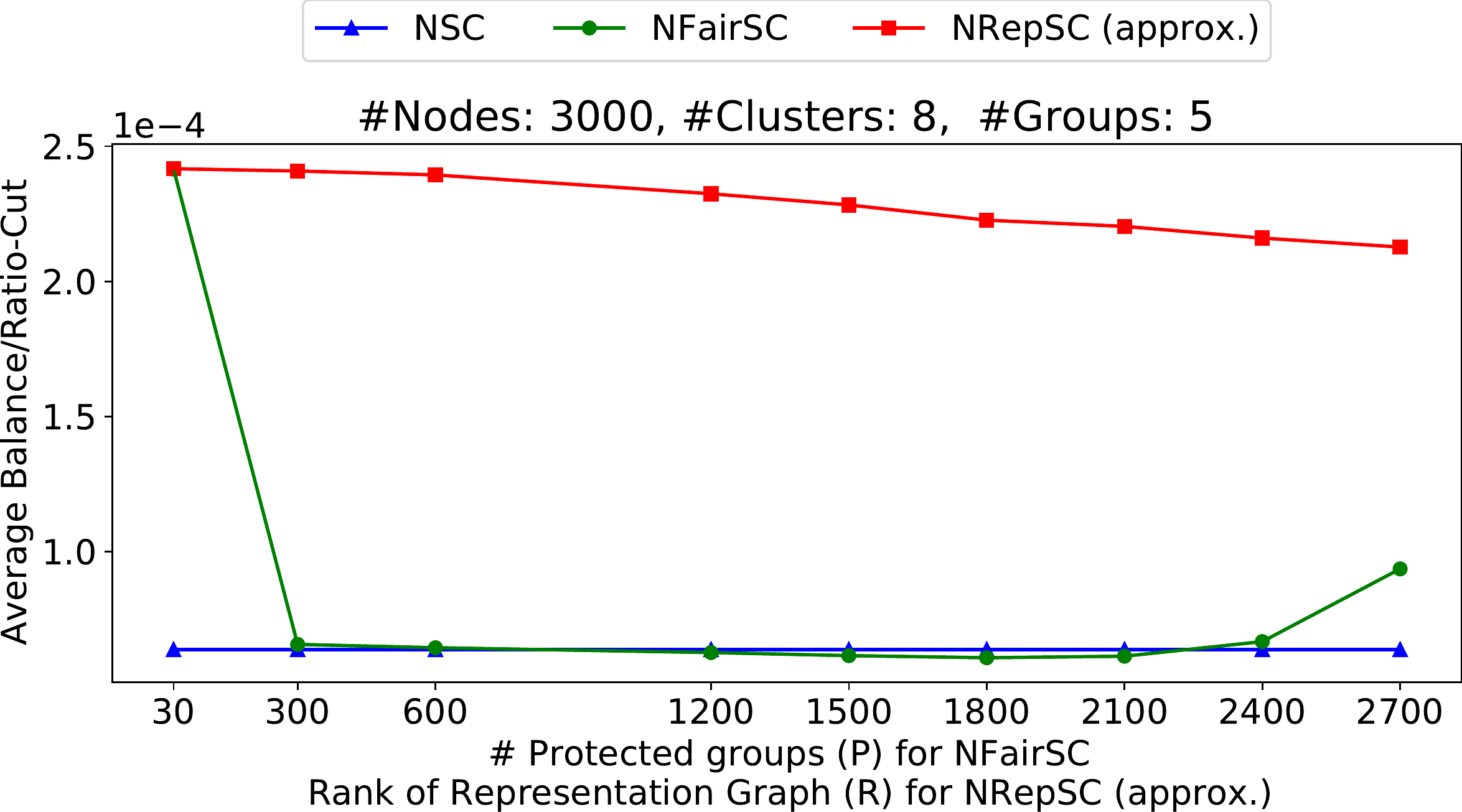}}
    \caption{Comparing \textsc{NRepSC (approx.)} with \textsc{NFairSC} using synthetically generated representation graphs sampled from an SBM.}
    \label{fig:sbm_comparison_norm}
\end{figure}

\subsection{Experiments with representation graphs sampled from SBM}
\label{chapter:fairness:section:sbm_experiments}

In this case, we divide the nodes into $P = 5$ protected groups and sample a representation graph $\calR$ using a Stochastic Block Model. Nodes in $\calR$ are connected with probability $p_{\mathrm{in}} = 0.8$ (resp. $p_{\mathrm{out}} = 0.2$) if they belong to the same (resp. different) protected group(s). Conditioned on $\calR$, we then sample an adjacency matrix from $\calR$-SBM as before. As an $\calR$ generated this way may violate the rank assumption, we only experiment with the approximate variants of \textsc{URepSC} and \textsc{NRepSC} in this case. Moreover, as such an $\calR$ may not be $d$-regular, the notion of accuracy no longer conveys information about the representation awareness of an algorithm. Thus, we instead compute the individual balance $\rho_i$ with respect to each node, as defined in \eqref{eq:balance}.
Recall that $0 \leq \rho_i \leq 1$ and higher values indicate that the representatives of node $v_i$ are well spread out across clusters $\hat{\calC}_1$, \dots, $\hat{\calC}_K$. We use average balance $\bar{\rho} = \frac{1}{N} \sum_{i = 1}^N \rho_i$ to measure the representation-awareness of the clusters.

While average balance measures the representation awareness of the clusters, we also need to ensure that they have a high quality. Thus, we compute the ratio of the average balance to the ratio-cut objective. A high value indicates balanced clusters with a high quality (low ratio-cut score). Figure \ref{fig:sbm_comparison_unnorm} fixes the value of $P = 5$ and shows the variation of the metric described above on $y$-axis as a function of the number of protected groups used by \textsc{UFairSC} and rank $R$ used by \textsc{URepSC (approx.)}, for various values of $N$ and $K$. We used the same values of parameters $p$, $q$, $r$, and $s$, as in Section \ref{chapter:fairness:section:d_reg_experiments}. The plots in Figure \ref{fig:sbm_comparison_unnorm} show a trade-off between clustering accuracy and representation awareness. One can choose an appropriate value of $R$ and use \textsc{URepSC (approx.)} to get good quality clusters with a high balance. Figure \ref{fig:sbm_comparison_norm} presents analogous results for \textsc{NRepSC (approx.)}.

% ==================================================== %

\begin{figure}[t]
    \centering
    \subfloat[][$K = 2$]{\includegraphics[width=0.48\textwidth]{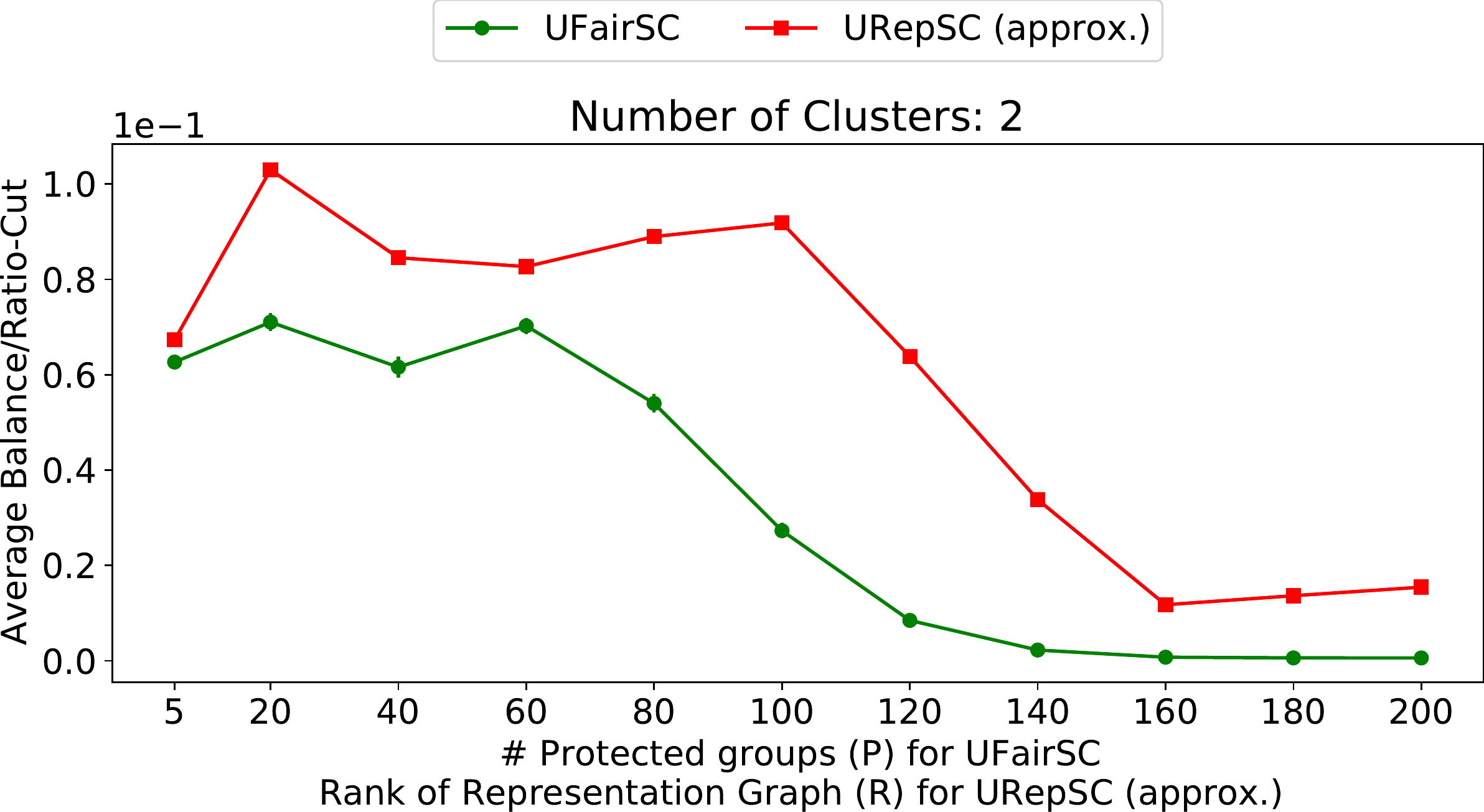}}%
    \hspace{0.5cm}\subfloat[][$K = 4$]{\includegraphics[width=0.48\textwidth]{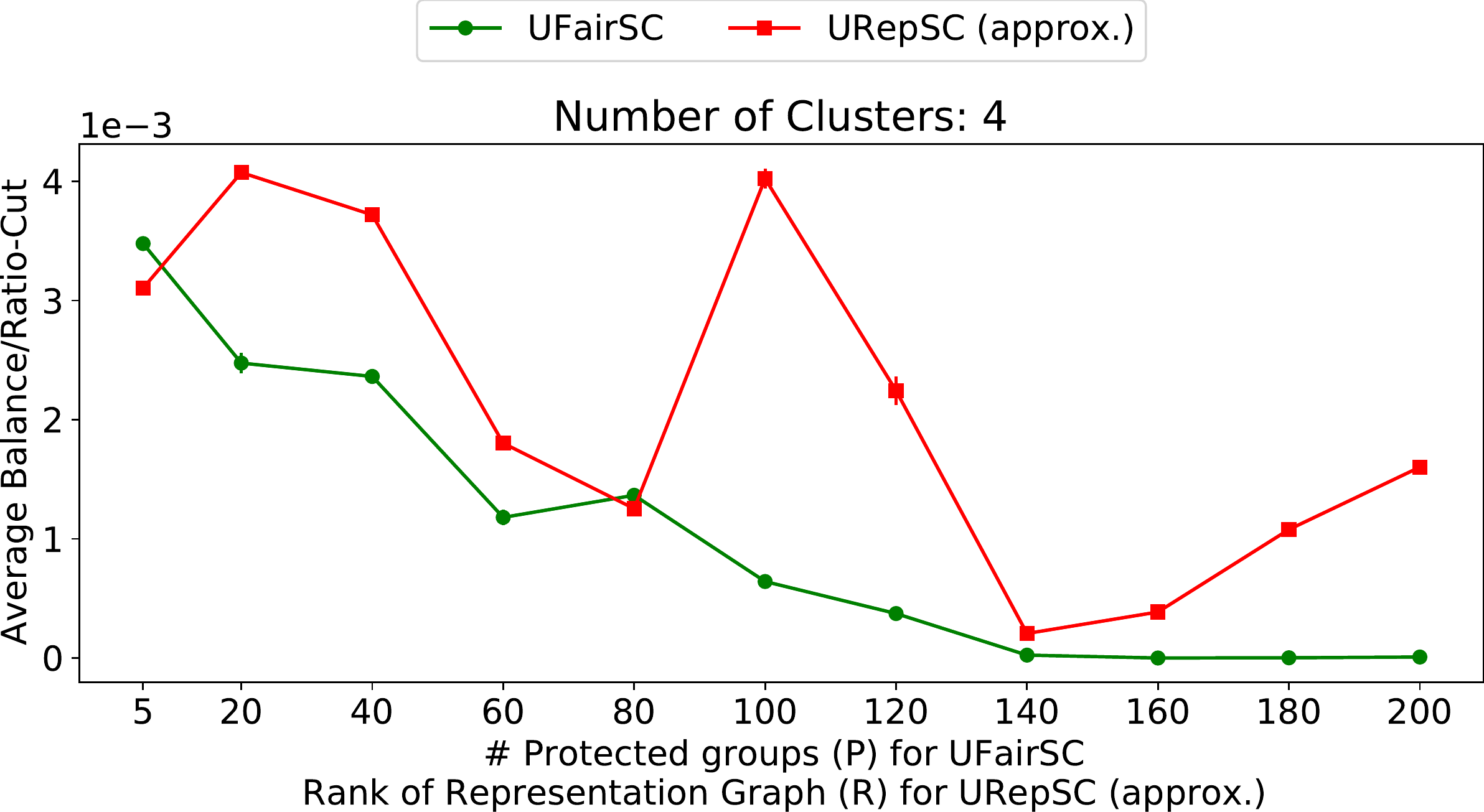}}

    \subfloat[][$K = 6$]{\includegraphics[width=0.48\textwidth]{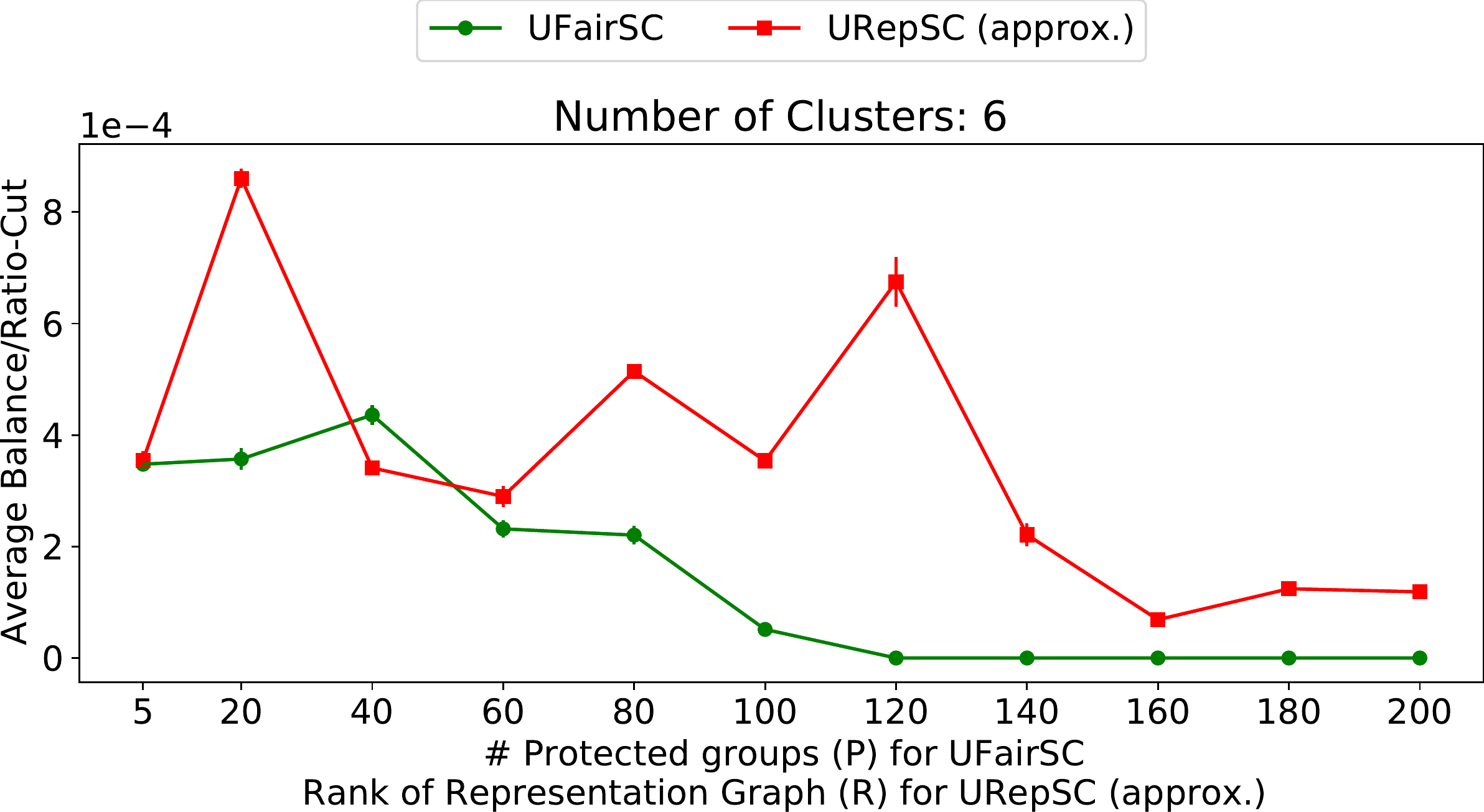}}%
    \hspace{0.5cm}\subfloat[][$K = 8$]{\includegraphics[width=0.48\textwidth]{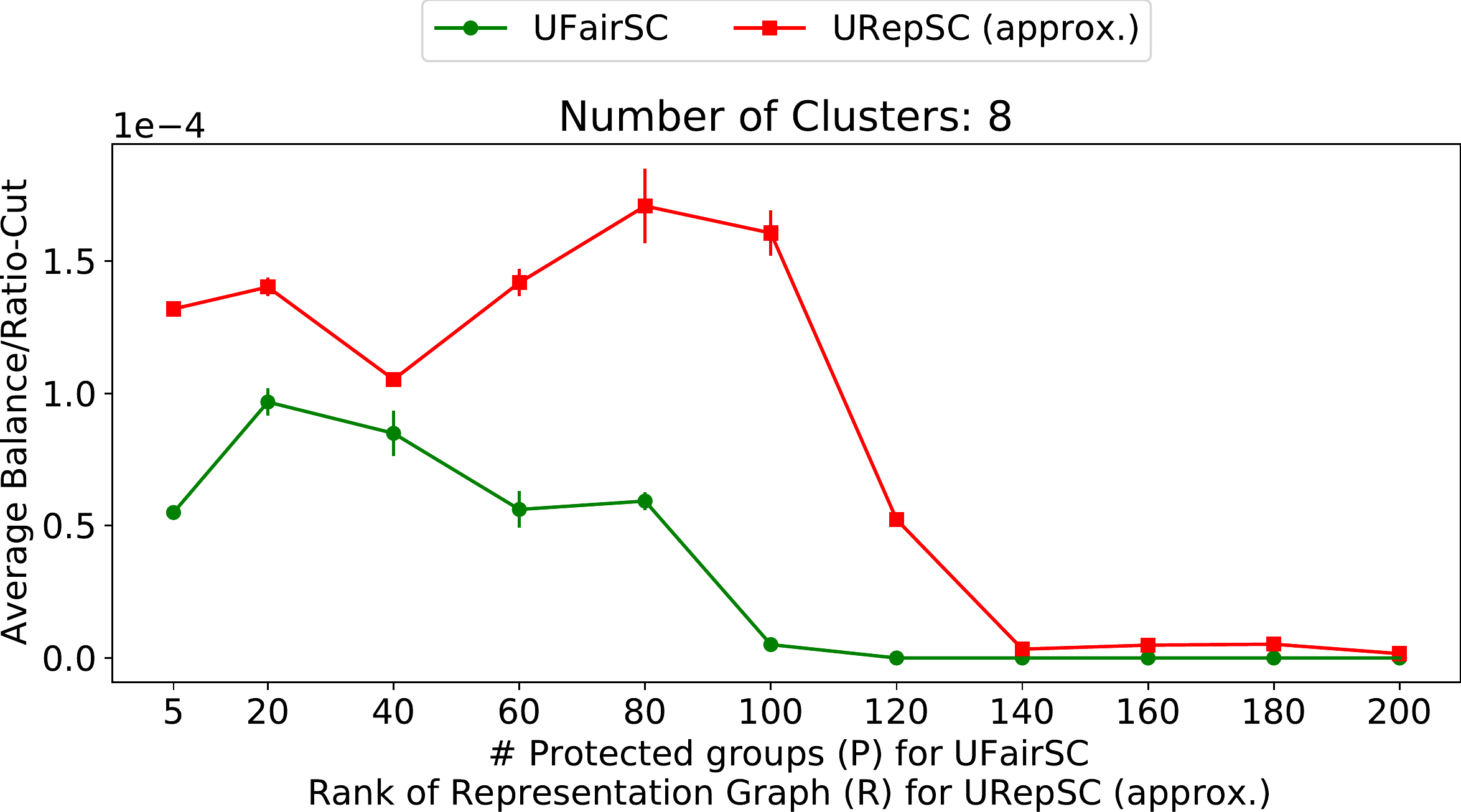}}
    \caption{Comparing \textsc{URepSC (approx.)} with \textsc{UFairSC} on FAO trade network.}
    \label{fig:real_data_comparison_unnorm}
\end{figure}

\begin{figure}[t]
    \centering
    \subfloat[][$K = 2$]{\includegraphics[width=0.48\textwidth]{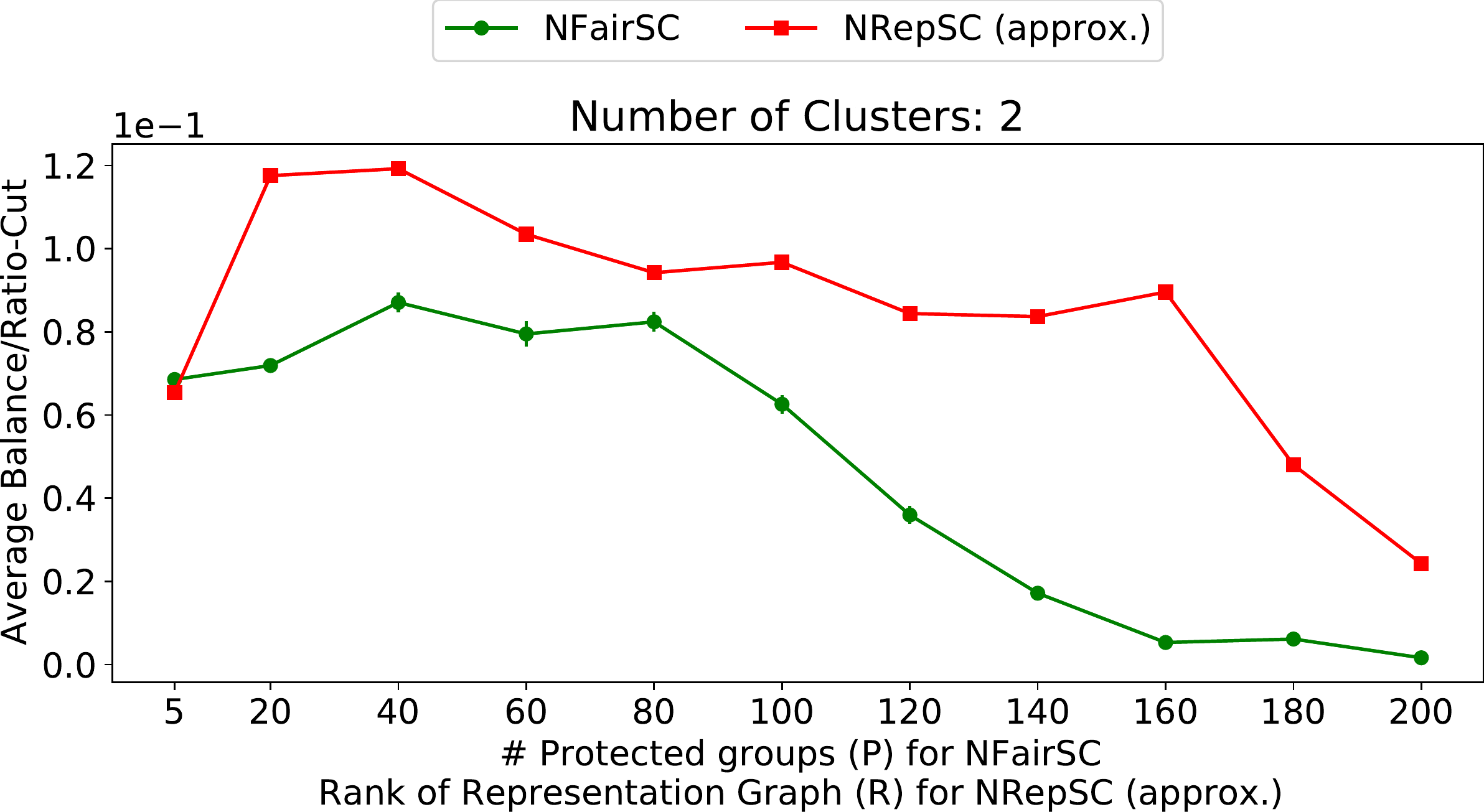}}%
    \hspace{0.5cm}\subfloat[][$K = 4$]{\includegraphics[width=0.48\textwidth]{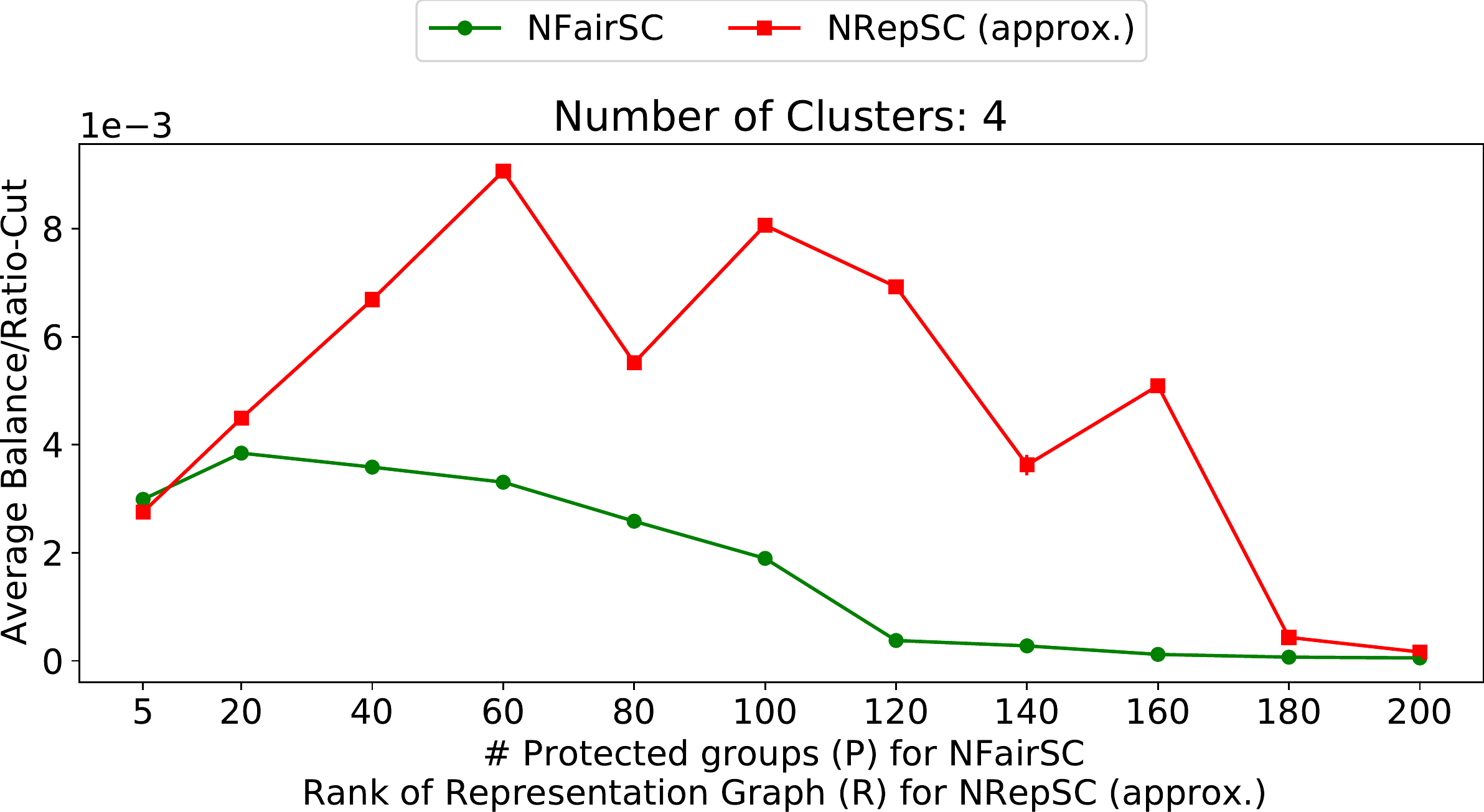}}

    \subfloat[][$K = 6$]{\includegraphics[width=0.48\textwidth]{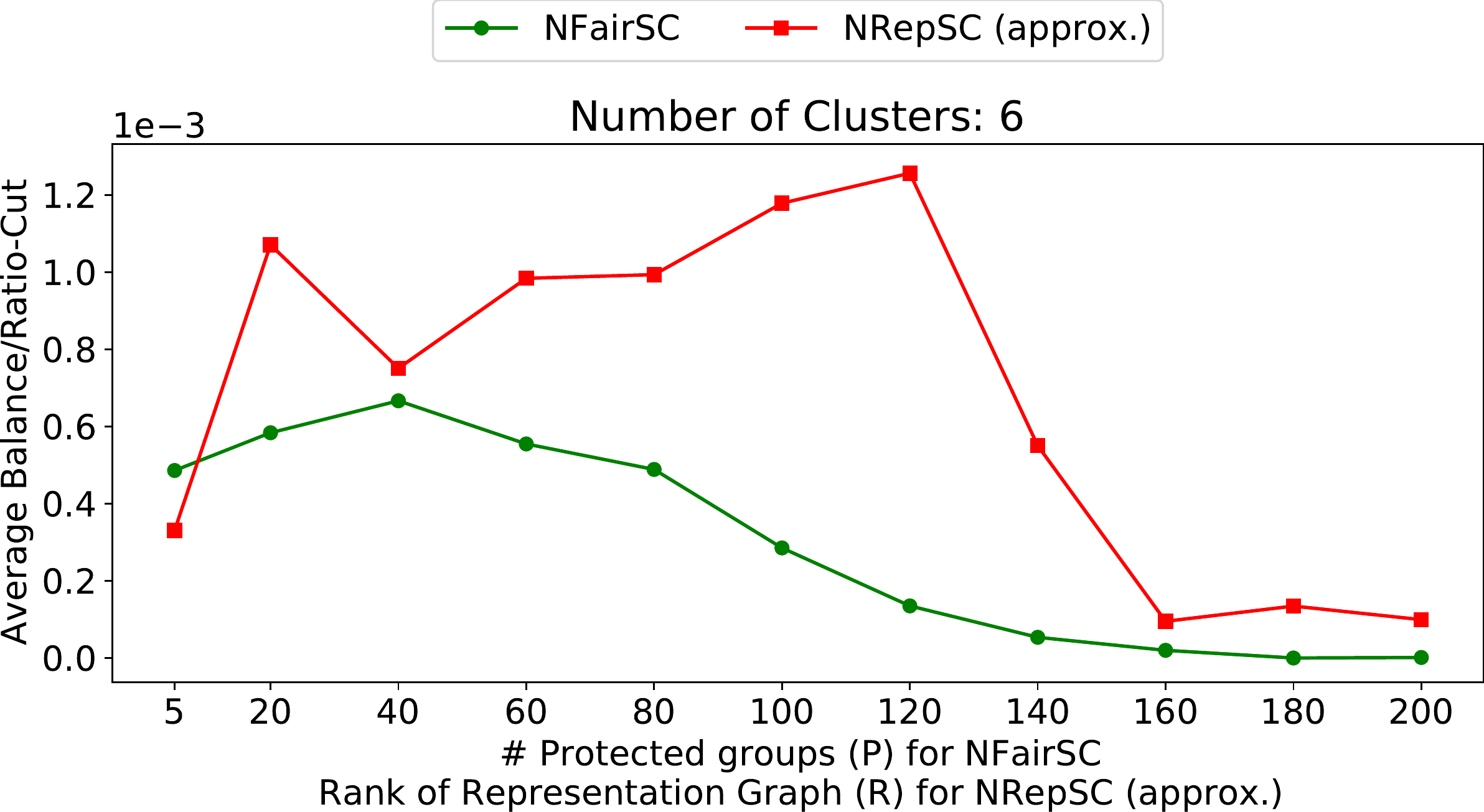}}%
    \hspace{0.5cm}\subfloat[][$K = 8$]{\includegraphics[width=0.48\textwidth]{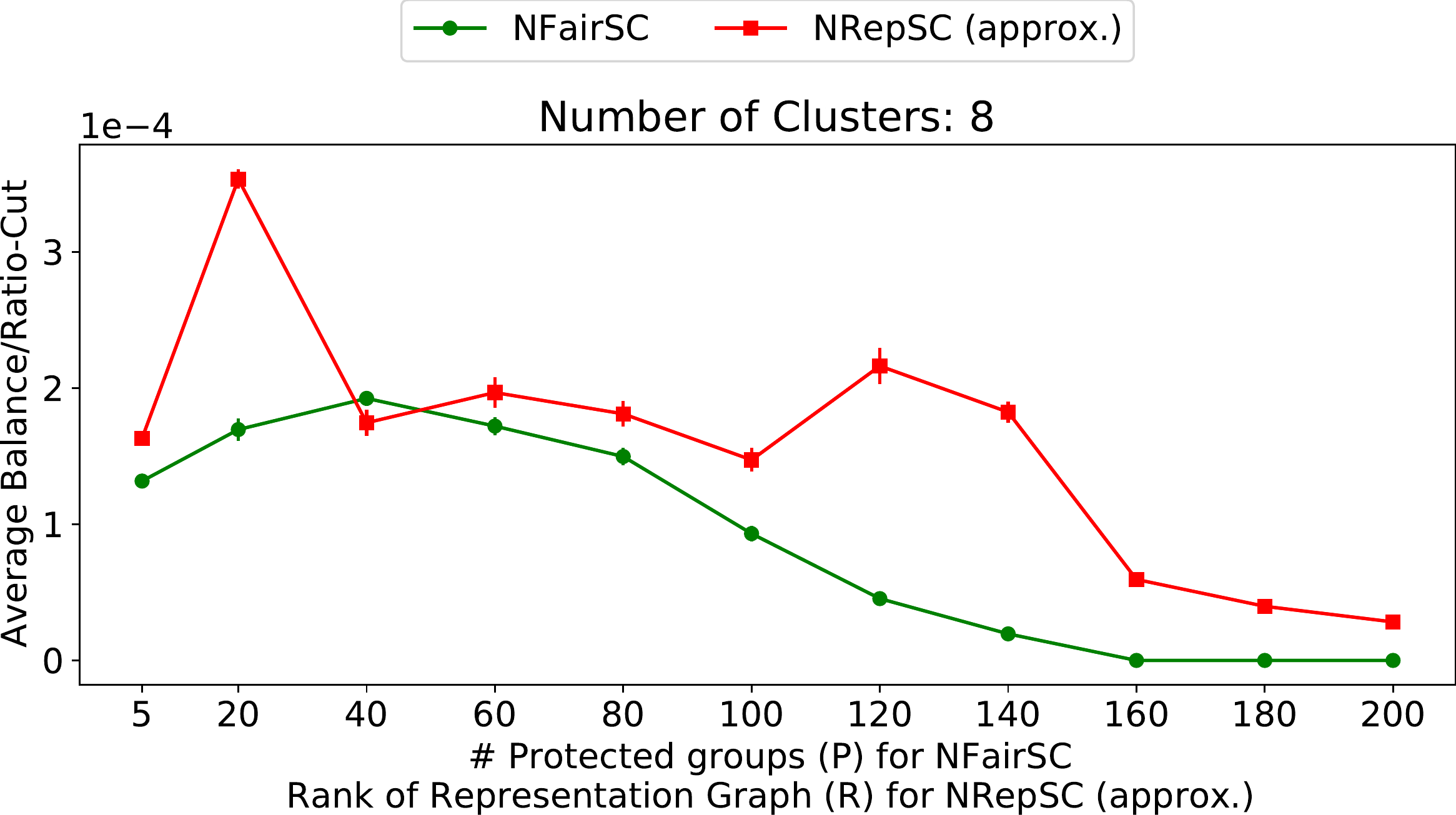}}
    \caption{Comparing \textsc{NRepSC (approx.)} with \textsc{NFairSC} on FAO trade network.}
    \label{fig:real_data_comparison_norm}
\end{figure}

\subsection{Experiments with a real-world network}
\label{chapter:fairness:section:trade_experiments}

For the final set of experiments, we use the FAO trade network \citep{DomenicoEtAl:2015:StructuralReducibilityOfMultilayerNetworks}, which is a multiplex network based on the data made available by the Food and Agriculture Organization (FAO) of the United Nations. It has $214$ nodes representing countries and $364$ layers corresponding to commodities like coffee, banana, barley, etc. An edge between two countries in a layer indicates the volume of the corresponding commodity traded between these countries. We convert the weighted graph in each layer to an unweighted graph by connecting every node with its five nearest neighbors. We then make all the edges undirected and use the first $182$ layers to construct the representation graph $\calR$. Nodes in $\calR$ are connected if they are linked in either of these layers. Similarly, the next $182$ layers are used to construct the similarity graph $\calG$. Note that $\calR$ constructed this way is not $d$-regular. The goal is to find clusters in $\calG$ that satisfy Definition \ref{def:representation_constraint} with respect to $\calR$.

To motivate this further, note that clusters based only on $\calG$ only consider the trade of commodities $183$--$364$. However, countries also have other trade relations in $\calR$, leading to shared economic interests. Assume that the members of each cluster would jointly formulate the economic policies for that cluster. However, the policies made in one cluster affect everyone, even if they are not part of the cluster, as they all share a global market. This incentivizes the countries to influence the economic policies of all the clusters. Being representation aware with respect to $\calR$ entails that each country has members in other clusters with shared interests. This enables a country to indirectly shape the policies of other clusters.

As before, we use the low-rank approximation for the representation graph in \textsc{URepSC (approx.)} and \textsc{NRepSC (approx.)}. Figure \ref{fig:real_data_comparison_unnorm} compares \textsc{URepSC (approx.)} with \textsc{UFairSC}, and has the same semantics as Figure \ref{fig:sbm_comparison_unnorm}. Different plots in Figure \ref{fig:real_data_comparison_unnorm} correspond to different choices of $K$. \textsc{URepSC (approx.)} achieves a higher ratio of average balance to ratio-cut. In practice, a user would choose $R$ by assessing the relative importance of a quality metric like ratio-cut and representation metric like average balance. Figure \ref{fig:real_data_comparison_norm} presents analogous results for \textsc{NRepSC (approx.)}.

%%%%%%%%%%%%%%%%%%%%%%%%%%%%%%%%%%%%%%%%%%%%%%

\section{Conclusion}
\label{section:conclusion}

The primary focus of this work has been on studying the consistency of constrained spectral clustering under an individual-level representation constraint. The proposed representation constraint naturally generalizes similar population-level constraints \citep{ChierichettiEtAl:2017:FairClusteringThroughFairlets} by using auxiliary information encoded in a representation graph $\calR$. We showed that the constraint can be expressed as a linear expression that when added to the optimization problem solved by spectral clustering results in the representation-aware variants of the algorithm. An interesting consequence of this problem is a variant of the stochastic block model that plants the properties of $\calR$ in a similarity graph in addition to the given clusters to provide a hard problem instance to our algorithms. Under this model, we derive a high-probability upper bound on the number of mistakes made by the algorithms and establish conditions under which they are weakly consistent. To the best of our knowledge, these are the first consistency results for constrained spectral clustering under individual-level constraints. Next, we make a few additional remarks.

\paragraph*{The $d$-regularity assumption} The $d$-regularity assumption on $\calR$ ensures the representation-awareness of the ground-truth clusters in our analysis. Note that the representation graph that recovers the statistical-level constraint is also $d$-regular (see Appendix \ref{appendix:constraint}), hence our analysis strictly generalizes the previously known results. It would be interesting to study the performance of our algorithms under weaker assumptions on $\calR$. One can also use a similar strategy as ours to modify more expressive variants of the stochastic block model, like the degree-corrected SBM \citep{KarrerNewman:2011:StochasticBlockmodelsAndCommunityStructureInNetworks}, to establish the consistency of the algorithms on more realistic similarity graphs.

\paragraph*{Computational complexity} Our current approach involves finding the null space of a $N \times N$ matrix. This operation has $O(N^3)$ complexity. Existing methods for speeding up the standard spectral clustering algorithm focus on making the eigen-decomposition and/or the $k$-means step faster (see \citet{TremblayEtAl:2016:CompressiveSpectralClustering} and the references within). However, even with these modifications, the null space computation would still be the computationally dominant step. \citet{XuEtAl:2009:FastNormalizedCutWithLinearConstraints} proposed an efficient algorithm for solving the normalized cut problem under a linear constraint. However, their algorithm assumes that $K = 2$. Developing similar algorithms for general values of $K$ and exploring their theoretical guarantees will enable the application domains that involve very large graphs to utilise our ideas.

% \paragraph*{Exploring other algorithms} The proposed representation constraint is a novel contribution. The representation graph encodes auxiliary information that may not affect the similarity computation. In contrast, existing individual-level constraints are defined only in terms of the similarity metric, for instance, by insisting on having a cluster centroid close to every data point \citep{MahabadiEtAl:2020:IndividualFairnessForKClustering}. While we focus only on spectral clustering in this paper, our proposed constraint is more broadly applicable. In fact, the feasibility issue in our current setting is a consequence of the desire to represent the constraint as a linear expression (from \eqref{eq:general_optimization_problem} to \eqref{eq:representation_constraint}). One can instead develop other algorithms that directly solve the more general optimization problem given in \eqref{eq:general_optimization_problem}. Alternatively, one can also solve an unconstrained optimization problem that penalizes the algorithm for having individuals with a low balance (defined in \eqref{eq:balance}).  

Other possible extensions of our work include similar algorithms for weighted similarity graphs, overlapping clusters, and other types of graphs such as hypergraphs. This paper provides the first step towards consistency analysis of spectral clustering under individual-level constraints.

%%%%%%%%%%%%%%%%%%%%%%%%%%%%%%%%%%%%%%%%%%%%%%
%% Multiple Appendixes:                     %%
%%%%%%%%%%%%%%%%%%%%%%%%%%%%%%%%%%%%%%%%%%%%%%
\begin{appendix}

\section{Representation constraint: Additional details}
\label{appendix:constraint}

%Here, we describe two example applications of the proposed constraint and highlight a few of its interesting properties.

%\paragraph*{Application - Fairness} As mentioned before, in the context of fairness, the representation graph specifies the ``trust'' relationship. A node $v_i$ finds the clusters fair if it has an adequate number of representatives in all clusters. The clusters are considered fair if all nodes find them fair. As fairness is defined from the perspective of each individual node, such a fairness notion is known as an \textit{individual fairness} notion.

%\paragraph*{Application - Load balancing} Consider an application where nodes in the similarity graph $\calG$ correspond to processes in cloud. Some of these processes are ``similar'' in that they operate on a common large dataset. The edges in $\calG$ encode similarity between processes from this perspective. It is desirable to keep similar processes on the same physical machine to improve efficiency. However, processes also compete with each other for shared resources like network, main memory, and cache. The edges in $\calR$ connect processes that compete for the same hardware resources. The objective is to cluster similar processes in $\calG$ together while ensuring that processes that are connected in $\calR$ are well spread out across clusters to minimize a collision.

In this section, we make two additional remarks about the properties of the proposed constraint, both in the context of fairness.

\paragraph*{Statistical fairness as a special case} Recall that our constraint specifies an individual fairness notion. Contrast this with several existing approaches that assign each node to one of the $P$ \textit{protected groups} $\calP_1, \dots, \calP_P \subseteq \calV$ \citep{ChierichettiEtAl:2017:FairClusteringThroughFairlets}, and require these protected groups to have a proportional representation in all clusters, i.e., 
\begin{equation*}
    \frac{\abs{\calP_i \cap \calC_j}}{\abs{\calC_j}} = \frac{\abs{\calP_i}}{N}, \,\, \forall i \in [P],\,\, j \in [K].
\end{equation*}
This is an example of \textit{statistical fairness}. In Example \ref{example:statistical_vs_individual_fairness}, we argued that statistical fairness may not be enough in some cases. We now show that the constraint in Definition \ref{def:representation_constraint} is equivalent to a statistical fairness notion for an appropriately constructed representation graph $\calR$ from the given protected groups $\calP_1, \dots, \calP_P$. Namely, let $\calR$ be such that $R_{ij} = 1$ if and only if $v_i$ and $v_j$ belong to the same protected group. In this case, it is easy to verify that the constraint in Definition \ref{def:representation_constraint} reduces to the statistical fairness criterion given above. In general, for other configurations of the representation graph, we strictly generalize the statistical fairness notion. We also strictly generalize the approach presented in \citet{KleindessnerEtAl:2019:GuaranteesForSpectralClusteringWithFairnessConstraints}, where the authors use spectral clustering to produce statistically fair clusters. Also noteworthy is the assumption made by statistical fairness, namely that every pair of vertices in a protected group can represent each others' interests ($R_{ij} = 1 \Leftrightarrow v_i$ and $v_j$ are in the same protected group), or they are very similar with respect to some sensitive attributes. This assumption becomes unreasonable as protected groups grow in size.

\paragraph*{Sensitive attributes and protected groups} Viewed as a fairness notion, the proposed constraint only requires a representation graph $\calR$. It has two advantages over existing fairness criteria: \textbf{(i)} it does not require observable sensitive attributes (such as age, gender, and sexual orientation), and \textbf{(ii)} even if sensitive attributes are provided, one need not specify the number of protected groups or explicitly compute them. This ensures data privacy and helps against individual profiling. Our constraint only requires access to the representation graph $\calR$. This graph can either be directly elicited from the individuals or derived as a function of several sensitive attributes. In either case, once $\calR$ is available, we no longer need to expose any sensitive attributes to the clustering algorithm. For example, individuals in $\calR$ may be connected if their age difference is less than five years and if they went to the same school. Crucially, the sensitive attributes used to construct $\calR$ may be numerical, binary, categorical, etc.

\end{appendix}

%% if your bibliography is in bibtex format, uncomment commands:
\bibliographystyle{plainnat}
\bibliography{bibliography}

\begin{thebibliography}{62}
\providecommand{\natexlab}[1]{#1}
\providecommand{\url}[1]{\texttt{#1}}
\expandafter\ifx\csname urlstyle\endcsname\relax
  \providecommand{\doi}[1]{doi: #1}\else
  \providecommand{\doi}{doi: \begingroup \urlstyle{rm}\Url}\fi

\bibitem[Abbe(2018)]{Abbe:2018:CommunityDetectionAndStochasticBlockModels}
Emmanuel Abbe.
\newblock Community detection and stochastic block models: Recent developments.
\newblock \emph{Journal of Machine Learning Research}, 18\penalty0
  (177):\penalty0 1--86, 2018.

\bibitem[Ahmadian et~al.(2017)Ahmadian, Norouzi-Fard, Svensson, and
  Ward]{AhmadianEtAl:2017:BetterGuaranteesForKMeansAndEuclideanKMedianByPrimalDualAlgorithms}
Sara Ahmadian, Ashkan Norouzi-Fard, Ola Svensson, and Justin Ward.
\newblock Better guarantees for k-means and euclidean k-median by primal-dual
  algorithms.
\newblock \emph{Symposium on Foundations of Computer Science}, 2017.

\bibitem[Ahmadian et~al.(2019)Ahmadian, Epasto, Kumar, and
  Mahdian]{AhmadianEtAl:2019:ClusteringWithoutOverRepresentation}
Sara Ahmadian, Alessandro Epasto, Ravi Kumar, and Mohammad Mahdian.
\newblock Clustering without over-representation.
\newblock \emph{In Proceedings of the 25th ACM SIGKDD International Conference
  on Knowledge Discovery \& Data Mining}, pages 267--275, 2019.

\bibitem[Anderson et~al.(2020)Anderson, Bera, Das, and
  Liu]{AndersonEtAl:2020:DistributionalIndividualFairnessInClustering}
Nihesh Anderson, Suman~K. Bera, Syamantak Das, and Yang Liu.
\newblock Distributional individual fairness in clustering.
\newblock \emph{arXiv}, 2006.12589, 2020.

\bibitem[Arthur and
  Vassilvitskii(2007)]{ArthurVassilvitskii:2007:KMeansTheAdvantagesOfCarefulSeeding}
David Arthur and Sergei Vassilvitskii.
\newblock k-means++: The advantages of careful seeding.
\newblock \emph{Symposium on Discrete Algorithms}, 2007.

\bibitem[Banerjee and
  Ghosh(2006)]{BanerjeeGhosh:2006:ScalableClusteringAlgorithmsWithBalancingConstraints}
Arindam Banerjee and Joydeep Ghosh.
\newblock Scalable clustering algorithms with balancing constraints.
\newblock \emph{Data Mining and Knowledge Discovery}, 13:\penalty0 365--395,
  2006.

\bibitem[Basu et~al.(2008)Basu, Davidson, and
  Wagstaff]{BasuEtAl:2008:ConstrainedClustering}
Sugato Basu, Ian Davidson, and Kiri Wagstaff.
\newblock \emph{Constrained clustering: Advances in algorithms, theory, and
  applications}.
\newblock CRC Press, 2008.

\bibitem[Bera et~al.(2019)Bera, Chakrabarty, Flores, and
  Negahbani]{BeraEtAl:2019:FairAlgorithmsForClustering}
Suman Bera, Deeparnab Chakrabarty, Nicolas Flores, and Maryam Negahbani.
\newblock Fair algorithms for clustering.
\newblock \emph{Advances in Neural Information Processing Systems},
  32:\penalty0 4954--4965, 2019.

\bibitem[Bercea et~al.(2019)Bercea, Gross, Khuller, Kumar, R\"{o}sner, Schmidt,
  and Schmidt]{BerceaEtAl:2019:OnTheCostOfEssentiallyFairClusterings}
Ioana~O. Bercea, Martin Gross, Samir Khuller, Aounon Kumar, Clemens R\"{o}sner,
  Daniel~R. Schmidt, and Melanie Schmidt.
\newblock On the cost of essentially fair clusterings.
\newblock \emph{APPROX-RANDOM}, pages 18:1--18:22, 2019.

\bibitem[Bie et~al.(2004)Bie, Suykens, and
  Moor]{BieEtAl:2004:LearningFromGeneralLabelConstraints}
Tijl~De Bie, Johan Suykens, and Bart~De Moor.
\newblock Learning from general label constraints.
\newblock \emph{Structural, Syntactic, and Statistical Pattern Recognition.
  Lecture Notes in Computer Science}, 3138, 2004.

\bibitem[Binkiewicz et~al.(2017)Binkiewicz, Vogelstein, and
  Rohe]{BinkiewiczEtAl:2017:CovariateAssistedSpectralClustering}
Norbert Binkiewicz, Joshua~T. Vogelstein, and Karl Rohe.
\newblock Covariate assisted spectral clustering.
\newblock \emph{Biometrika}, 104\penalty0 (2):\penalty0 361--377, 2017.

\bibitem[Chen et~al.(2019)Chen, Fain, Lyu, and
  Munagala]{ChenEtAl:2019:ProportionallyFairClustering}
Xingyu Chen, Brandon Fain, Liang Lyu, and Kamesh Munagala.
\newblock Proportionally fair clustering.
\newblock \emph{In Proceedings of the 36th International Conference on Machine
  Learning}, 97:\penalty0 1032--1041, 2019.

\bibitem[Chierichetti et~al.(2017)Chierichetti, Kumar, Lattanzi, and
  Vassilvitskii]{ChierichettiEtAl:2017:FairClusteringThroughFairlets}
Flavio Chierichetti, Ravi Kumar, Silvio Lattanzi, and Sergei Vassilvitskii.
\newblock Fair clustering through fairlets.
\newblock \emph{Advances in Neural Information Processing Systems},
  30:\penalty0 5029--5037, 2017.

\bibitem[Coleman et~al.(2008)Coleman, Saunderson, and
  Wirth]{ColemanEtAl:2008:SpectralClusteringWithInconsistentAdvice}
Tom Coleman, James Saunderson, and Anthony Wirth.
\newblock Spectral clustering with inconsistent advice.
\newblock \emph{In Proceedings of the 25th International Conference on Machine
  Learning}, pages 152--159, 2008.

\bibitem[Cucuringu et~al.(2016)Cucuringu, Koutis, Chawla, Miller, and
  Peng]{CucuringuEtAl:2016:SimpleAndScalableConstrainedClustering}
Mihai Cucuringu, Ioannis Koutis, Sanjay Chawla, Gary Miller, and Richard Peng.
\newblock Simple and scalable constrained clustering: A generalized spectral
  method.
\newblock \emph{Proceedings of the 19th International Conference on Artificial
  Intelligence and Statistics}, 41:\penalty0 445--454, 2016.

\bibitem[Davidson and Ravi(2005)]{DavidsonRavi:2005:ClusteringWithConstraints}
Ian Davidson and S.~S. Ravi.
\newblock Clustering with constraints: Feasibility issues and the k-means
  algorithm.
\newblock \emph{In Proceedings of the 5th SIAM Data Mining Conference}, pages
  138--149, 2005.

\bibitem[Demiriz et~al.(2008)Demiriz, Bennett, and
  Bradley]{DemirizEtAl:2008:UsingAssignmentConstraintsToAvoidEmptyClustersInKMeansClustering}
Ayhan Demiriz, Kristin Bennett, and P.S. Bradley.
\newblock \emph{Constrained Clustering: Advances in Algorithms, Theory, and
  Applications}, chapter Using assignment constraints to avoid empty clusters
  in k-means clustering, pages 201--220.
\newblock 2008.

\bibitem[Domenico et~al.(2015)Domenico, Nicosia, Arenas, and
  Latora]{DomenicoEtAl:2015:StructuralReducibilityOfMultilayerNetworks}
Manlio~De Domenico, Vincenzo Nicosia, Alexandre Arenas, and Vito Latora.
\newblock Structural reducibility of multilayer networks.
\newblock \emph{Nature Communications}, 6\penalty0 (1), 2015.

\bibitem[Dwork et~al.(2012)Dwork, Hardt, Pitassi, Reingold, and
  Zemel]{DworkEtAl:2012:FairnessThroughAwareness}
Cynthia Dwork, Moritz Hardt, Toniann Pitassi, Omer Reingold, and Richard Zemel.
\newblock Fairness through awareness.
\newblock \emph{ITCS {'}12}, pages 214–--226, 2012.

\bibitem[Eriksson et~al.(2011)Eriksson, Olsson, and
  Kahl]{ErikssonEtAl:2011:NormalizedCutsRevisited}
Anders Eriksson, Carl Olsson, and Fredrik Kahl.
\newblock Normalized cuts revisited: A reformulation for segmentation with
  linear grouping constraints.
\newblock \emph{Journal of Mathematical Imaging and Vision}, 39:\penalty0
  45--61, 2011.

\bibitem[Gao et~al.(2017)Gao, Ma, Zhang, and
  Zhou]{GaoEtAl:2017:AchievingOptimalMisclassificationProportionInStochasticBlockModels}
Chao Gao, Zongming Ma, Anderson~Y. Zhang, and Harrison~H. Zhou.
\newblock Achieving optimal misclassification proportion in stochastic block
  models.
\newblock \emph{Journal of Machine Learning Research}, 18:\penalty0 1--45,
  2017.

\bibitem[Ghoshdastidar and
  Dukkipati(2017{\natexlab{a}})]{GhoshdastidarDukkipati:2017:ConsistencyOfSpectralHypergraphPartitioningUnderPlantedPartitionModel}
Debarghya Ghoshdastidar and Ambedkar Dukkipati.
\newblock Consistency of spectral hypergraph partitioning under planted
  partition model.
\newblock \emph{Annals of Statistics}, 45\penalty0 (1):\penalty0 289--315,
  2017{\natexlab{a}}.

\bibitem[Ghoshdastidar and
  Dukkipati(2017{\natexlab{b}})]{GhoshdastidarDukkipati:2017:UniformHypergraphPartitioning}
Debarghya Ghoshdastidar and Ambedkar Dukkipati.
\newblock Uniform hypergraph partitioning: Provable tensor methods and sampling
  techniques.
\newblock \emph{Journal of Machine Learning Research}, 18\penalty0
  (1):\penalty0 1638--1678, 2017{\natexlab{b}}.

\bibitem[Gupta and Dukkipati(2021{\natexlab{a}})]{ThisPaperArXiv}
Shubham Gupta and Ambedkar Dukkipati.
\newblock Protecting individual interests across clusters: Spectral clustering
  with guarantees.
\newblock \emph{arXiv}, 2105.03714, 2021{\natexlab{a}}.

\bibitem[Gupta and Dukkipati(2021{\natexlab{b}})]{ThisPaperSupp}
Shubham Gupta and Ambedkar Dukkipati.
\newblock Supplement to ``on consistency of constrained spectral clustering
  under representation-aware stochastic block model''.
\newblock 2021{\natexlab{b}}.

\bibitem[Hofmann and
  Buhmann(1997)]{HofmannBuhmann:1997:PairwiseDataClusteringByDeterministicAnnealing}
Thomas Hofmann and Joachim~M. Buhmann.
\newblock Pairwise data clustering by deterministic annealing.
\newblock \emph{IEEE Transactions on Pattern Analysis and Machine
  Intelligence}, 19\penalty0 (1):\penalty0 1--14, 1997.

\bibitem[Holland et~al.(1983)Holland, Laskey, and
  Leinhardt]{HollandEtAl:1983:StochasticBlockmodelsFirstSteps}
Paul~W. Holland, Kathryn~Blackmond Laskey, and Samuel Leinhardt.
\newblock Stochastic blockmodels: First steps.
\newblock \emph{Social Networks}, 5\penalty0 (2):\penalty0 109--137, 1983.

\bibitem[Kamvar et~al.(2003)Kamvar, Klein, and
  Manning]{KamvarEtAl:2003:SpectralLearning}
Sepandar~D. Kamvar, Dan Klein, and Christopher~D. Manning.
\newblock Spectral learning.
\newblock \emph{In Proceedings of the 18th International Joint Conference on
  Artificial Intelligence}, pages 561--566, 2003.

\bibitem[Karrer and
  Newman(2011)]{KarrerNewman:2011:StochasticBlockmodelsAndCommunityStructureInNetworks}
Brian Karrer and Mark Newman.
\newblock Stochastic blockmodels and community structure in networks.
\newblock \emph{Physical Review E}, 83\penalty0 (1):\penalty0 016107, 2011.

\bibitem[Kawale and
  Boley(2013)]{KawaleBoley:2013:ConstrainedSpectralClusteringUsingL1Regularization}
Jaya Kawale and Daniel Boley.
\newblock Constrained spectral clustering using l1 regularization.
\newblock \emph{In Proceedings of the International Conference on Data Mining},
  pages 103--111, 2013.

\bibitem[Kleindessner et~al.(2019)Kleindessner, Samadi, Awasthi, and
  Morgenstern]{KleindessnerEtAl:2019:GuaranteesForSpectralClusteringWithFairnessConstraints}
Matth\"{a}us Kleindessner, Samira Samadi, Pranjal Awasthi, and Jamie
  Morgenstern.
\newblock Guarantees for spectral clustering with fairness constraints.
\newblock \emph{In Proceedings of the 36th International Conference on Machine
  Learning}, 97:\penalty0 3458--3467, 2019.

\bibitem[Kumar et~al.(2004)Kumar, Sabharwal, and
  Sen]{KumarEtAl:2004:ASimpleLinearTimeApproximateAlgorithmForKMeansClusteringInAnyDimension}
Amit Kumar, Yogish Sabharwal, and Sandeep Sen.
\newblock A simple linear time (1 + $\epsilon$)-approximation algorithm for
  k-means clustering in any dimensions.
\newblock \emph{Symposium on Foundations of Computer Science}, 2004.

\bibitem[Lei and
  Rinaldo(2015)]{LeiEtAl:2015:ConsistencyOfSpectralClusteringInSBM}
Jing Lei and Alessandro Rinaldo.
\newblock Consistency of spectral clustering in stochastic block models.
\newblock \emph{The Annals of Statistics}, 43\penalty0 (1):\penalty0 215--237,
  2015.

\bibitem[Lei and
  Zhu(2017)]{LeiZhu:2017:AGenericSampleSplittingApproachForRefinedCommunityRecoveryInSBMs}
Jing Lei and Lingxue Zhu.
\newblock A generic sample splitting approach for refined community recovery in
  stochastic block models.
\newblock \emph{Statistica Sinica}, 27\penalty0 (4):\penalty0 1639--1659, 2017.

\bibitem[Li et~al.(2009)Li, Liu, and
  Tang]{LiEtAl:2009:ConstrainedClusteringViaSpectralRegularization}
Zhenguo Li, Jianzhuang Liu, and Xiaoou Tang.
\newblock Constrained clustering via spectral regularization.
\newblock \emph{In Proceedings of the IEEE Conference on Computer Vision and
  Pattern Recognition}, pages 421--428, 2009.

\bibitem[Lloyd(1982)]{Lloyd:1982:LeastSquaresQuantisationInPCM}
Stuart~P. Lloyd.
\newblock Least squares quantisation in pcm.
\newblock \emph{IEEE Transactions on Information Theory}, 28\penalty0
  (2):\penalty0 129--137, 1982.

\bibitem[Lu and
  Carreira-Perpin\'{a}n(2008)]{LuCarreiraPerpinan:2008:ConstrainedSpectralClusteringThroughAffinityPropagation}
Zhengdong Lu and Miguel~\'{A}. Carreira-Perpin\'{a}n.
\newblock Constrained spectral clustering through affinity propagation.
\newblock \emph{In Proceedings of the IEEE Conference on Computer Vision and
  Pattern Recognition}, 2008.

\bibitem[L\"{u}tkepohl(1996)]{Lutkepohl:1996:HandbookOfMatrices}
Helmut L\"{u}tkepohl.
\newblock \emph{Handbook of matrices}.
\newblock Wiley, 1996.

\bibitem[Mahabadi and
  Vakilian(2020)]{MahabadiEtAl:2020:IndividualFairnessForKClustering}
Sepideh Mahabadi and Ali Vakilian.
\newblock Individual fairness for k-clustering.
\newblock \emph{In Proceedings of the 37th International Conference on Machine
  Learning}, 119:\penalty0 6586--6596, 2020.

\bibitem[Ng et~al.(2001)Ng, Jordan, and
  Weiss]{NgEtAl:2001:OnSpectralClustering}
Andrew Ng, Michael Jordan, and Yair Weiss.
\newblock On spectral clustering: Analysis and an algorithm.
\newblock \emph{Advances in Neural Information Processing Systems}, 14, 2001.

\bibitem[Rangapuram and
  Hein(2012)]{RangapuramHein:2012:Constrained1SpectralClustering}
Syama~S. Rangapuram and Matthias Hein.
\newblock Constrained 1-spectral clustering.
\newblock \emph{In Proceedings of the 15th International Conference on
  Artificial Intelligence and Statistics}, 22:\penalty0 1143--1151, 2012.

\bibitem[Rohe et~al.(2011)Rohe, Chatterjee, and
  Yu]{RoheEtAl:2011:SpectralClusteringAndTheHighDimensionalSBM}
Karl Rohe, Sourav Chatterjee, and Bin Yu.
\newblock Spectral clustering and the high-dimensional stochastic blockmodel.
\newblock \emph{The Annals of Statistics}, 39\penalty0 (4):\penalty0
  1878--1915, 2011.

\bibitem[R\"{o}sner and
  Schmidt(2018)]{RosnerSchmidt:2018:PrivacyPreservingClusteringWithConstraints}
Clemens R\"{o}sner and Melanie Schmidt.
\newblock Privacy preserving clustering with constraints.
\newblock \emph{ICALP}, 2018.

\bibitem[Schmidt et~al.(2018)Schmidt, Schwiegelshohn, and
  Sohler]{SchmidtEtAl:2018:FairCoresetsAndStreamingAlgorithmsForFairKMeansClustering}
Melanie Schmidt, Chris Schwiegelshohn, and Christian Sohler.
\newblock Fair coresets and streaming algorithms for fair k-means clustering.
\newblock \emph{arXiv}, 1812.10854, 2018.

\bibitem[Shental et~al.(2003)Shental, Bar-Hillel, Hertz, and
  Weinshall]{ShentalEtAl:2003:ComputingGaussianMixtureModelsWithEMUsingEquivalenceConstraints}
Noam Shental, Aharon Bar-Hillel, Tomer Hertz, and Daphna Weinshall.
\newblock Computing gaussian mixture models with em using equivalence
  constraints.
\newblock \emph{In Proceedings of the 16th International Conference on Neural
  Information Processing Systems}, pages 465--472, 2003.

\bibitem[Shi and Malik(2000)]{ShiMalik:2000:NormalizedCutsAndImageSegmentation}
Jianbo Shi and Jitendra Malik.
\newblock Normalized cuts and image segmentation.
\newblock \emph{IEEE Transactions on Pattern Analysis and Machine
  Intelligence}, 22\penalty0 (8):\penalty0 888--905, 2000.

\bibitem[Tremblay et~al.(2016)Tremblay, Puy, Gribonval, and
  Vandergheynst]{TremblayEtAl:2016:CompressiveSpectralClustering}
Nicolas Tremblay, Gilles Puy, Remi Gribonval, and Pierre Vandergheynst.
\newblock Compressive spectral clustering.
\newblock \emph{In Proceedings of The 33rd International Conference on Machine
  Learning}, 48:\penalty0 1002--1011, 2016.

\bibitem[von Luxburg(2007)]{Luxburg:2007:ATutorialOnSpectralClustering}
Ulrike von Luxburg.
\newblock A tutorial on spectral clustering.
\newblock \emph{Statistics and Computing}, 17\penalty0 (4):\penalty0 395--416,
  2007.

\bibitem[von Luxburg et~al.(2008)von Luxburg, Belkin, and
  Bousquet]{LuxburgEtAl:2008:ConsistencyOfSpectralClustering}
Ulrike von Luxburg, Mikhail Belkin, and Olivier Bousquet.
\newblock Consistency of spectral clustering.
\newblock \emph{The Annals of Statistics}, 36\penalty0 (2):\penalty0 555--586,
  2008.

\bibitem[Vu(2018)]{VuEtAl:2018:ASimpleSVDAlgorithmForFindingHiddenPartitions}
Van Vu.
\newblock A simple svd algorithm for finding hidden partitions.
\newblock \emph{Combinatorics, Probability and Computing}, 27\penalty0
  (1):\penalty0 124--140, 2018.

\bibitem[Vu and
  Lei(2013)]{VuLei:2013:MinimaxSparsePrincipalSubspaceEstimationInHighDimensions}
Vincent~Q. Vu and Jing Lei.
\newblock Minimax sparse principal subspace estimation in high dimensions.
\newblock \emph{The Annals of Statistics}, 41\penalty0 (6):\penalty0
  2905--2947, 2013.

\bibitem[Wagner and
  Wagner(1993)]{WagnerWagner:1993:BetweenMinCutAndGraphBisection}
Dorothea Wagner and Frank Wagner.
\newblock Between min cut and graph bisection.
\newblock \emph{Mathematical Foundations of Computer Science}, 711:\penalty0
  744--750, 1993.

\bibitem[Wagstaff et~al.(2001)Wagstaff, Cardie, Rogers, and
  Schr\H{o}dl]{WagstaffEtAl:2001:ConstrainedKMeansClusteringWithBackgroundKnowledge}
Kiri Wagstaff, Claire Cardie, Seth Rogers, and Stefan Schr\H{o}dl.
\newblock Constrained k-means clustering with background knowledge.
\newblock \emph{In Proceedings of the 18th International Conference on Machine
  Learning}, pages 577--584, 2001.

\bibitem[Wang et~al.(2009)Wang, Ding, and
  Li]{WangEtAl:2009:IntegratedKLClustering}
Fei Wang, Chris Ding, and Tao Li.
\newblock Integrated kl (k-means- laplacian) clustering: A new clustering
  approach by combining attribute data and pairwise relations.
\newblock \emph{In Proceedings of the SIAM International Conference on Data
  Mining}, pages 38--48, 2009.

\bibitem[Wang and
  Davidson(2010{\natexlab{a}})]{WangDavidson:2010:ActiveSpectralClustering}
Xiang Wang and Ian Davidson.
\newblock Active spectral clustering.
\newblock \emph{In Proceedings of the IEEE International Conference on Data
  Mining}, pages 561--568, 2010{\natexlab{a}}.

\bibitem[Wang and
  Davidson(2010{\natexlab{b}})]{WangDavidson:2010:FlexibleConstrainedSpectralClustering}
Xiang Wang and Ian Davidson.
\newblock Flexible constrained spectral clustering.
\newblock \emph{In Proceedings of the 16th ACM SIGKDD International Conference
  on Knowledge Discovery and Data Mining}, pages 563--572, 2010{\natexlab{b}}.

\bibitem[Wang et~al.(2014)Wang, Qian, and
  Davidson]{WangEtAl:2014:OnConstrainedSpectralClusteringAndItsApplications}
Xiang Wang, Buyue Qian, and Ian Davidson.
\newblock On constrained spectral clustering and its applications.
\newblock \emph{Data Mining and Knowledge Discovery}, 28:\penalty0 1--30, 2014.

\bibitem[Xu et~al.(2009)Xu, Li, and
  Schuurmans]{XuEtAl:2009:FastNormalizedCutWithLinearConstraints}
Linli Xu, Wenye Li, and Dale Schuurmans.
\newblock Fast normalized cut with linear constraints.
\newblock \emph{IEEE Conference on Computer Vision and Pattern Recognition},
  2009.

\bibitem[Yu and Shi(2001)]{YuShi:2001:GroupingWithBias}
Stella~X. Yu and Jianbo Shi.
\newblock Grouping with bias.
\newblock \emph{Advances in Neural Information Processing Systems},
  14:\penalty0 1327--1334, 2001.

\bibitem[Yu and
  Shi(2004)]{YuShi:2004:SegmentationGivenPartialGroupingConstraints}
Stella~X. Yu and Jianbo Shi.
\newblock Segmentation given partial grouping constraints.
\newblock \emph{IEEE Transactions on Pattern Analysis and Machine
  Intelligence}, 26\penalty0 (2):\penalty0 173--183, 2004.

\bibitem[Zhang et~al.(2014)Zhang, Levina, and
  Zhu]{ZhangEtAl:2014:DetectingOverlappingCommunitiesInNetworksUsingSpectralMethods}
Yuan Zhang, Elizaveta Levina, and Ji~Zhu.
\newblock Detecting overlapping communities in networks using spectral methods.
\newblock \emph{SIAM Journal on Mathematics of Data Science}, 2\penalty0
  (2):\penalty0 265--283, 2014.

\bibitem[Zhu et~al.(2013)Zhu, Loy, and
  Gong]{ZhuEtAl:2013:ConstrainedClustering}
Xiatian Zhu, Chen~Change Loy, and Shaogang Gong.
\newblock Constrained clustering: Effective constraint propagation with
  imperfect oracles.
\newblock \emph{In Proceedings of the 13th International Conference on Data
  Mining}, pages 1307--1312, 2013.

\end{thebibliography}

\newpage
\setcounter{page}{1}

{
\centering
\Large{Supplement to} \\
\Large{\textbf{``On Consistency of constrained spectral clustering under Representation-Aware Stochastic Block Model''}} \\
\vspace{5mm}
\Large{Shubham Gupta and Ambedkar Dukkipati} \\
}

\normalsize

%%%%%%%%%%%%%%%%%%%%%%%%%%%%%%%%%%%%%%%%%%%%%%

\setcounter{section}{0}
\def\thesection{\arabic{section}}

\section{Proof of technical lemmas from Section \ref{section:algorithms}}
\label{appendix:proof_of_technical_lemmas_from_algorithms}

%%%%%%%%%%%%%%%%%%%%%%%%%%%%%%%%%%%%%%%%%%%%%%

\subsection*{Proof of Lemma \ref{lemma:constraint_matrix_unnorm}}
Fix an arbitrary node $v_i \in \calV$ and $k \in [K]$. Because $\bfR(\bfI - \bmone \bmone^\intercal / N) \bfH = \mathbf{0}$,
\begin{align*}
    &\sum_{j=1}^N R_{ij} H_{jk} = \frac{1}{N} \left(\sum_{j=1}^N R_{ij} \right) \left(\sum_{j=1}^N H_{jk} \right) \\
    \Rightarrow& \,\, \frac{1}{\sqrt{\abs{\calC_k}}} \abs{\{v_j \in \calV : R_{ij} = 1 \land v_j \in \calC_k\}} = \frac{1}{N} \abs{\{v_j \in \calV : R_{ij} = 1\}} \frac{\abs{\calC_k}}{\sqrt{\abs{\calC_k}}} \\
    \Rightarrow&  \,\,\frac{\abs{\{v_j \in \calV : R_{ij} = 1 \land v_j \in \calC_k\}}}{\abs{\calC_k}} = \frac{\abs{\{v_j \in \calV : R_{ij} = 1\}}}{N}. 
\end{align*}
Because this holds for an arbitrary $v_i \in \calV$ and $k \in [K]$, $\bfR(\bfI - \bmone \bmone^\intercal / N) \bfH = \mathbf{0}$ implies the constraint in Definition \ref{def:representation_constraint}.

%%%%%%%%%%%%%%%%%%%%%%%%%%%%%%%%%%%%%%%%%%%%%%

\subsection*{Proof of Lemma \ref{lemma:constraint_matrix_norm}}
Fix an arbitrary node $v_i \in \calV$ and $k \in [K]$. Because $\bfR(\bfI - \bmone \bmone^\intercal / N) \bfT = 0$,
\begin{align*}
  &\sum_{j=1}^N R_{ij} T_{jk} = \frac{1}{N}  \left(\sum_{j=1}^N R_{ij} \right) \left(\sum_{j=1}^N T_{jk} \right) \\
  \Rightarrow& \,\, \frac{1}{\sqrt{\mathrm{Vol}(\calC_k)}} \abs{\{v_j \in \calV : R_{ij} = 1 \land v_j \in \calC_k\}} = \frac{1}{N} \abs{\{v_j \in \calV : R_{ij} = 1\}} \frac{\abs{\calC_k}}{\sqrt{\mathrm{Vol}(\calC_k)}} \\
  \Rightarrow&  \,\,\frac{\abs{\{v_j \in \calV : R_{ij} = 1 \land v_j \in \calC_k\}}}{\abs{\calC_k}} = \frac{\abs{\{v_j \in \calV : R_{ij} = 1\}}}{N}. 
\end{align*}
Here, recall that $\mathrm{Vol}(\calC_k) = \sum_{v_i \in \calC_k} D_{ii}$ is the volume of the cluster $\calC_k$, which is used in \eqref{eq:T_def}. Because this holds for an arbitrary $v_i \in \calV$ and $k \in [K]$, $\bfR(\bfI - \bmone \bmone^\intercal / N) \bfT = \mathbf{0}$ implies the constraint in Definition \ref{def:representation_constraint}.

%%%%%%%%%%%%%%%%%%%%%%%%%%%%%%%%%%%%%%%%%%%%%%

\section{Proof of technical lemmas from Section \ref{section:analysis}}
\label{appendix:technical_lemmas_consistency}

% ===================================== %

\subsection{Proof of Lemma \ref{lemma:introducing_uks}}

Because $\calR$ is a $d$-regular graph, it is easy to see that $\bfR \bmone = d \bmone$. Recall from Section \ref{section:consistency_results} that $\bfI - \frac{1}{N}\bmone \bmone^\intercal$ is a projection matrix that removes the component of any vector $\bfx \in \bbR^N$ along the all ones vector $\bmone$. Thus, $(\bfI - \frac{1}{N}\bmone \bmone^\intercal) \bmone = 0$ and hence $\bmone \in \nullspace{\bfR(\bfI - \frac{1}{N}\bmone \bmone^\intercal)}$. Moreover, as all clusters have the same size, $$\bmone^\intercal \bfu_k = \sum_{i = 1}^N u_{ki} = \sum_{i \,:\, v_i \in \calC_k} u_{ki} + \sum_{i \,:\, v_i \notin \calC_k} u_{ki} = \frac{N}{K} - \frac{1}{K - 1} \left(N - \frac{N}{K}\right) = 0.$$

Thus, $\bfR(\bfI - \frac{1}{N} \bmone \bmone^\intercal) \bfu_k = \bfR \bfu_k$. Let us compute the $i^{th}$ element of the vector $\bfR\bfu_k$ for an arbitrary $i \in [N]$. $$(\bfR\bfu_k)_i = \sum_{j = 1}^N R_{ij} u_{kj} = \sum_{\substack{j \,:\, R_{ij} = 1 \\ \& v_j \in \calC_k}} 1 - \sum_{\substack{j \,:\, R_{ij} = 1 \\ \& v_j \notin \calC_k}} \frac{1}{K - 1} = \frac{d}{K} - \frac{1}{K - 1}\left(d - \frac{d}{K}\right) = 0.$$
Here, the second last equality follows from Assumption \ref{assumption:R_is_d_regular} and the assumption that all clusters have the same size. Thus, $\bfR\bfu_k = 0$ and hence $\bfu_k \in \nullspace{\bfR(\bfI - \frac{1}{N} \bmone \bmone^\intercal)}$. 

Because $\bmone^\intercal \bfu_k = 0$ for all $k \in [K - 1]$, to show that $\bmone, \bfu_1, \dots, \bfu_{K - 1}$ are linearly independent, it is enough to show that $\bfu_1, \dots, \bfu_{K - 1}$ are linearly independent. Consider the $i^{th}$ component of $\sum_{k = 1}^{K - 1} \alpha_k \bfu_k$ for arbitrary $\alpha_1 \dots, \alpha_{K - 1} \in \bbR$ and $i \in [N]$. If $v_i \in \calC_K$, then
\begin{equation*}
  \left(\sum_{k = 1}^{K - 1} \alpha_k \bfu_k \right)_i = -\frac{1}{K - 1} \sum_{k = 1}^{K - 1} \alpha_k.
\end{equation*}
Similarly, when $v_i \in \calC_{k{'}}$ for some $k{'} \in [K - 1]$, we have,
\begin{equation*}
  \left(\sum_{k = 1}^{K - 1} \alpha_k \bfu_k \right)_i = \alpha_{k{'}} -\frac{1}{K - 1} \sum_{\substack{k = 1 \\ k \neq k{'}}}^{K - 1} \alpha_{k}.
\end{equation*}
Thus, $\sum_{k = 1}^{K - 1} \alpha_k \bfu_k = 0$ implies that $-\frac{1}{K - 1} \sum_{k = 1}^{K - 1} \alpha_k = 0$ and $\alpha_{k{'}} - \frac{1}{K - 1} \sum_{k = 1, k \neq k{'}}^{K - 1} \alpha_{k} = 0$ for all $k{'} \in [K - 1]$. Subtracting the first equation from the second gives $\alpha_{k{'}} + \frac{1}{K - 1} \alpha_{k{'}} = 0$, which in turn implies that $\alpha_{k{'}} = 0$ for all $k{'} \in [K - 1]$. Thus, $\bmone, \bfu_1, \dots, \bfu_{K - 1}$ are linearly independent.

% ===================================== %

\subsection{Proof of Lemma \ref{lemma:uk_eigenvector_of_tildeA}}

Using the representation of $\tilde{\calA}$ from \eqref{eq:tilde_cal_A_def}, Lemma \ref{lemma:introducing_uks}, and the assumption on equal size of the clusters, we get,
\begin{align*}
  \tilde{\calA} \bmone &= q \bfR \bmone + s (\bmone \bmone^\intercal - \bfR) \bmone + (p - q)\sum_{k = 1}^K \bfG_k \bfR \bfG_k \bmone + (r - s) \sum_{k = 1}^K \bfG_k (\bmone \bmone^\intercal - \bfR) \bfG_k \bmone \\
  &= qd \bmone + sN \bmone - sd \bmone + (r - s)\sum_{k = 1}^K \bfG_k \bmone \bmone^\intercal \bfG_k \bmone + [(p - q) - (r - s)] \sum_{k = 1}^K \bfG_k \bfR \bfG_k \bmone \\
  &= \left[qd + s(N - d) + (p - q) \frac{d}{K} + (r - s)\frac{N - d}{K} \right] \bmone.
\end{align*}
Similarly, for any $k{'} \in [K]$,
\small
{
\begin{align*}
  \allowdisplaybreaks
  \tilde{\calA} \bfu_{k{'}} &= q \bfR \bfu_{k{'}} + s (\bmone \bmone^\intercal - \bfR) \bfu_{k{'}} + (p - q)\sum_{k = 1}^K \bfG_k \bfR \bfG_k \bfu_{k{'}} + (r - s) \sum_{k = 1}^K \bfG_k (\bmone \bmone^\intercal - \bfR) \bfG_k \bfu_{k{'}} \\
  &= 0 + 0 + (r - s) \sum_{k = 1}^K \bfG_k \bmone \bmone^\intercal \bfG_k \bfu_{k{'}} + [(p - q) - (r - s)] \sum_{k = 1}^K \bfG_k \bfR \bfG_k \bfu_{k{'}} \\
  &= \left[ (p - q) \frac{d}{K} + (r - s) \frac{N - d}{K} \right] \bfu_{k{'}}.
\end{align*}
}

% ===================================== %

\subsection{Proof of Lemma \ref{lemma:orthonormal_eigenvectors_y2_yK}}

It is easy to verify that vectors $\bfy_2, \dots, \bfy_K$ are obtained by applying the Gram-Schmidt normalization process to the vectors $\bfu_1, \dots, \bfu_{K - 1}$. Thus, $\bfy_2, \dots, \bfy_K$ span the same space as $\bfu_1, \dots, \bfu_{K - 1}$. Recall that $\bfy_1 = \bmone / \sqrt{N}$. We start by showing that $\bfy_1^\intercal \bfy_{1 + k} = 0$.
\begin{align*}
  \bfy_1^\intercal \bfy_{1 + k} = \frac{1}{\sqrt{N}} \sum_{i = 1}^N y_{(1+k)i} &= \frac{1}{\sqrt{N}} \left[ \sum_{i : v_i \in \calC_k} (K - k) q_k - \sum_{i : v_i \in \calC_{k{'}}, k{'} > k} q_k \right] \\
  &= \frac{1}{\sqrt{N}} \left[ \frac{N}{K} (K - k) q_k - \left(N - k\frac{N}{K} \right)q_k \right] = 0.
\end{align*}
Here, $q_k = \frac{1}{\sqrt{\frac{N}{K} (K - k) (K - k + 1)}}$. Now consider $\bfy_{1 + k_1}^\intercal \bfy_{1 + k_2}$ for $k_1, k_2 \in [K - 1]$ such that $k_1 \neq k_2$. Assume without loss of generality that $k_1 < k_2$.
\begin{align*}
  \bfy_{1 + k_1}^\intercal \bfy_{1 + k_2} &= \sum_{i : v_i \in \calC_{k_2}} (-q_{k_1})(K - k_2)q_{k_2} + \sum_{i : v_i \in \calC_k, k > k_2} (-q_{k_1})(-q_{k_2}) \\
  &= -q_{k_1} q_{k_2} (K - k_2) \frac{N}{K} + q_{k_1} q_{k_2} \left(N - k_2 \frac{N}{K}\right) = 0.
\end{align*}
Finally, for any $k \in [K - 1]$,
\begin{align*}
  \bfy_{1 + k}^\intercal \bfy_{1 + k} = \sum_{i : v_i \in \calC_k} (K - k)^2 q_k^2 + \sum_{i : v_i \in \calC_{k{'}}, k{'} > k} q_k^2 = q_k^2 \left[ \frac{N}{K} (K - k)^2 + N - k\frac{N}{K} \right] = 1,
\end{align*}
where the last equality follows from the definition of $q_k$.

% ===================================== %

\subsection{Proof of Lemma \ref{lemma:first_K_eigenvectors_of_L}}

Note that the columns of $\calZ$ are also the dominant $K$ eigenvectors of $\bfY^\intercal \tilde{\calA} \bfY$, as $\calZ$ is the solution to \eqref{eq:optimization_problem} with $\bfL$ set to $\calL$. The calculations below show that for all $k \in [K]$, $\bfe_k \in \bbR^{N - r}$, the $k^{th}$ standard basis vector, is an eigenvector of $\bfY^\intercal \tilde{\calA} \bfY$ with eigenvalue $\lambda_k$, where $\lambda_1, \dots, \lambda_K$ are defined in Lemma \ref{lemma:uk_eigenvector_of_tildeA}.
\begin{equation*}
  \bfY^\intercal \tilde{\calA} \bfY \bfe_k = \bfY^\intercal \tilde{\calA} \bfy_k = \lambda_k \bfY^\intercal \bfy_k = \lambda_k \bfe_k.
\end{equation*}
The second equality follows from Lemma \ref{lemma:uk_eigenvector_of_tildeA} because $\bfy_2, \dots, \bfy_K \in \spanof{\bfu_1, \dots, \bfu_{K - 1}}$, and $\bfu_1, \dots, \bfu_{K - 1}$ are all eigenvectors of $\tilde{\calA}$ with the same eigenvalue. To show that the columns of $\bfY \calZ$ lie in the span of $\bfy_1, \dots, \bfy_K$, it is enough to show that $\bfe_1, \dots, \bfe_K$ are the dominant $K$ eigenvectors of $\bfY^\intercal \tilde{\calA} \bfY$.

Let $\bmalpha \in \bbR^{N - r}$ be an eigenvector of $\bfY^\intercal \tilde{\calA} \bfY$ such that $\bmalpha \notin \spanof{\bfe_1, \dots, \bfe_K}$ and $\norm{\bmalpha}[2][2] = 1$. Then, because $\bfY^\intercal \tilde{\calA} \bfY$ is symmetric, $\bmalpha^\intercal \bfy_1 = 0$, i.e. $\alpha_1 = 0$, where $\alpha_i$ denotes the $i^{th}$ element of $\bmalpha$. The eigenvalue corresponding to $\bmalpha$ is given by
\begin{equation*}
  \lambda_\alpha = \bmalpha^\intercal \bfY^\intercal \tilde{\calA} \bfY \bmalpha.
\end{equation*}
Let $\bfx = \bfY \bmalpha = \sum_{i = 1}^{N - r} \alpha_i \bfy_i$, then $\lambda_{\alpha} = \bfx^\intercal \tilde{\calA} \bfx$. Using the definition of $\tilde{\calA}$ from \eqref{eq:tilde_cal_A_def}, we get,
{
\small
\begin{equation}
  \label{eq:x_calA_x}
  \bfx^\intercal \tilde{\calA} \bfx = (q - s)\bfx^\intercal \bfR \bfx + s \bfx^\intercal \bmone \bmone^\intercal \bfx + [(p - q) - (r - s)]\sum_{k = 1}^K \bfx^\intercal \bfG_k \bfR \bfG_k \bfx + (r - s) \sum_{k = 1}^K \bfx^\intercal \bfG_k \bmone \bmone^\intercal \bfG_k \bfx.
\end{equation}
}%
We will consider each term in \eqref{eq:x_calA_x} separately. Before that, note that $\bfy_2, \dots, \bfy_{N - r} \in \nullspace{\bfR}$. This is because $\bfy_1 = \bmone / \sqrt{N}$ and $\bfy_2, \dots, \bfy_{N - r}$ are orthogonal to $\bfy_1$. Thus,
\begin{equation}
  \label{eq:null_space_F}
  \bfR (\bfI - \bmone \bmone^\intercal / N) \bfy_i = 0 \Rightarrow \bfR(\bfI - \bfy_1 \bfy_1^\intercal) \bfy_i = 0 \Rightarrow \bfR \bfy_i = 0, \,\,\,\,\, i = 2, 3, \dots, N - r.
\end{equation}
Now consider the first term in \eqref{eq:x_calA_x}.
\begin{equation*}
  \bfx^\intercal \bfR \bfx = \sum_{i = 1}^{N - r} \sum_{j = 1}^{N - r} \alpha_i \alpha_j \bfy_i^\intercal \bfR \bfy_j = \alpha_1^2 \bfy_1^\intercal \bfR \bfy_1 = 0.
\end{equation*}
Here, the second equality follows from \eqref{eq:null_space_F}, and the third equality follows as $\alpha_1 = 0$. Similarly, for the second term in \eqref{eq:x_calA_x},
\begin{equation*}
  \bfx^\intercal \bmone \bmone^\intercal \bfx = N \bfx^\intercal \bfy_1 \bfy_1^\intercal \bfx = N \sum_{i = 1}^{N - r} \sum_{j = 1}^{N - r} \alpha_i \alpha_j \bfy_i^\intercal \bfy_1 \bfy_1^\intercal \bfy_j = N \alpha_1^2 (\bfy_1^\intercal \bfy_1)^2 = 0.
\end{equation*}
Note that $\bfG_k = \bfG_k \bfG_k$ as $\bfG_k$ is a diagonal matrix with either $0$ or $1$ on its diagonal. For the third term in \eqref{eq:x_calA_x},
\begin{equation}
  \label{eq:x_Fk_x}
  \bfx^\intercal \bfG_k \bfR \bfG_k \bfx = \bfx^\intercal \bfG_k \bfG_k \bfR \bfG_k \bfG_k \bfx = \bfx_{[k]} \bfR_{[k]} \bfx_{[k]} \leq \frac{d}{K} \norm{\bfx_{[k]}}[2][2],
\end{equation}
where $\bfx_{[k]} \in \bbR^{N/K}$ contains elements of $\bfx$ corresponding to vertices in $\calC_k$. Similarly, $\bfR_{[k]} \in \bbR^{N/K \times N/K}$ contains the submatrix of $\bfR$ restricted to rows and columns corresponding to vertices in $\calC_k$. The last inequality holds because $\bfR_{[k]}$ is a $d/K$-regular graph by Assumption \ref{assumption:R_is_d_regular}, hence its maximum eigenvalue is $d/K$. Further, 
\begin{equation*}
  \sum_{k = 1}^K \bfx^\intercal \bfG_k \bfR \bfG_k \bfx \leq \frac{d}{K} \sum_{k = 1}^K \norm{\bfx_{[k]}}[2][2] = \frac{d}{K} \norm{\bfx}[2][2] = \frac{d}{K}.
\end{equation*}
Similarly, for the fourth term in \eqref{eq:x_calA_x},
\begin{equation*}
  \bfx^\intercal \bfG_k \bmone \bmone^\intercal \bfG_k \bfx = \bfx^\intercal \bfG_k \bfG_k \bmone \bmone^\intercal \bfG_k \bfG_k \bfx = \bfx_{[k]}^\intercal \bmone_{N/K} \bmone_{N/K}^\intercal \bfx_{[k]} \leq \frac{N}{K} \norm{\bfx_{[k]}}[2][2].
\end{equation*}
Here, $\bmone_{N/K} \in \bbR^{N/K}$ is an all-ones vector and the last inequality holds because $\bmone_{N/K} \bmone_{N/K}^\intercal$ is a $N/K$-regular graph. Because $\bfx_{[k]} \notin \spanof{\bfy_1, \dots, \bfy_K}$, there is at least one $k \in [K]$ for which $\bfx_{[k]}$ is not a constant vector (if this was not true, $\bfx_{[k]}$ will belong to span of $\bfy_1, \dots, \bfy_K$). Thus, at least for one $k \in [K]$, $\bfx^\intercal \bfG_k \bmone \bmone^\intercal \bfG_k \bfx < \frac{N}{K} \norm{\bfx_{[k]}}[2][2]$. Summing over $k \in [K]$, we get,
\begin{equation*}
  \sum_{k = 1}^K \bfx^\intercal \bfG_k \bmone \bmone^\intercal \bfG_k \bfx <  \frac{N}{K} \sum_{k = 1}^K \norm{\bfx_{[k]}}[2][2] = \frac{N}{K} \norm{\bfx}[2][2] = \frac{N}{K}.
\end{equation*}
Adding the four terms we get the following bound. For eigenvector $\bmalpha$ of $\bfY^\intercal \tilde{\calA} \bfY$ such that $\bmalpha \notin \spanof{\bfe_1, \dots, \bfe_K}$ and $\norm{\bmalpha}[2][2] = 1$,
\begin{equation}
  \label{eq:x_calA_x_bound}
  \lambda_\alpha = \bfx^\intercal \tilde{\calA} \bfx < [(p - q) - (r - s)] \frac{d}{K} + (r - s) \frac{N}{K} = \lambda_K.
\end{equation}
Thus, $\lambda_1, \dots, \lambda_K$ are the highest $K$ eigenvalues of $\bfY^\intercal \tilde{\calA} \bfY$ and hence $\bfe_1, \dots, \bfe_K$ are the top $K$ eigenvectors. Thus, the columns of $\bfY \calZ$ lie in the span of $\bfy_1, \dots, \bfy_K$.

% ===================================== %

\subsection{Proof of Lemma \ref{lemma:bound_on_D-calD}}
As $\bfD$ and $\calD$ are diagonal matrices, $\norm{\bfD - \calD} = \max_{i \in [N]} \abs{D_{ii} - \calD_{ii}}$. Applying union bound, we get,
\begin{equation*}
  \rmP(\max_{i \in [N]} \abs{D_{ii} - \calD_{ii}} \geq \epsilon) \leq \sum_{i = 1}^N \rmP(\abs{D_{ii} - \calD_{ii}} \geq \epsilon).
\end{equation*}
We consider an arbitrary term in this summation. For any $i \in [N]$, note that $D_{ii} = \sum_{j \neq i} A_{ij}$ is a sum of independent Bernoulli random variables such that $\rmE[D_{ii}] = \calD_{ii}$. We consider two cases depending on the value of $p$.

\paragraph*{Case 1: $p > \frac{1}{2}$} By Hoeffding's inequality,
\begin{equation*}
  \rmP(\abs{D_{ii} - \calD_{ii}} \geq \epsilon) \leq 2 \exp \left( -\frac{2\epsilon^2}{N} \right).
\end{equation*}
Setting $\epsilon = \sqrt{2(\alpha + 1)} \sqrt{p N \ln N}$, we get for any $\alpha > 0$,
\begin{equation*}
  \rmP(\abs{D_{ii} - \calD_{ii}} \geq \sqrt{2(\alpha + 1)} \sqrt{p N \ln N}) \leq 2 \exp \left( - \frac{4p(\alpha + 1)N \ln N}{N} \right) \leq N^{-(\alpha + 1)}.
\end{equation*}

\paragraph*{Case 2: $p \leq \frac{1}{2}$}  By Bernstein's inequality, as $\abs{A_{ij} - \calA_{ij}} \leq 1$ for all $i, j \in [N]$,
\begin{equation*}
  \rmP(\abs{D_{ii} - \calD_{ii}} \geq \epsilon) \leq 2 \exp \left( -\frac{\epsilon^2/2}{\sum_{j \neq i} \rmE[(A_{ij} - \calA_{ij})^2]  + \epsilon/3} \right).
\end{equation*}
Also note that,
\begin{equation*}
  \rmE[(A_{ij} - \calA_{ij})^2] \leq \calA_{ij}(1 - \calA_{ij})^2 + (1 - \calA_{ij})(-\calA_{ij})^2 = \calA_{ij} (1 - \calA_{ij}) \leq \calA_{ij} \leq p.
\end{equation*}
Thus,
\begin{equation*}
  \rmP(\abs{D_{ii} - \calD_{ii}} \geq \epsilon) \leq 2 \exp \left( -\frac{\epsilon^2/2}{Np  + \epsilon/3} \right).
\end{equation*}
Let $\epsilon = c \sqrt{p N \ln N}$ for some constant $c > 0$ and assume that $p \geq C\frac{\ln N}{N}$ for some $C > 0$. We get,
\begin{align*}
  2 \exp \left( -\frac{\epsilon^2/2}{Np  + \epsilon/3} \right) = 2 \exp \left(- \frac{c^2 p N \ln N}{2(Np + c \sqrt{p N \ln N}/3)} \right) &= 2 \exp \Big(- \frac{c^2 \ln N}{2(1 + \frac{c}{3} \sqrt{\frac{\ln N}{p N}})} \Big) \\
  &\leq 2 \exp \Big(- \frac{c^2 \ln N}{2(1 + \frac{c}{3 \sqrt{C}})} \Big).
\end{align*}
Let $c$ be such that $\frac{c^2}{2(1 + c/3\sqrt{C})} \geq 2 (\alpha + 1)$. Such a $c$ can always be chosen as $\underset{c \rightarrow \infty}{\lim} \frac{c^2}{2(1 + c/3\sqrt{C})} = \infty$. Then,
\begin{equation*}
  \rmP(\abs{D_{ii} - \calD_{ii}} \geq \epsilon) \leq N^{-(\alpha + 1)}.
\end{equation*}

Thus, there always exists a constant $\const_1(C, \alpha)$ that depends only on $C$ and $\alpha$ such that for all $\alpha > 0$ and for all values of $p \geq C \ln N / N$,
\begin{equation*}
  \rmP(\abs{D_{ii} - \calD_{ii}} \geq \const_1(C, \alpha) \sqrt{p N \ln N}) \leq N^{-(\alpha + 1)}.
\end{equation*}
Applying the union bound over all $i \in [N]$ yields the desired result.

% ===================================== %

\subsection{Proof of Lemma \ref{lemma:bound_on_A-calA}}

Note that $\max_{i, j \in [N]} \calA_{ij} = p$. Define $g = p N = N \max_{i, j \in [N]} \calA_{ij}$. Note that, $g \geq C \ln N$ as $p \geq C\frac{\ln N}{N}$. By Theorem 5.2 from \cite{LeiEtAl:2015:ConsistencyOfSpectralClusteringInSBM}, for any $\alpha > 0$, there exists a constant $\const_2(C, \alpha)$ such that, $$\norm{\bfA - \calA} \leq \const_2(C, \alpha) \sqrt{g} = \const_2(C, \alpha) \sqrt{pN}$$ with probability at least $1 - N^{-\alpha}$.

% ===================================== %

\subsection{Proof of Lemma \ref{lemma:bound_on_eigenvector_diff}}

Because $\bfY^\intercal \bfY = \bfI$, for any orthonormal matrix $\bfU \in \bbR^{K \times K}$ such that $\bfU \bfU^\intercal = \bfU^\intercal \bfU = \bfI$, $$\norm{\bfY \calZ - \bfY \bfZ \bfU}[F][2] = \norm{\bfY(\calZ - \bfZ \bfU)}[F][2] = \trace{(\calZ - \bfZ \bfU)^\intercal \bfY^\intercal \bfY (\calZ - \bfZ \bfU)} = \norm{\calZ - \bfZ \bfU}[F][2].$$ Thus, it is enough to show an upper bound on $\norm{\calZ - \bfZ \bfU}[F]$, where recall that columns of $\calZ \in \bbR^{N - r \times K}$ and $\bfZ \in \bbR^{N - r \times K}$ contain the leading $K$ eigenvectors of $\bfY^\intercal \calL \bfY$ and $\bfY^\intercal \bfL \bfY$ respectively. Thus, $$\inf_{\bfU \in \bbR^{K \times K} : \bfU \bfU^\intercal = \bfU^\intercal \bfU = \bfI} \norm{\bfY \calZ - \bfY \bfZ \bfU}[F] = \inf_{\bfU \in \bbR^{K \times K} : \bfU \bfU^\intercal = \bfU^\intercal \bfU = \bfI} \norm{\calZ - \bfZ \bfU}[F].$$ By equation (2.6) and Proposition 2.2 in \cite{VuLei:2013:MinimaxSparsePrincipalSubspaceEstimationInHighDimensions}, $$\inf_{\bfU \in \bbR^{K \times K} : \bfU \bfU^\intercal = \bfU^\intercal \bfU = \bfI}\norm{\calZ - \bfZ\bfU}[F] \leq \sqrt{2} \norm{\calZ\calZ^\intercal (\bfI - \bfZ \bfZ^\intercal)}[F].$$
Moreover, $\norm{\calZ\calZ^\intercal (\bfI - \bfZ \bfZ^\intercal)}[F] \leq \sqrt{K} \norm{\calZ\calZ^\intercal (\bfI - \bfZ \bfZ^\intercal)}$ as $\rank{\calZ\calZ^\intercal (\bfI - \bfZ \bfZ^\intercal)} \leq K$. Thus, we get, 
\begin{equation}
  \label{eq:Z-ZU_leq_ZZ_I-ZZ}
  \inf_{\bfU \in \bbR^{K \times K} : \bfU\bfU^\intercal = \bfU^\intercal \bfU = \bfI}\norm{\calZ - \bfZ\bfU}[F] \leq \sqrt{2K} \norm{\calZ\calZ^\intercal (\bfI - \bfZ \bfZ^\intercal)}.
\end{equation}

Let $\mu_1 \leq \mu_2 \leq \dots \leq \mu_{N - r}$ be eigenvalues of $\bfY^\intercal \calL \bfY$ and $\alpha_1 \leq \alpha_2 \leq \dots \leq \alpha_{N - r}$ be eigenvalues of $\bfY^\intercal \bfL \bfY$. By Weyl's perturbation theorem, $$\abs{\mu_i - \alpha_i} \leq \norm{\bfY^\intercal \calL \bfY - \bfY^\intercal \bfL \bfY}, \,\,\,\, \forall i \in [N - r].$$ Define $\gamma = \mu_{K + 1} - \mu_{K}$ to be the eigen-gap between the $K^{th}$ and $(K + 1)^{th}$ eigenvalues of $\bfY^\intercal \calL \bfY$. 

\paragraph*{Case 1: $\norm{\bfY^\intercal \calL \bfY - \bfY^\intercal \bfL \bfY} \leq \frac{\gamma}{4}$} If $\norm{\bfY^\intercal \calL \bfY - \bfY^\intercal \bfL \bfY} \leq \frac{\gamma}{4}$, then $\abs{\mu_i - \alpha_i} \leq \frac{\gamma}{4}$ for all $i \in [N - r]$ by the inequality given above. Thus, $\alpha_1, \alpha_2, \dots, \alpha_K \in [0, \mu_K + \frac{\gamma}{4}]$ and $\alpha_{K + 1}, \alpha_{K + 2}, \dots, \alpha_{N - r} \in [\mu_{K + 1} - \frac{\gamma}{4}, \infty)$. Let $S = [0, \mu_K + \frac{\gamma}{4}]$, then $\mu_1, \dots, \mu_K \in S$ and $\alpha_{K + 1}, \dots, \alpha_{N - r} \notin S$. Define $\delta$ as, $$\delta = \min\{ \abs{\alpha_i - s}, \alpha_i \notin S, s \in S\}.$$ Then, $\delta \geq [\mu_{K + 1} - \gamma/4] - [\mu_{K} + \gamma/4] = \gamma/2$. By Davis-Kahan $\sin \Theta$ theorem, $$\norm{\calZ\calZ^\intercal (\bfI - \bfZ \bfZ^\intercal)} \leq \frac{1}{\delta} \norm{\bfY^\intercal \calL \bfY - \bfY^\intercal \bfL \bfY} = \frac{2}{\gamma} \norm{\bfY^\intercal \calL \bfY - \bfY^\intercal \bfL \bfY}.$$

\paragraph*{Case 2: $\norm{\bfY^\intercal \calL \bfY - \bfY^\intercal \bfL \bfY} > \frac{\gamma}{4}$} Note that $\norm{\calZ\calZ^\intercal (\bfI - \bfZ \bfZ^\intercal)} \leq 1$ as, $$\norm{\calZ\calZ^\intercal (\bfI - \bfZ \bfZ^\intercal)} \leq \norm{\calZ\calZ^\intercal}\,\,\norm{\bfI - \bfZ \bfZ^\intercal} = 1 . 1 = 1.$$ Thus, if $\norm{\bfY^\intercal \calL \bfY - \bfY^\intercal \bfL \bfY} > \frac{\gamma}{4}$, then, $$\norm{\calZ\calZ^\intercal (\bfI - \bfZ \bfZ^\intercal)} \leq \frac{4}{\gamma} \norm{\bfY^\intercal \calL \bfY - \bfY^\intercal \bfL \bfY}.$$

In both cases, $\norm{\calZ\calZ^\intercal (\bfI - \bfZ \bfZ^\intercal)} \leq \frac{4}{\gamma} \norm{\bfY^\intercal \calL \bfY - \bfY^\intercal \bfL \bfY}$. Using \eqref{eq:L-calL_bound} and \eqref{eq:Z-ZU_leq_ZZ_I-ZZ}, we get with probability at least $1 - 2N^{-\alpha}$, $$\inf_{\bfU \in \bbR^{K \times K} : \bfU\bfU^\intercal = \bfU^\intercal \bfU = \bfI}\norm{\calZ - \bfZ\bfU}[F] \leq \const_3(C, \alpha) \frac{4\sqrt{2K}}{\gamma} \sqrt{p N \ln N}.$$

% ===================================== %

\subsection{Proof of Lemma \ref{lemma:k_means_error}}

Equation \eqref{eq:num_mistakes_bound} directly follows from Lemma 5.3 in \cite{LeiEtAl:2015:ConsistencyOfSpectralClusteringInSBM}. We only need to show that $$\frac{8 (2 + \epsilon)}{\delta^2} \norm{\bfX \bfU - \bfY \calZ}[F][2] < \frac{N}{K}.$$ Equation \eqref{eq:correct_solution_on_non-mistakes} then follows from Lemma 5.3 in \cite{LeiEtAl:2015:ConsistencyOfSpectralClusteringInSBM}. Recall that $\delta = \sqrt{\frac{2K}{N}}$. Using Lemma \ref{lemma:bound_on_eigenvector_diff}, we get
\begin{equation*}
  \frac{8 (2 + \epsilon)}{\delta^2} \norm{\bfX \bfU^\intercal - \bfY \calZ}[F][2] \leq \const_3(C, \alpha)^2 \frac{128(2 + \epsilon)}{\gamma^2} p N^2 \ln N < \frac{N}{K}.
\end{equation*}
Here, the last inequality follows from the assumption that $\gamma^2 > \const_3(C, \alpha)^2 . 128 (2 + \epsilon) p NK \ln N$.

% ===================================== %

\subsection{Proof of Lemma \ref{lemma:calQ-Q_bound}}

We begin by showing a simple result. Let $a, b > 0$. Then,
\begin{equation*}
    \abs{\sqrt{a} - \sqrt{b}} = \frac{\abs{(\sqrt{a} - \sqrt{b})(\sqrt{a} + \sqrt{b})}}{\sqrt{a} + \sqrt{b}} = \frac{\abs{a - b}}{\sqrt{a} + \sqrt{b}} \leq \frac{\abs{a - b}}{\sqrt{b}}.
\end{equation*}
Further,
\begin{equation*}
    \Big\vert \frac{1}{\sqrt{a}} - \frac{1}{\sqrt{b}}  \Big\vert = \frac{\abs{\sqrt{a} - \sqrt{b}}}{\sqrt{ab}} \leq \frac{\abs{a - b}}{b\sqrt{a}}.
\end{equation*}

Coming back to the bound on $\norm{\calQ^{-1} - \bfQ^{-1}}$, note that, as $\calQ^{-1} = (\lambda_1 - p)^{-1/2} \bfI$, we have that
\begin{equation*}
    \norm{\calQ^{-1} - \bfQ^{-1}} = \max \left\{ \Big\vert \nu_i - \frac{1}{\sqrt{\lambda_1 - p}} \Big\vert \;:\; \nu_i \text{ is an eigenvalue of } \bfQ^{-1} \right\}.
\end{equation*}
As $\bfQ = \sqrt{\bfY^\intercal \bfD \bfY}$, the eigenvalues of $\bfQ^{-1}$ are given by $1/\sqrt{\mu{'}_1}$, \dots, $1/\sqrt{\mu{'}_{N - r}}$, where, $\mu{'}_1$, \dots, $\mu{'}_{N - r}$ are the eigenvalues of $\bfY^\intercal \bfD \bfY$. Moreover, by substituting $a = \mu{'}_i$ and $b = \lambda_1 - p$ in the inequality derived above, we get
\begin{equation}
    \label{eq:Q_calQ_eigenvalue_diff}
    \Big\vert \frac{1}{\sqrt{\mu{'}_i}} - \frac{1}{\sqrt{\lambda_1 - p}} \Big\vert \leq \frac{\abs{\mu{'}_i - (\lambda_1 - p)}}{(\lambda_1 - p) \sqrt{\mu{'}_i}}, \;\;\; \forall\; i \in [N - r].
\end{equation}
By Weyl's perturbation theorem, for any $i \in [N - r]$,
\begin{equation*}
    \abs{\mu{'}_i - (\lambda_1 - p)} \leq \norm{\bfY^\intercal \bfD \bfY - \bfY^\intercal \calD \bfY} \leq \norm{\bfD - \calD},
\end{equation*}
where the last inequality follows as $\bfY^\intercal \bfY = \bfI$. Let us assume for now that $\norm{\bfD - \calD} \leq \frac{\lambda_1 - p}{2}$ (we prove this below). Then, $\abs{\mu{'}_i - (\lambda_1 - p)} \leq \frac{\lambda_1 - p}{2}$ for all $i \in [N - r]$. Hence,
\begin{equation*}
    \mu{'}_i \geq \frac{\lambda_1 - p}{2}, \;\;\; \forall \; i \in [N - r].
\end{equation*}
Using this in \eqref{eq:Q_calQ_eigenvalue_diff} results in
\begin{equation*}
    \Big\vert \frac{1}{\sqrt{\mu{'}_i}} - \frac{1}{\sqrt{\lambda_1 - p}} \Big\vert \leq \frac{\sqrt{2}}{\sqrt{(\lambda_1 - p)^3}} \norm{\bfD - \calD}, \;\;\; \forall\; i \in [N - r].
\end{equation*}
Taking the maximum over all $i \in [N - r]$ yields the desired result. Next, we prove that $\norm{\bfD - \calD} \leq \frac{\lambda_1 - p}{2}$.

Recall from Lemma \ref{lemma:bound_on_D-calD} that $\norm{\bfD - \calD} \leq \const_1(C, \alpha) \sqrt{p N \ln N}$, where the proof of Lemma \ref{lemma:bound_on_D-calD} requires that $\const_1(C, \alpha)$ satisfies the following condition:
\begin{equation*}
    \frac{\const_1(C, \alpha)^2}{2\left(1 + \frac{\const_1(C, \alpha)}{3\sqrt{C}}\right)} \geq 2(\alpha + 1).
\end{equation*}
Additionally, to show that $\norm{\bfD - \calD} \leq \frac{\lambda_1 - p}{2}$, we need to ensure that $\const_1(C, \alpha) \leq \frac{\lambda_1 - p}{2 \sqrt{p N \ln N}}$. A constant $\const_1(C, \alpha)$ that satisfies both these conditions exists if:
\begin{equation*}
    \frac{(\lambda_1 - p)^2 / 4pN \ln N}{2\left(1 + \frac{(\lambda_1 - p) / 2\sqrt{pN \ln N}}{3\sqrt{C}} \right)} \geq 2(\alpha + 1).
\end{equation*}
Simplifying the expression above results in
\begin{equation*}
    \frac{1}{\left(\frac{\sqrt{pN \ln N}}{\lambda_1 - p}\right)\left(\frac{\sqrt{pN \ln N}}{\lambda_1 - p} + \frac{1}{6\sqrt{C}} \right)} \geq 16(\alpha + 1).
\end{equation*}
The assumption made in the lemma guarantees that such a condition is satisfied. Hence, $\const_1(C, \alpha)$ can be set such that $\norm{\bfD - \calD} \leq \frac{\lambda_1 - p}{2}$.

% ===================================== %

\subsection{Proof of Lemma \ref{lemma:eigenvector_diff_bound_normalized}}

As in the proof of Lemma \ref{lemma:bound_on_eigenvector_diff}, because $\bfY^\intercal \bfY = \bfI$, for any orthonormal matrix $\bfU \in \bbR^{K \times K}$ such that $\bfU^\intercal \bfU = \bfU \bfU^\intercal = \bfI$,
\begin{equation*}
    \norm{\bfY \calQ^{-1} \calZ - \bfY \bfQ^{-1} \bfZ \bfU}[F] =  \norm{\calQ^{-1} \calZ - \bfQ^{-1} \bfZ \bfU}[F].
\end{equation*}
As $\calQ, \bfQ \in \bbR^{N - r \times N - r}$ and $\calZ, \bfZ \in \bbR^{N - r \times K}$, we have that $\rank{\calQ^{-1} \calZ} \leq K$ and $\rank{\bfQ^{-1} \bfZ \bfU} \leq K$, and hence $\rank{\calQ^{-1} \calZ - \bfQ^{-1} \bfZ \bfU} \leq 2K$. Therefore,
\begin{equation*}
    \norm{\calQ^{-1} \calZ - \bfQ^{-1} \bfZ \bfU}[F] \leq \sqrt{2K} \norm{\calQ^{-1} \calZ - \bfQ^{-1} \bfZ \bfU}.
\end{equation*}
Moreover, using $\calQ^{-1} = (\sqrt{\lambda_1 - p})^{-1} \bfI$ and Lemma \ref{lemma:calQ-Q_bound},
\begin{align*}
    \norm{\calQ^{-1} \calZ - \bfQ^{-1} \bfZ \bfU} &\leq \norm{\calQ^{-1}} \cdot \norm{\calZ - \bfZ \bfU} + \norm{\calQ^{-1} - \bfQ^{-1}} \cdot \norm{\bfZ \bfU} \\
    &\leq \frac{1}{\sqrt{\lambda_1 - p}} \norm{\calZ - \bfZ \bfU} + \sqrt{\frac{2}{(\lambda_1 - p)^3}} \norm{\bfD - \calD} \cdot \norm{\bfZ \bfU}.
\end{align*}
Note that $\norm{\bfZ \bfU} = \sqrt{\lambdamax{\bfU^\intercal \bfZ^\intercal \bfZ \bfU}} = \sqrt{\lambdamax{\bfU^\intercal \bfU}} = \sqrt{\lambdamax{\bfI}} = 1$. Also note that $\norm{\calZ - \bfZ \bfU} \leq \norm{\calZ - \bfZ \bfU}[F]$. Combining all of this information, we get,
{
\small
\begin{eqnarray*}
    \inf_{\bfU : \bfU^\intercal \bfU = \bfU\bfU^\intercal = \bfI} \norm{\bfY \calQ^{-1} \calZ - &\bfY& \bfQ^{-1} \bfZ \bfU}[F] \leq \\ &&\sqrt{\frac{2K}{\lambda_1 - p}} \inf_{\bfU : \bfU^\intercal \bfU = \bfU\bfU^\intercal = \bfI} \norm{\calZ - \bfZ \bfU}[F] + \sqrt{\frac{4K}{(\lambda_1 - p)^3}} \norm{\bfD - \calD}.
\end{eqnarray*}
}%
Let $\mu_1 \leq \mu_2 \leq \dots \leq \mu_{N - r}$ be eigenvalues of $\calQ^{-1} \bfY^\intercal \calL \bfY \calQ^{-1}$ and $\alpha_1 \leq \alpha_2 \leq \dots \leq \alpha_{N - r}$ be eigenvalues of $\bfQ^{-1} \bfY^\intercal \bfL \bfY \bfQ^{-1}$. Define $\gamma = \mu_{K + 1} - \mu_K$. Using a strategy similar to the one used in the proof of Lemma \ref{lemma:bound_on_eigenvector_diff}, we get:
\begin{equation*}
    \inf_{\bfU : \bfU^\intercal \bfU = \bfU\bfU^\intercal = \bfI} \norm{\calZ - \bfZ \bfU}[F] \leq \frac{4\sqrt{2K}}{\gamma} \norm{\calQ^{-1} \bfY^\intercal \calL \bfY \calQ^{-1} - \bfQ^{-1} \bfY^\intercal \bfL \bfY \bfQ^{-1}}.
\end{equation*}
Using \eqref{eq:calQYcalLYcalQ-QYLYQ_bound} and Lemma \ref{lemma:bound_on_D-calD} results in
{
\small
\begin{align*}
    \inf_{\bfU : \bfU^\intercal \bfU = \bfU\bfU^\intercal = \bfI}  &\norm{\calZ - \bfZ \bfU}[F] \\
    \leq& \frac{8\sqrt{2K}}{\gamma (\lambda_1 - p)} \Bigg[ \left( \frac{(\lambda_1 - \bar{\lambda}) \const_1(C, \alpha) \sqrt{2}}{\lambda_1 - p} + \frac{\const_3(C, \alpha)}{2}\right) \sqrt{pN \ln N} + \\
    &\left( \frac{(\lambda_1 - \bar{\lambda}) \const_1(C, \alpha)^2}{\lambda_1 - p} + \const_1(C, \alpha) \const_3(C, \alpha) \sqrt{2} \right) \frac{p N \ln N}{\lambda_1 - p} + \\
    &\const_1(C, \alpha)^2 \const_3(C, \alpha) \frac{(pN \ln N)^{3/2}}{(\lambda_1 - p)^2} \Bigg] \\
    \leq & \frac{8\sqrt{2K}}{\gamma (\lambda_1 - p)} \Bigg[ \frac{(\lambda_1 - \bar{\lambda}) \const_1(C, \alpha) \sqrt{2}}{\lambda_1 - p} + \frac{\const_3(C, \alpha)}{2} + \\
    & \left( \frac{(\lambda_1 - \bar{\lambda}) \const_1(C, \alpha)^2}{\lambda_1 - p} + \const_1(C, \alpha) \const_3(C, \alpha) \sqrt{2} \right) \const_4(C, \alpha)  + \\
    & \const_1(C, \alpha)^2 \const_3(C, \alpha)\const_4(C, \alpha)^2 \Bigg] \sqrt{p N \ln N} \\
    \leq & \frac{8\sqrt{2K} \const_5(C, \alpha)}{\gamma (\lambda_1 - p)} \sqrt{p N \ln N}.
\end{align*}
}%
Here, the second inequality follows from the assumption that there is a constant $\const_4(C, \alpha)$ that satisfies $\frac{\sqrt{p N \ln N}}{\lambda_1 - p} \leq \const_4(C, \alpha)$. The last inequality follows by choosing $\const_5(C, \alpha)$ such that the expression between the square brackets in the second inequality is less than $\const_5(C, \alpha)$. Using the expression above, we get:
{
\small
\begin{equation*}
    \inf_{\bfU : \bfU^\intercal \bfU = \bfU\bfU^\intercal = \bfI} \norm{\bfY \calQ^{-1} \calZ - \bfY \bfQ^{-1} \bfZ \bfU}[F] \leq \left[\frac{16 K \const_5(C, \alpha)}{\gamma (\lambda_1 - p)^{3/2}} + \frac{2 \const_1(C, \alpha) \sqrt{K}}{(\lambda_1 - p)^{3/2}} \right] \sqrt{p N \ln N}.
\end{equation*}
}

\end{document}